\documentclass{article}
\usepackage{iclr2027_conference,times}
\usepackage[T1]{fontenc}
\usepackage{amsmath,amssymb,graphicx,booktabs,multirow,array,tabularx}
\usepackage{colortbl}
\usepackage{hyperref,url}
\usepackage{enumitem}
\usepackage{listings}
\usepackage{placeins}
\hypersetup{colorlinks=true,citecolor=blue!45!black,linkcolor=blue!45!black,urlcolor=blue!45!black}
\newcommand{\method}{HarnessHOI}
\newcommand{\E}{\mathcal{E}}
\newcommand{\Hh}{\mathcal{H}}
\newcommand{\Pp}{\mathcal{P}}
\newcommand{\K}{\mathcal{K}}

\newcommand{\ind}{\mathbf{1}}
\title{Harnessing Multimodal Large Language Models for Training-Free Human-Object Interaction Detection}

\author{
\textbf{
Zhaolin Cai$^{1}$ \quad
Huiyu Duan$^{1,*}$ \quad
Liu Yang$^{1}$ \quad
Yanjun Qin$^{2}$
}\\[2pt]
\textbf{
Bo Ai$^{3}$ \quad
Wei Chen$^{1}$ \quad
Xiongkuo Min$^{3,*}$ \quad
Guangtao Zhai$^{1,4}$\thanks{Corresponding authors.}
}\\[4pt]
$^{1}$Shanghai Jiao Tong University
\qquad
$^{2}$Xinjiang University
\\
$^{3}$Beijing Jiaotong University
\qquad
$^{4}$Shanghai AI Laboratory
}

\iclrfinalcopy   

\begin{document}
\maketitle

\vspace{-2mm}
\begin{abstract}
\vspace{-2mm}

Human-object interaction (HOI) detection aims to localize human-object pairs and recognize their interactions. Traditional supervised methods perform strongly but rely on task-specific training. Recent multimodal large language models (MLLMs) offer a promising route to training-free HOI detection through their broad visual-semantic knowledge and versatile perceptual and reasoning capabilities. However, existing approaches largely invoke these capabilities through loosely coordinated inference stages. This fragmented execution restricts the role of interaction hypotheses in guiding visual exploration, leaving key participants overlooked and local ambiguities unresolved. Furthermore, propagating early semantic assumptions through subsequent visual grounding and relation prediction induces self-reinforcing semantic circularity. To resolve these challenges, we propose HarnessHOI, a training-free framework that transforms passive MLLM inference into an active interaction-centric harness. Specifically, we introduce an interaction-guided perception mechanism that projects emerging interaction hypotheses back into the visual space to discover missing participants and refine ambiguous evidence through targeted observation. Furthermore, a relation-agnostic geometric adjudication module reconciles multi-source evidence to establish a unified spatial basis for grounded interaction reasoning across multiple actions and semantic roles. Extensive experiments on HICO-DET and V-COCO demonstrate that HarnessHOI achieves state-of-the-art performance among training-free methods, confirming the effectiveness of the proposed harness for complex interaction understanding. Code will be released upon publication.

\vspace{-2mm}

\end{abstract}
\vspace{-2mm}
\section{Introduction}
\vspace{-2mm}

Human-object interaction (HOI) detection aims to localize human-object pairs and recognize the interactions between them, typically represented as $\langle\mathrm{human}, \mathrm{action}, \mathrm{object}\rangle$ triplets~\citep{chao2018hicodet,gupta2015vcoco}. As a foundational task for scene understanding, it supports applications ranging from visual surveillance~\citep{dogariu2020surveillance,li2024ripple} to embodied AI~\citep{mascaro2023hoi4abot,bahl2023affordances}. Traditional supervised methods have achieved strong performance by learning task-specific interaction representations and prediction functions from HOI annotations~\citep{gkioxari2018interactnet,liao2022genvlkt}. However, their reliance on extensive annotations and dedicated training limits their scalability to the compositional and long-tailed interaction space encountered in real-world scenes~\citep{hou2020vcl,hou2021fcl}.

Recent vision-language models and multimodal large language models (MLLMs) have substantially expanded their ability to visual understanding and visual-semantic knowledge~\citep{radford2021clip,bai2025qwen25vl}. Initial approaches integrate this knowledge into HOI detectors through feature alignment, prompt adaptation, or supervised optimization~\citep{lei2024cmmp,lei2024ezhoi}. Recent work further develops part-aware interaction representations~\citep{tonini2026finehoi} and unifies HOI detection with generation~\citep{yang2026unihoi}. Although effective, these approaches still require specific training, reintroducing the annotation and optimization burden that foundation models are expected to alleviate. Training-free paradigms therefore directly exploit frozen pretrained models for interaction grounding and prediction. Retrieval-based approaches score candidate interactions against visual-semantic knowledge~\citep{tonini2025dysco}, while the variant formulates recognition as candidate-conditioned semantic assessment~\citep{xuan2026zeroshothoidetectionmllmbased}. Recent agentic methods further revisit candidate interactions and recover visual entities through multi-stage semantic reasoning~\citep{lei2026unleashing}. Despite this progress, existing approaches leave semantic reasoning and visual perception through a separate process, leaving perception passive to the emerging interaction hypotheses.

Human-object relations are fundamentally constrained by both relational semantics and fine-grained visual evidence distributed across the scene~\citep{wang2019deepcontext,wan2019pmfnet,zhang2023pvic}.An emerging interaction hypothesis provides cues to recover overlooked participants or resolve local visual ambiguity. Conversely, grounded visual evidence provides necessary spatial constraints to verify candidate relations. Existing methods fail to regulate this reciprocal dependency, resulting in two coupled limitations. First, visual perception remains static when emerging interaction hypotheses are not systematically projected back into the image space to guide observation. This creates \emph{perceptual passivity}, leaving missing entities and ambiguous regions unresolved. Second, directly propagating semantic proposals through visual grounding and subsequent relation evaluation induces \emph{semantic circularity}, where initial semantic assumptions bias the visual interpretation and produce self-reinforcing predictions and spurious relational judgments. The central challenge is therefore to harness the complementary semantic and perceptual capabilities of MLLMs within a unified framework, enabling interaction hypotheses to actively guide visual exploration while grounding the resulting judgments in instance-level visual evidence.

\begin{figure}[t]
\centering
\includegraphics[width=\linewidth]{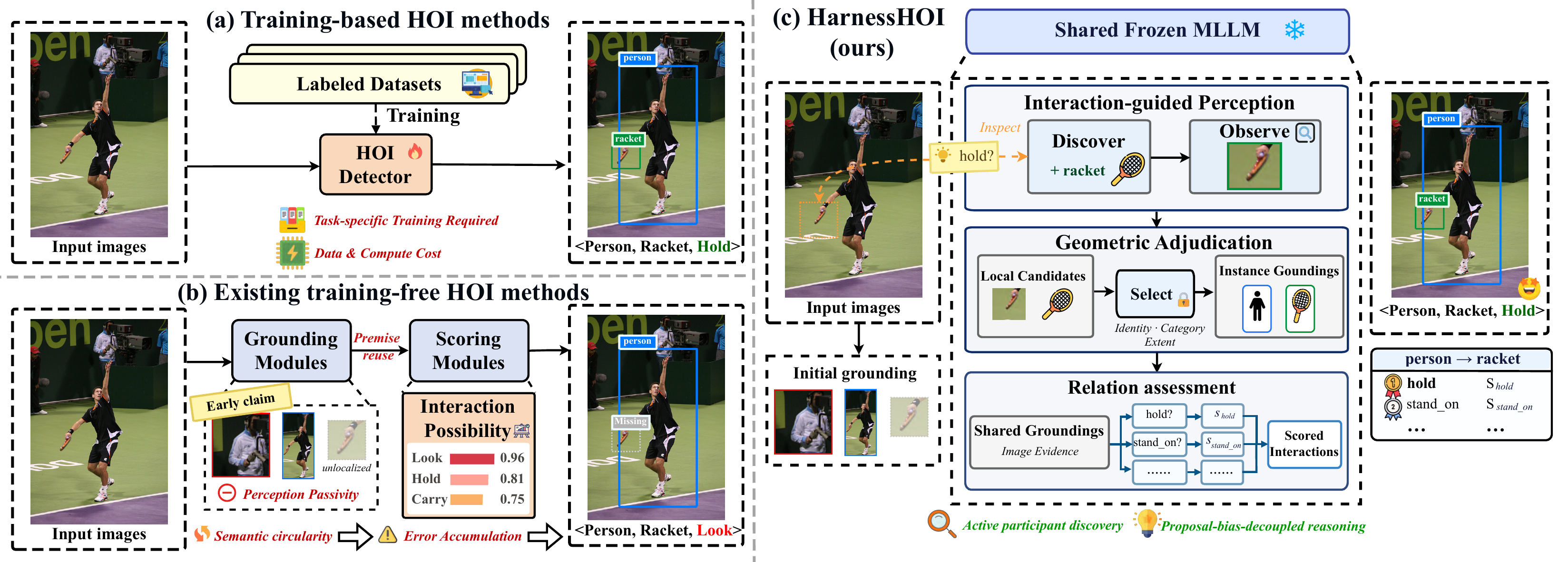}
\vspace{-2em}
\caption{Comparison of HOI methods. (a) Training-based HOI methods learn
from annotated interactions but require task-specific training. (b) Existing training-free methods rely on loosely coordinated perception and reasoning modules, resulting in passive visual exploration and semantic circularity. (c) HarnessHOI (ours) introduces interaction-centric harness to actively search and select visual evidence, and establish reliable grounding for final HOI predictions.}
\vspace{-1em}
\label{fig:teaser}
\end{figure}

To address these limitations, we propose \textbf{HarnessHOI}, a training-free framework that transforms MLLM interpretation into an active interaction-centric harness paradigm(Figure~\ref{fig:teaser}). To overcome perceptual passivity, HarnessHOI incorporates \emph{interaction-guided perception} that projects emerging hypotheses back into image space, actively discovering interaction participants and requesting targeted observations for ambiguous evidence. To mitigate semantic circularity, \emph{relation-agnostic geometric adjudication} reconciles the resulting localization candidates under persistent instance identities, establishing a unified visual grounding according to participant evidence. Grounded interaction reasoning then assesses multiple actions and semantic roles using these shared groundings and the acquired visual evidence. Together, these mechanisms establish an interaction-centric harness that coordinates interaction-guided evidence acquisition with relation-agnostic visual grounding, enabling coherent and reliable training-free HOI prediction with frozen MLLMs.

Our main contributions are summarized as follows:
\begin{itemize}
\vspace{-2mm}

    \item We present \textbf{HarnessHOI}, a training-free framework that harnesses complementary perceptual and semantic capabilities of frozen MLLMs for HOI detection, moving beyond isolated semantic scoring or loosely coupled multi-stage reasoning.
\vspace{-1mm}

    \item We introduce interaction-guided perception, relation-agnostic geometric adjudication, and grounded interaction reasoning to connect hypothesis-driven visual exploration with shared participant groundings and evidence-grounded relation assessment, addressing perceptual passivity and semantic circularity.
\vspace{-1mm}

    \item Extensive experiments on HICO-DET and V-COCO demonstrate that HarnessHOI achieves state-of-the-art performance among training-free methods, validating the effectiveness of the proposed interaction-centric harness for human-object interaction detection.
\end{itemize}

\section{Related Work}
\label{sec_related_work}
\vspace{-1mm}

\subsection{Traditional Human-Object Interaction Detection}
\vspace{-1mm}

Traditional human-object interaction (HOI) detection frameworks primarily rely on fully supervised paradigms, which are broadly grouped into two-stage pipelines and single-stage set-prediction architectures~\citep{zhang2021cdn,zhang2022stip}. Two-stage methods detect human and object proposals prior to classifying interaction predicates~\citep{gao2018ican,li2019tin,zhang2021scg,zhang2022upt}. Conversely, single-stage transformer architectures utilize learnable queries to directly decode relation triplets from global visual contexts~\citep{kim2021hotr,zou2021hoitransformer,chen2021asnet,tamura2021qpic}. To improve generalization across unseen combinations, recent work incorporates pretrained vision-language representations by aligning visual features with text embeddings~\citep{yuan2022rlip,yuan2023rlipv2,ning2023hoiclip,mao2023clip4hoi}.  Although these supervised methods achieve competitive performance on closed-world benchmarks, they remain inherently constrained by task-specific parameter optimization and heavy reliance on exhaustive triplet annotations, impeding their scalability to open-world scenarios.
\vspace{-1mm}

\subsection{Foundation Models for HOI Detection}
\vspace{-1mm}

The evolution of vision-language models and multimodal large language models offers expansive visual-semantic knowledge, prompting a shift toward parameter-efficient and training-free HOI detection. Early adaptation strategies align interaction representations with pretrained visual-semantic knowledge~\citep{wu2023eoid,cao2023unihoi,lei2025hola}. Recent methods learn part-aware representations~\citep{tonini2026finehoi}, distill MLLM knowledge into HOI detectors~\citep{gao2025clhoi}, or unify detection and generation~\citep{yang2026unihoi}. Supervised fine-tuning and reinforcement learning further adapt MLLMs for interaction reasoning~\citep{wu2026hoimllm}. Although effective, these approaches still require specific optimization. Training-free methods therefore directly exploit frozen foundation models for interaction prediction. Visual-semantic retrieval approaches calculate similarity scores between representations and text embeddings~\citep{tonini2025dysco}. The variant formulates HOI detection as candidate-conditioned generation~\citep{xuan2026zeroshothoidetectionmllmbased}. Recent agentic frameworks introduce multi-agent and multi-turn reasoning to propose and revisit candidate interactions prior to visual grounding~\citep{wang2026agenthoi,lei2026unleashing}. Despite these advancements, current training-free paradigms treat semantic reasoning and visual grounding as isolated steps. As a result, visual perception remains fundamentally passive to initial semantic hypotheses, leaving models susceptible to perceptual omission and semantic circularity.
\vspace{-1mm}

\subsection{MLLM Harnessing for Visual Reasoning}
\vspace{-1mm}

Active perception principles advocate dynamically adapting visual observation based on task demands rather than relying on static image features \cite{bajcsy1988active}. Recent vision-language research operationalizes active perception by incorporating selective region re-encoding or localized attention mechanisms during MLLM inference \cite{zhang2025vicrop,shalam2026proposeattend}. In parallel, agentic workflows decompose complex visual tasks into sequential reasoning and grounding operations \cite{lei2026unleashing}. However, existing active harnessing strategies are primarily designed for free-form visual question answering or coarse region grounding, lacking effective and controllable mechanisms to manage complex relational understanding inherent to HOI detection. Applying generic workflows directly to HOI detection allows unconstrained semantic assumptions to distort visual search. HarnessHOI overcomes these limitations by unifying interaction-guided perception with relational reasoning, enabling a controlled harnessing paradigm that prevents semantic bias while maintaining rigorous spatial and relational grounding.

\vspace{-1mm}

\section{Method}
\label{sec_method}
\vspace{-1mm}

\begin{figure}[t]
\centering
\includegraphics[width=\linewidth]{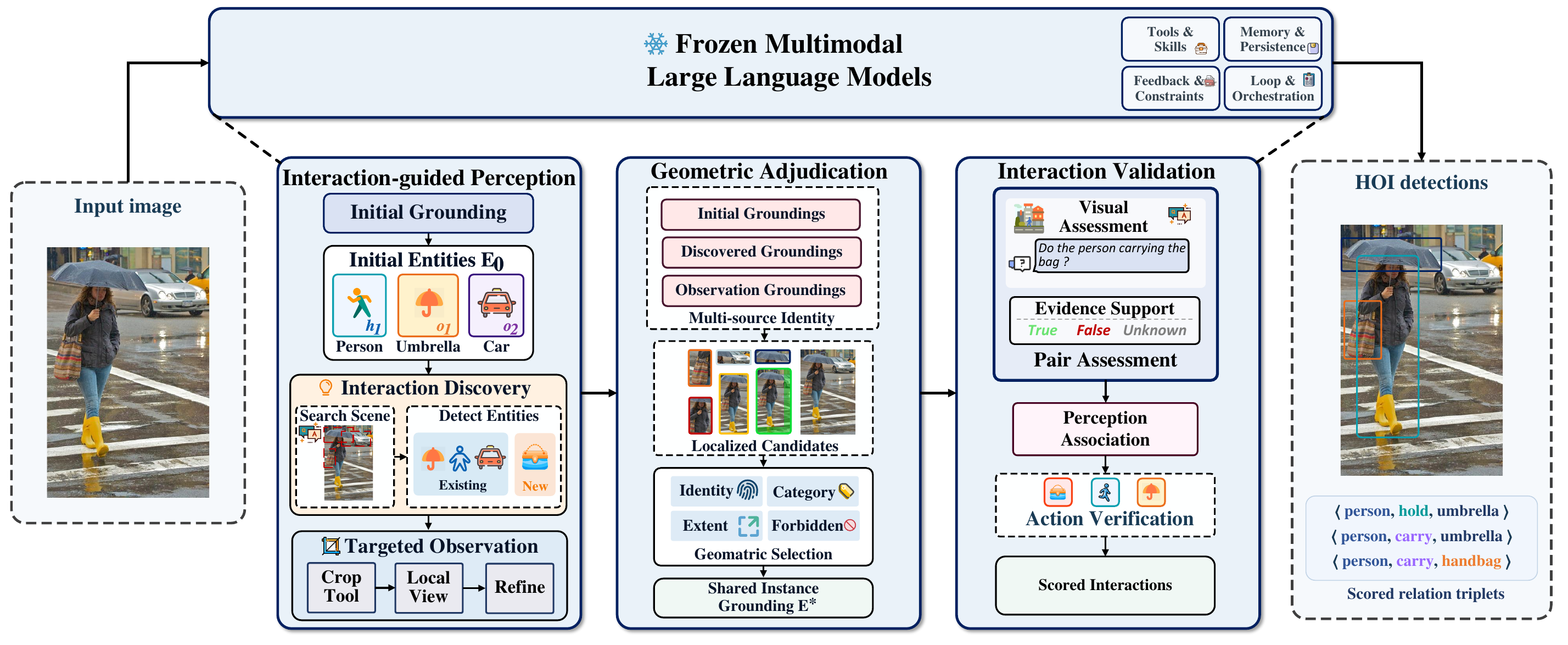}
\vspace{-2em}
\caption{\textbf{Overview of HarnessHOI.} Interaction-guided perception uses
interaction hypotheses to discover participants and acquire targeted
observations. Geometrics adjudication reconciles multisource identities to establish shared instance-level groundings. Interaction validation assesses evidence support, associates instances with semantic roles, and verifies candidate actions to produce scored HOI triplets. The frozen MLLM orchestrates all components using external tools, persistent memory, feedback constraints, and iterative loops.}
\vspace{-1em}
\label{fig:framework}
\end{figure}
\vspace{-1mm}

\subsection{Interaction-Centric HOI Harness}
\label{sec_method_overview}

HarnessHOI introduces an interaction-centric harness in which semantic
hypotheses guide participant acquisition and the acquired evidence supports
grounded relation assessment (Figure~\ref{fig:framework}). Shared participant
identities connect these operations across multiple actions and semantic roles,
allowing evidence acquired under one hypothesis to support further interaction
judgments.

\paragraph{Problem formulation.}
Given an image $I$ and an interaction taxonomy $\mathcal{T}$, our goal is to
predict a scored set of grounded interactions
$\widehat{\mathcal{Y}}=\{(k,s(k))\}$. Each entity
$e_i=(i,b_i,c_i)\in\mathcal{E}$ comprises an image-level identity, a normalized
bounding box, and a category. An interaction $k=(h,a,o)$
associates a human $h\in\mathcal{H}\subseteq\mathcal{E}$ with an action $a$
and a target $o\in\mathcal{E}$ under the semantics of $\mathcal{T}$.
The score $s(k)$ expresses visual support for the interaction between the
localized participants. We denote the MLLM by $F_\theta$, with $\theta$ fixed
throughout inference.

\paragraph{HarnessHOI overview.}
As illustrated in Figure~\ref{fig:framework}, HarnessHOI consists of
interaction-guided perception,
relation-agnostic geometric adjudication, and grounded interaction reasoning.
Specifically, we use the interaction hypotheses generated by $F_\theta$ to
guide participant discovery and targeted observation, associating the
resulting localization candidates and their source views with persistent
instance identities. We then apply relation-agnostic geometric adjudication
to select a consistent grounding for each participant according to its
identity, category, and visible extent. Based on these shared groundings
and the acquired image evidence, grounded interaction reasoning evaluates
multiple actions and semantic roles through visual question answering.
The same frozen MLLM performs all three stages, with its answer
distributions determining geometric selection and interaction scores.

\subsection{Interaction-Guided Perception}
\label{sec_interaction_guided_perception}

The semantics of an interaction provide cues to its participating objects and
their spatial relation to the human. Interaction-guided perception exploits
these cues to recover overlooked participants and obtain local observations
for precise instance grounding.

\paragraph{Hypothesis-directed discovery.}
Starting from an initial entity set $\mathcal{E}_0$ localized by $F_\theta$ in $I$,
we expand the participant set by querying the model about the visible
interactions of each human in $\mathcal{H}_0\subseteq\mathcal{E}_0$.
Conditioned on the full image, initial entities, and interaction semantics in
$\mathcal{T}$, $F_\theta$ jointly generates interaction hypotheses,
localizations of newly discovered participants, and observation requests:
\begin{equation}
    \bigl(\mathcal{P}_{D},\Delta\mathcal{E}_{D},\mathcal{R}\bigr)
    = D_\theta\bigl(I,\mathcal{E}_0,\mathcal{T}\bigr),
    \label{eq:interaction_discovery}
\end{equation}
where $D_\theta$ denotes the discovery operation of $F_\theta$.
The positive interaction hypotheses define human-target associations
$\mathcal{P}_{D}$ over the expanded participant set
$\mathcal{E}_0\cup\Delta\mathcal{E}_{D}$, where $\Delta\mathcal{E}_{D}$
contains participants newly localized in the full image.
For hypotheses involving unlocalized targets or ambiguous instance boundaries,
the observation requests $\mathcal{R}$ specify the relevant image regions and
participants for further visual inspection.

\paragraph{Targeted visual observation.}
Targeted observation complements global interaction context with local visual
evidence of participant identity and extent. Using high-resolution crops
acquired from $I$ according to $\mathcal{R}$, $F_\theta$ refines existing
localizations and grounds additional participants, conditioned on the
requested target description and local instance context. To integrate
observations across image scales, we associate candidate localizations with
persistent instance identities and express their boxes in image coordinates.
The corresponding human-target associations
form $\mathcal{P}_{O}$, connecting humans to the existing targets addressed by
their observations and to newly localized participants.

Across initial localization, discovery, and local observation, the shared
instance representation organizes visual evidence under $N$ participant
identities. Each identity $i$ has a candidate localization set
$\mathcal{B}_i=\{(b_{i,m},c_{i,m},v_{i,m})\}_{m=1}^{M_i}$. Here $b_{i,m}$
and $c_{i,m}$ specify the box and category of candidate $m$, and $v_{i,m}$
identifies its source view. The global image and acquired crops form the
shared view set $\mathcal{V}$. The candidate sets
$\{\mathcal{B}_i\}_{i=1}^{N}$ preserve alternative spatial interpretations of
each participant for geometric adjudication, while $\mathcal{V}$ provides
the global context and local detail for grounded interaction reasoning.
The observation policy and coordinate handling are detailed in
Appendix~\ref{app:policy}.

\subsection{Relation-Agnostic Geometric Adjudication}
\label{sec_geometric_adjudication}

We introduce relation-agnostic geometric adjudication to establish a
consistent localization for each participant across its interactions.
For each instance, we compare the candidate localizations against the
shared views $\mathcal{V}$ according to its identity, category, and complete
visible extent. This instance-level criterion decouples geometric selection
from support for individual actions or roles.

Let $x$ denote a visual question together with its instance context and image
evidence, and $\mathcal{L}_x$ the candidate answer set. We define the normalized
support of $F_\theta$ for an answer $\ell\in\mathcal{L}_x$ as
\begin{equation}
    q_\theta(\ell\mid x)
    = \frac{\exp\bigl(z_\ell(x)\bigr)}
           {\sum_{j\in\mathcal{L}_x}\exp\bigl(z_j(x)\bigr)},
    \label{eq:support}
\end{equation}
where $z_\ell(x)$ is the output logit of $F_\theta$ for answer label $\ell$.
For an instance with $M_i>1$, $x_i^{\mathrm{geo}}$ contains the comparison
question, candidate localizations $\mathcal{B}_i$, and shared views
$\mathcal{V}$. Each answer label $\ell_{i,m}$ identifies a candidate $m$,
and we select the candidate with the greatest visual support:
\begin{equation}
    m_i^{*}
    = \operatorname*{arg\,max}_{m=1,\ldots,M_i}
      q_\theta\bigl(\ell_{i,m}\mid x_i^{\mathrm{geo}}\bigr).
    \label{eq:geometry_selection}
\end{equation}
The selected candidate defines
$(b_i^{*},c_i^{*})=(b_{i,m_i^{*}},c_{i,m_i^{*}})$; identities with one
candidate retain that localization. Collecting these selections yields the
grounded entity set $\mathcal{E}^{*}=\{(i,b_i^{*},c_i^{*})\}_{i=1}^{N}$.
All relations involving an entity in $\mathcal{E}^{*}$ use its selected box
and category, so hypotheses concerning the same participant are evaluated
with consistent spatial grounding.

\subsection{Grounded Interaction Reasoning}
\label{sec_grounded_reasoning}

Given the shared instance groundings $\mathcal{E}^{*}$, we formulate visual
questions for candidate interactions using the action or role definitions
in $\mathcal{T}$. The frozen MLLM $F_\theta$ evaluates each question with the
global image and relevant local views from $\mathcal{V}$. Pair, action, agent,
and binding questions use the answer set
$\mathcal{L}_{\mathrm{rel}}=\{\mathrm{false},\mathrm{true},\mathrm{unknown}\}$,
representing rejection, support, and unresolved visual evidence, respectively,
with normalized support given by Equation~\ref{eq:support}.

\paragraph{Candidate pair construction.}
The relation candidates draw on both the associations established during
perception and the visual evidence across the expanded participant set.
Let $\mathcal{H}^{*}\subseteq\mathcal{E}^{*}$ denote the localized humans and
$\mathcal{P}=\{(h,o)\in\mathcal{H}^{*}\times\mathcal{E}^{*}:h\ne o\}$ the
human-entity pairs. For each pair, $F_\theta$ evaluates whether any positive
action or physical role binding in $\mathcal{T}$ is visually supported, yielding
$q_{\mathrm{F}}^{h,o}$, $q_{\mathrm{T}}^{h,o}$, and $q_{\mathrm{U}}^{h,o}$.
These assessments extend the discovery and observation associations to define
the candidate pair set:
\begin{equation}
    \begin{aligned}
        \mathcal{K}
        &= \mathcal{P}_{D}\cup\mathcal{P}_{O}\cup
        \bigl\{(h,o)\in\mathcal{P}\bigm|
        q_{\mathrm{F}}^{h,o}\le
        \max\bigl(q_{\mathrm{T}}^{h,o},q_{\mathrm{U}}^{h,o}\bigr)\bigr\}.
    \end{aligned}
    \label{eq:route}
\end{equation}

\paragraph{Multi-label action verification.}
To recognize multiple interactions involving the same participants, we
evaluate each pair in $\mathcal{K}$ over its category-compatible action set.
The action domain extends beyond the hypotheses used for participant
discovery:
\begin{equation}
    \mathcal{Q}_{+}
    = \bigl\{(h,a,o)\bigm| (h,o)\in\mathcal{K},\;
      a\in\mathcal{A}^{+}(c_o^{*})\bigr\},
    \label{eq:action_domain}
\end{equation}
where $\mathcal{A}^{+}(c)$ is the set of positive actions associated with
category $c$ in $\mathcal{T}$. For each $(h,a,o)\in\mathcal{Q}_{+}$,
$x_{h,a,o}^{\mathrm{rel}}$ denotes the visual question context for verifying
action $a$ between the selected instances $h,o$. The interaction score is
\begin{equation}
    s(h,a,o)
    = q_\theta\bigl(\mathrm{true}\mid x_{h,a,o}^{\mathrm{rel}}\bigr).
    \label{eq:hico_score}
\end{equation}
Separate normalization over $\mathcal{L}_{\mathrm{rel}}$ for each action
preserves the multi-label nature of HOI.

\paragraph{Compositional role verification.}
In semantic-role prediction, action occurrence and the visibility of its
target are distinct aspects of the interaction.
We extend the prediction to $k=(h,a,r,o)$, where $r$ specifies the semantic
role and $o\in\mathcal{E}^{*}\cup\{\varnothing\}$ accommodates physical
and null targets. To distinguish target visibility from target identity,
we decompose role prediction into action support, conditional target visibility,
and instance-specific binding. For a given interaction, we suppress the
arguments of the component supports and write
\begin{equation}
    \begin{aligned}
        q_A
        &= q_\theta\bigl(\mathrm{true}\mid x_{h,a}^{\mathrm{agent}}\bigr), \\[3pt]
        q_\sigma
        &= q_\theta\bigl(\sigma\mid x_{h,a,r}^{\mathrm{role}}\bigr), \\[3pt]
        q_B
        &= q_\theta\bigl(\mathrm{true}\mid x_{h,a,r,o}^{\mathrm{bind}}\bigr),
    \end{aligned}
    \label{eq:role_factors}
\end{equation}
where $x_{h,a}^{\mathrm{agent}}$ denotes the context for verifying whether
human $h$ performs action $a$. Target visibility is assessed in
$x_{h,a,r}^{\mathrm{role}}$ conditional on action occurrence, and the binding
of $o$ to role $r$ is assessed in $x_{h,a,r,o}^{\mathrm{bind}}$ conditional on
both action occurrence and target visibility. Action and binding questions use
$\mathcal{L}_{\mathrm{rel}}$, while the role-state answer $\sigma$ ranges over
$\mathcal{L}_{\mathrm{role}}=
\{\mathrm{missing},\mathrm{visible},\mathrm{unknown}\}$, whose outcomes denote
the absence of a visible, localizable target, the presence of such a target,
and unresolved visual evidence, respectively.

Action and role-state factors cover every human in $\mathcal{H}^{*}$ and all
corresponding actions and roles in $\mathcal{T}$, while physical binding
factors are evaluated for pairs in $\mathcal{K}$.
Writing $q_M$ and $q_V$ for the missing and visible supports, we obtain scores
for person-level actions $k=(h,a)$, physical role targets, and null targets:
\begin{equation}
    s_{\mathrm{agent}} = q_A,\qquad
    s_{\mathrm{physical}} = q_A\,q_V\,q_B,\qquad
    s_{\mathrm{NULL}} = q_A\,q_M.
    \label{eq:vcoco}
\end{equation}

We apply the action-triplet formulation to HICO-DET and the semantic-role
formulation to V-COCO. For HICO-DET, we additionally evaluate the
$\mathrm{no\_interaction}$ predicate over all pairs in $\mathcal{P}$,
using its independently assessed true support as the detection score.

The resulting prediction set $\widehat{\mathcal{Y}}$ couples each interaction
score $s(k)$ with the selected participant boxes from $\mathcal{E}^{*}$.
Participants discovered and observations acquired under individual interaction
hypotheses thus contribute to action and role verification across the expanded
scene representation. Further details on inference, implementation, and
question semantics are provided in Appendices~\ref{app:policy},
\ref{app:implementation}, and~\ref{app:prompts}.

\section{Experiments}

\subsection{Experimental Setup}
\label{sec:setup}

\paragraph{Benchmarks and metrics.}
We evaluate \method{} on the complete HICO-DET~\citep{chao2018hicodet} and V-COCO~\citep{gupta2015vcoco} test sets, containing 9,658 and 4,946 images, respectively. HICO-DET reports Full, Rare, and Non-Rare mAP over 600 HOI categories under Default and Known Object evaluation. Open-vocabulary evaluation follows RF-UC, NF-UC, UO, and UV. V-COCO reports role AP over 25 semantic roles under scenarios S1 and S2. Both benchmarks require participant matching at IoU 0.5. Detailed metric definitions and evaluation protocols appear in Appendix~\ref{app:protocol}.

\paragraph{Implementation details.}
All perceptual and reasoning operations use a single frozen Qwen3.5-9B~\citep{qwen2026qwen35}, with at most two local observations per image. The harness receives the image and public interaction vocabulary. Prompts and inference policy are selected using the official training sets, following the protocol in Appendix~\ref{app:protocol}. Inference settings, implementation, and visual questions appear in Appendices~\ref{app:policy}, \ref{app:implementation}, and~\ref{app:prompts}.

\subsection{Comparison with State-of-the-Art Methods}

\begin{table}[!t]
\caption{Comparison with state-of-the-art methods under regular settings using predicted boxes. HICO-DET reports Default and Known Object mAP (\%); V-COCO reports role AP (\%) under S1 and S2, with external entries following each source's role definitions. HarnessHOI uses the 25-role definition. Within each method group, the best results are bold; the strongest previous training-free results are underlined. Dashes denote unavailable entries.}
\vspace{-1em}
\label{tab:main}
\begin{center}
\small
\setlength{\tabcolsep}{2pt}
\renewcommand{\arraystretch}{1.08}
\resizebox{\linewidth}{!}{%
\begin{tabular}{llcccccccc}
\toprule
& & \multicolumn{3}{c}{HICO-DET Default} & \multicolumn{3}{c}{HICO-DET Known Object} & \multicolumn{2}{c}{V-COCO}\\
\cmidrule(lr){3-5}\cmidrule(lr){6-8}\cmidrule(l){9-10}
Method & Model & Full & Rare & Non-Rare & Full & Rare & Non-Rare & S1 & S2\\
\midrule
\multicolumn{10}{l}{\emph{Task-optimized methods}}\\
QPIC~\citep{tamura2021qpic} & R101 & 29.90 & 23.92 & 31.69 & 32.38 & 26.06 & 34.27 & 58.30 & 60.70\\
UPT~\citep{zhang2022upt} & R101-DC5 & 32.62 & 28.62 & 33.81 & 36.08 & 31.41 & 37.47 & 61.30 & 67.10\\
GEN-VLKT~\citep{liao2022genvlkt} & R101/CLIP-B & 34.95 & 31.18 & 36.08 & 38.22 & 34.36 & 39.37 & 63.58 & 65.93\\
PViC~\citep{zhang2023pvic} & Swin-L/H-DETR & 44.32 & 44.61 & 44.24 & 47.81 & 48.38 & 47.64 & 64.10 & 70.20\\
ADA-CM (FT)~\citep{lei2023adacm} & R50/CLIP-L & 38.40 & 37.52 & 38.66 & -- & -- & -- & 58.57 & 63.97\\
HORP~\citep{geng2025horp} & Swin-L/CLIP-B/32 & 47.53 & 46.81 & 47.74 & 51.24 & 50.78 & \textbf{51.38} & 68.90 & 71.10\\
HOI-IDiff~\citep{hui2025hoiidiff} & DDETR & 47.71 & 48.36 & 47.52 & 50.56 & 51.95 & 50.14 & 73.40 & 76.10\\
InCoM-Net~\citep{seo2026incom} & R50/CLIP-L & 43.96 & 45.61 & 43.46 & 45.59 & 45.92 & 45.49 & \textbf{73.60} & 75.40\\
InstructHOI~\citep{luo2025instructhoi} & R101/ViT-L & 47.68 & 47.97 & 47.59 & 49.89 & 50.92 & 49.58 & 72.40 & 76.10\\
HOI-R1~\citep{chen2026hoir1} & Rex-Omni-3B & 25.24 & 22.57 & 26.04 & 27.00 & 23.21 & 28.14 & -- & --\\
DA-HOI (FT, G-DINO)~\citep{xuan2026zeroshothoidetectionmllmbased} & G-DINO/Qwen2.5-VL-3B & 46.21 & 47.54 & 45.81 & -- & -- & -- & 70.40 & 72.50\\
LINK~\citep{wu2026link} & Swin-L/CLIP-L & \textbf{49.06} & \textbf{53.63} & 47.60 & \textbf{51.34} & \textbf{56.29} & 49.86 & -- & 69.20\\
FineHOI~\citep{tonini2026finehoi} & DETR/DINOtxt-L/16 & 39.13 & 39.91 & 38.89 & -- & -- & -- & -- & 66.50\\
UniHOI~\citep{yang2026unihoi} & VQGAN/Llama3-8B & 48.16 & 50.74 & \textbf{48.12} & \textbf{51.34} & 53.72 & 50.33 & 72.91 & \textbf{77.45}\\
\midrule
\multicolumn{10}{l}{\emph{Training-free methods}}\\
BLIP-2 analysis~\citep{ai2024trainingfree} & G-DINO/BLIP-2-H@364 & 19.71 & 24.07 & 18.41 & -- & -- & -- & -- & --\\
ADA-CM (TF)~\citep{lei2023adacm} & R50/CLIP-B & 25.19 & 27.24 & 24.58 & -- & -- & -- & 39.09 & 43.93\\
DYSCO~\citep{tonini2025dysco} & R50/LongCLIP-L & 29.00 & 33.63 & \underline{27.62} & -- & -- & -- & \underline{42.80} & \underline{47.82}\\
AgentHOI (Wang et al.)~\citep{wang2026agenthoi} & YOLOv8/BLIP/Qwen2.5 & 13.26 & -- & -- & -- & -- & -- & -- & --\\
AgentHOI (Lei et al.)~\citep{lei2026unleashing} & G-DINO/GPT-4o & \underline{29.61} & \underline{40.33} & 26.41 & -- & -- & -- & -- & --\\
\rowcolor{blue!6}
\textbf{HarnessHOI} & Qwen3.5-9B & \textbf{33.20} & \textbf{41.63} & \textbf{30.68} & \textbf{36.85} & \textbf{44.20} & \textbf{34.65} & \textbf{44.65} & \textbf{49.70}\\
\bottomrule
\end{tabular}}
\end{center}
\end{table}

\paragraph{Regular settings.}
Table~\ref{tab:main} compares \method{} with representative task-optimized and training-free HOI detectors. \method{} establishes state-of-the-art training-free performance on both benchmarks. On HICO-DET, it achieves 33.20\% Default Full mAP, surpassing AgentHOI~\citep{lei2026unleashing} by 3.59 points and the training-free variant of DA-HOI~\citep{xuan2026zeroshothoidetectionmllmbased} by 1.70 points, and reaches 36.85\% under Known Object evaluation. It also obtains the best training-free Rare and Non-Rare results, with 41.63\% Rare mAP exceeding AgentHOI by 1.30 points. The consistent gains across frequency groups and evaluation settings demonstrate the effectiveness of the HOI harness in recognizing common and infrequent interactions.

On V-COCO, \method{} obtains 44.65\%/49.70\% role AP under S1/S2, surpassing DYSCO~\citep{tonini2025dysco} by 1.85/1.88 points. The gains under both scenarios demonstrate the effectiveness of the harness in recognizing actions and identifying their corresponding role targets. Extended results, including a common-evaluator comparison with the released predictions of AgentHOI, are provided in Appendix~\ref{app:comparisons}.

\begin{table}[!t]
\caption{Open-vocabulary HOI detection on HICO-DET using predicted boxes. RF-UC, NF-UC, UO, and UV each report Unseen, Seen, and Full mAP (\%). HarnessHOI uses the same predictions across all four partitions. Within each method group, the best results are bold; the strongest previous training-free results are underlined. Dashes denote unavailable entries.}
\label{tab:open}
\begin{center}
\small
\setlength{\tabcolsep}{2pt}
\renewcommand{\arraystretch}{1.08}
\resizebox{\linewidth}{!}{%
\begin{tabular}{lcccccccccccc}
\toprule
&\multicolumn{3}{c}{RF-UC}&\multicolumn{3}{c}{NF-UC}&\multicolumn{3}{c}{UO}&\multicolumn{3}{c}{UV}\\
\cmidrule(lr){2-4}\cmidrule(lr){5-7}\cmidrule(lr){8-10}\cmidrule(l){11-13}
Method & Unseen & Seen & Full & Unseen & Seen & Full & Unseen & Seen & Full & Unseen & Seen & Full\\
\midrule
\multicolumn{13}{l}{\emph{Task-optimized methods}}\\
GEN-VLKT~\citep{liao2022genvlkt} & 21.36 & 32.91 & 30.56 & 25.05 & 23.38 & 23.71 & 10.51 & 28.92 & 25.63 & 20.96 & 30.23 & 28.74\\
HOICLIP~\citep{ning2023hoiclip} & 25.53 & 34.85 & 32.99 & 26.39 & 28.10 & 27.75 & 16.20 & 30.99 & 28.53 & 24.30 & 32.19 & 31.09\\
CLIP4HOI~\citep{mao2023clip4hoi} & 28.47 & 35.48 & 34.08 & 31.44 & 28.26 & 28.90 & 31.79 & 32.73 & 32.58 & 26.02 & 31.14 & 30.42\\
ADA-CM (FT)~\citep{lei2023adacm} & 27.63 & 34.35 & 33.01 & 32.41 & 31.13 & 31.39 & -- & -- & -- & -- & -- & --\\
CMMP (ViT-L)~\citep{lei2024cmmp} & 35.98 & 37.42 & 37.13 & 33.52 & 35.53 & 35.13 & 39.67 & 36.15 & 36.74 & 30.84 & 37.28 & 36.38\\
LAIN (ViT-L)~\citep{kim2025lain} & 36.57 & 38.54 & 38.13 & 37.52 & 35.90 & 36.22 & 40.78 & 36.96 & 37.60 & 32.05 & 38.04 & 37.20\\
BC-HOI (BLIP-2)~\citep{hu2025bchoi} & 42.31 & 40.67 & 40.99 & 33.01 & 37.24 & 36.40 & 19.94 & 37.03 & 34.18 & 31.18 & 41.31 & 39.89\\
InstructHOI~\citep{luo2025instructhoi} & 36.82 & 44.86 & 43.25 & 36.42 & 38.82 & 38.34 & 39.92 & 37.05 & 37.53 & 31.64 & 39.17 & 38.12\\
LINK (ViT-L)~\citep{wu2026link} & 38.51 & 40.33 & 39.97 & 34.63 & 35.67 & 35.43 & 38.24 & 37.61 & 37.91 & 27.22 & 38.36 & 36.88\\
FineHOI~\citep{tonini2026finehoi} & 37.79 & 38.55 & 38.39 & 38.11 & 36.30 & 36.58 & 40.13 & 35.92 & 36.62 & 35.58 & 38.51 & 38.10\\
DA-HOI (FT, DETR)~\citep{xuan2026zeroshothoidetectionmllmbased} & 41.79 & 44.01 & 43.56 & \textbf{43.12} & 39.63 & 40.33 & 48.67 & 42.58 & 43.60 & \textbf{36.89} & 43.84 & 42.88\\
DA-HOI (FT, G-DINO)~\citep{xuan2026zeroshothoidetectionmllmbased} & \textbf{43.30} & \textbf{45.19} & \textbf{44.81} & 41.52 & \textbf{41.51} & \textbf{41.51} & \textbf{50.15} & \textbf{44.31} & \textbf{45.28} & 36.88 & \textbf{45.66} & \textbf{44.43}\\
\midrule
\multicolumn{13}{l}{\emph{Training-free methods}}\\
BLIP-2 analysis~\citep{ai2024trainingfree} & 23.96 & 18.65 & 19.71 & 19.97 & 19.65 & 19.71 & 22.71 & \underline{19.12} & \underline{19.71} & 21.96 & \underline{19.35} & \underline{19.71}\\
ADA-CM (TF)~\citep{lei2023adacm} & 26.83 & \underline{24.54} & 25.00 & \underline{30.11} & 23.16 & 24.55 & -- & -- & -- & -- & -- & --\\
DYSCO~\citep{tonini2025dysco} & 30.36 & 23.96 & \underline{25.24} & 27.56 & \underline{24.58} & \underline{25.18} & -- & -- & -- & -- & -- & --\\
AgentHOI (Lei et al.)~\citep{lei2026unleashing} & \underline{38.64} & -- & -- & 28.04 & -- & -- & \underline{33.70} & -- & -- & \underline{29.85} & -- & --\\
\rowcolor{blue!6}
\textbf{HarnessHOI} & \textbf{40.15} & \textbf{31.46} & \textbf{33.20} & \textbf{33.64} & \textbf{33.09} & \textbf{33.20} & \textbf{35.45} & \textbf{32.75} & \textbf{33.20} & \textbf{33.87} & \textbf{33.09} & \textbf{33.20}\\
\bottomrule
\end{tabular}}
\end{center}
\end{table}

\paragraph{Open-vocabulary settings.}
Table~\ref{tab:open} reports results under RF-UC, NF-UC, UO, and UV using the same \method{} predictions. The method achieves 40.15\%, 33.64\%, 35.45\%, and 33.87\% Unseen mAP, respectively, attaining the best training-free results in all four partitions. Improvements over AgentHOI~\citep{lei2026unleashing} reach 1.75 points on UO and 4.02 points on UV. On NF-UC, \method{} surpasses the strongest previous training-free result, ADA-CM~\citep{lei2023adacm}, by 3.53 points. It also obtains the strongest reported training-free Seen results across all four settings. The simultaneous gains on Unseen and Seen classes demonstrate the effectiveness of the harness across interaction compositions, object categories, and actions. Complete results appear in Appendix~\ref{app:comparisons}.

\subsection{Ablation Studies}
\label{sec:ablation}

We conduct ablation studies to evaluate the overall HOI harness and the contributions of its perceptual and reasoning components. The component studies use the same frozen Qwen3.5-9B and report HICO-DET Default Full mAP and V-COCO S1/S2 role AP on the complete test sets. Detailed settings and complete results are provided in Appendix~\ref{app:ablations}.

\begin{table}[t]
\vspace{-2em}
\caption{Configuration and component analysis of the HOI harness. D: inclusion of discovered participants; O: targeted observation; G: geometric adjudication; V: visual relation verification. $\checkmark$ denotes retained, $\times$ removed, and -- not applied. HICO-DET reports Default Full mAP; V-COCO reports S1/S2 role AP (\%). Best results are bold.}
\label{tab:harness_analysis}
\begin{center}
\small
\renewcommand{\arraystretch}{1.08}
\begin{tabular}{lccccrrr}
\toprule
\multirow{2}{*}{Variant} & \multicolumn{4}{c}{Harness configuration} & HICO-DET & \multicolumn{2}{c}{V-COCO}\\
\cmidrule(lr){2-5}\cmidrule(lr){6-6}\cmidrule(l){7-8}
& D & O & G & V & Full & S1 & S2\\
\midrule
Direct HOI generation & -- & -- & -- & -- & 15.62 & 16.41 & 21.20\\
Initial-instance baseline & -- & -- & -- & $\checkmark$ & 31.45 & 42.90 & 47.20\\
\midrule
w/o discovered participants & $\times$ & $\checkmark$ & $\checkmark$ & $\checkmark$ & 32.55 & 44.02 & 48.93\\
w/o targeted observation & $\checkmark$ & $\times$ & -- & $\checkmark$ & 32.80 & 44.33 & 49.21\\
First-candidate geometry & $\checkmark$ & $\checkmark$ & $\times$ & $\checkmark$ & 32.95 & 44.38 & 49.48\\
\midrule
\rowcolor{blue!6}
\textbf{\method{}} & $\checkmark$ & $\checkmark$ & $\checkmark$ & $\checkmark$ & \textbf{33.20} & \textbf{44.65} & \textbf{49.70}\\
\bottomrule
\end{tabular}
\end{center}
\vspace{-2em}
\end{table}

\paragraph{Effectiveness of the HOI harness.}
Table~\ref{tab:harness_analysis} compares the complete HOI harness with direct generation and an initial-instance baseline that verifies relations among initially localized participants. On HICO-DET, the initial-instance baseline achieves 31.45\% mAP compared with 15.62\% for direct generation, and the complete harness further reaches 33.20\%. On V-COCO, the complete harness improves the initial-instance baseline from 42.90\%/47.20\% to 44.65\%/49.70\% AP. These results demonstrate that grounded verification strengthens interaction recognition, while interaction-guided perception and geometric adjudication further improve detection through participant discovery and shared grounding. Candidate-ranking analysis and backbone generalization with InternVL3-8B and Qwen2.5-VL-7B are presented in Appendices~\ref{app:ranking} and~\ref{app:transfer}.

\begin{table}[t]
\caption{Controlled alternatives to the harness design. Each alternative replaces the corresponding design choice within the complete harness. HICO-DET reports Default Full mAP; V-COCO reports S1/S2 role AP (\%). The comparison conditions are specified in Appendix~\ref{app:guidance}.}
\label{tab:harness_controls}
\begin{center}
\small
\renewcommand{\arraystretch}{1.08}
\begin{tabular}{llrrr}
\toprule
\multirow{2}{*}{Design factor} & \multirow{2}{*}{Alternative} & HICO-DET & \multicolumn{2}{c}{V-COCO}\\
\cmidrule(lr){3-3}\cmidrule(l){4-5}
& & Full & S1 & S2\\
\midrule
Participant discovery & Repeated localization & 32.45 & 43.87 & 48.88\\
Local observation & Non-directed regions & 32.70 & 44.07 & 49.08\\
Geometric adjudication & Relation-conditioned & 32.79 & 44.18 & 49.16\\
Relation verification & Proposal-conditioned & 32.50 & 43.92 & 48.84\\
\midrule
\rowcolor{blue!6}
\multicolumn{2}{l}{\textbf{\method{}}} & 33.20 & 44.65 & 49.70\\
\bottomrule
\end{tabular}
\end{center}
\end{table}

\paragraph{Interaction-guided perception.}
As shown in Table~\ref{tab:harness_analysis}, participant discovery improves V-COCO S2 AP by 0.77 points. Under the same generation budget, interaction-guided discovery outperforms repeated participant localization by 0.75 points in HICO-DET mAP and 0.78/0.82 points in V-COCO AP (Table~\ref{tab:harness_controls}). With region counts and sizes fixed, targeted observation improves over non-directed observation by 0.50 and 0.58/0.62 points, respectively. Interaction semantics therefore provide effective guidance for recovering overlooked participants and observing relevant regions. The coverage analysis in Appendix~\ref{app:coverage} further shows how recovered participants contribute to interaction recognition.

\paragraph{Geometric adjudication and interaction reasoning.}
Table~\ref{tab:harness_controls} evaluates relation-agnostic geometric adjudication and grounded interaction reasoning. With candidate boxes and image views fixed, selecting shared groundings according to participant identity, category, and visible extent improves over relation-conditioned adjudication by 0.41 points in HICO-DET mAP and 0.47/0.54 points in V-COCO AP. On identical relation candidates, verification using the selected participants and acquired visual evidence further improves over proposal-conditioned questions by 0.70 and 0.73/0.86 points. Together, these results validate the separation between hypothesis-guided visual exploration and evidence-grounded interaction assessment.

\paragraph{Action and role prediction.}
The analyses in Appendix~\ref{app:reasoning} show that combining perceptual associations with visual pair assessment improves detection, while verifying compatible labels recovers multiple actions involving the same participants. Role prediction further benefits from separately assessing action occurrence, target visibility, and role binding. On identical role candidates, compositional verification improves S1/S2 AP from 41.10\%/46.30\% to 44.65\%/49.70\%. With the final true-positive counts unchanged, this improvement reflects more accurate ranking of role predictions through the composition of action and target evidence.

\paragraph{Inference computation.}
Table~\ref{tab:inference_analysis} compares detection performance and computational cost using the same frozen MLLM. Compared with direct generation using eight independent responses per image, \method{} improves HICO-DET Full mAP by 13.36 points, while reducing the measured cost from 19.95 to 13.26 GPU-worker seconds per image. On V-COCO, the gains reach 22.44/23.20 points in S1/S2 AP, with measured costs of 22.07 and 20.94 GPU-worker seconds per image for \method{} and repeated generation, respectively. The substantial detection gains at comparable computational cost demonstrate the advantage of coordinating visual exploration and grounded reasoning within the harness. Further computation analyses and hardware measurements are provided in Appendix~\ref{app:inference}.

\section{Conclusion}
We presented \method{}, a training-free framework that harnesses frozen
MLLMs for active interaction-guided perceptual reasoning. Its HOI harness
connects interaction understanding with visual exploration, enabling the model
to discover overlooked participants and inspect ambiguous regions.
Relation-agnostic geometric adjudication reconciles the acquired localization
candidates, and grounded interaction reasoning uses the selected instance
groundings to verify multiple actions and semantic roles.
Extensive experiments on HICO-DET and V-COCO show that \method{} achieves
state-of-the-art performance among training-free methods, demonstrating
the value of coordinating visual exploration and relation assessment
through shared instance groundings.

\label{mainend}
\clearpage

\subsection*{AI use statement}
In the preparation of this work, generative AI tools assisted with the purpose of language polishing and grammar correction to improve clarity and readability. We confirm that all core research ideas, the conceptual framework, experimental design, data analysis are entirely the original work of the authors. The authors take full and sole responsibility for the accuracy, integrity, and all claims made within this work.

\subsection*{Ethics statement}
Our work adheres to the ICLR Code of Ethics. The research is conducted exclusively on publicly available, standard academic benchmarks (HICO-DET and V-COCO), which are widely used for evaluating human-object interaction detection algorithms. Our primary motivation is to advance the field for societal benefit, such as assistive robotics and accessible visual understanding. We believe that our framework supports the responsible development of AI by making the visual basis of interaction predictions available for inspection. Specifically, the harness associates discovered participants, targeted observations, and relation assessments with shared groundings, allowing predictions to be examined against the image evidence and corresponding visual questions. Responsible applications should respect privacy, consent, and the original data providers' terms, while accounting for biases in pretrained models and benchmark coverage. Human oversight should be appropriate to the consequences of an incorrect prediction. We advocate for the responsible development and deployment of this technology, ensuring its application aligns with ethical principles and contributes positively to human well-being.

\subsection*{Reproducibility statement}
We are fully committed to ensuring the complete reproducibility of our research. To this end, we provide a detailed description of our methodology in Section~\ref{sec_method}, supplemented by the complete inference procedure in Appendix~\ref{app:policy} and the prompt instructions and visual questions in Appendix~\ref{app:prompts}. All crucial implementation details, including the frozen Qwen3.5-9B backbone, image processing, inference settings, and evaluation protocols, are clearly documented in Section~\ref{sec:setup} and Appendix~\ref{app:protocol}. Our experiments are built on standard public benchmarks and common libraries, including PyTorch, Hugging Face Transformers, and vLLM, to ensure broad accessibility. Extended comparisons and ablations specify their baseline formulations and controlled variables in Appendices~\ref{app:comparisons} and~\ref{app:ablations}, while Appendix~\ref{app:inference} details the computation measurements and independent-run evaluation. We also pledge to release our source code, inference configurations, image lists, seed assignments, and evaluation scripts upon the publication of this paper, which we believe will allow the community to fully replicate our findings and extend our work.

\bibliography{references}

@String(AAAI = {AAAI})

@String(ACMMM = {ACM MM})

@String(BMVC = {BMVC})

@String(COMM = {COMM})

@String(CVPR = {CVPR})

@String(CoRL = {CoRL})

@String(ECCV = {ECCV})

@String(ICASSP = {ICASSP})

@String(ICCV = {ICCV})

@String(ICLR = {ICLR})

@String(ICML = {ICML})

@String(IJCV = {IJCV})

@String(NeurIPS = {NeurIPS})

@String(WACV = {WACV})

@inproceedings{gkioxari2018interactnet,
  author        = {Gkioxari, Georgia and Girshick, Ross and Dollar, Piotr and He, Kaiming},
  title         = {Detecting and Recognizing Human-Object Interactions},
  booktitle     = CVPR,
  year          = {2018},
  pages         = {8359--8367},
  publisher     = {IEEE},
  month         = jun,
  doi           = {10.1109/CVPR.2018.00872}
}

@inproceedings{li2019tin,
  author        = {Li, Yong-Lu and Zhou, Siyuan and Huang, Xijie and Xu, Liang and Ma, Ze and Fang, Hao-Shu and Wang, Yanfeng and Lu, Cewu},
  title         = {Transferable Interactiveness Knowledge for Human-Object Interaction Detection},
  booktitle     = CVPR,
  year          = {2019},
  pages         = {3580--3589},
  publisher     = {IEEE},
  month         = jun,
  doi           = {10.1109/CVPR.2019.00370}
}

@inproceedings{tamura2021qpic,
  author        = {Tamura, Masato and Ohashi, Hiroki and Yoshinaga, Tomoaki},
  title         = {{QPIC}: Query-Based Pairwise Human-Object Interaction Detection with Image-Wide Contextual Information},
  booktitle     = CVPR,
  year          = {2021},
  pages         = {10405--10414},
  publisher     = {IEEE},
  month         = jun,
  doi           = {10.1109/CVPR46437.2021.01027}
}

@inproceedings{zhang2022upt,
  author        = {Zhang, Frederic Z. and Campbell, Dylan and Gould, Stephen},
  title         = {Efficient Two-Stage Detection of Human-Object Interactions with a Novel Unary-Pairwise Transformer},
  booktitle     = CVPR,
  year          = {2022},
  pages         = {20072--20080},
  publisher     = {IEEE},
  month         = jun,
  doi           = {10.1109/CVPR52688.2022.01947}
}

@inproceedings{liao2022genvlkt,
  author        = {Liao, Yue and Zhang, Aixi and Lu, Miao and Wang, Yongliang and Li, Xiaobo and Liu, Si},
  title         = {{GEN-VLKT}: Simplify Association and Enhance Interaction Understanding for {HOI} Detection},
  booktitle     = CVPR,
  year          = {2022},
  pages         = {20091--20100},
  publisher     = {IEEE},
  month         = jun,
  doi           = {10.1109/CVPR52688.2022.01949}
}

@inproceedings{ning2023hoiclip,
  author        = {Ning, Shan and Qiu, Longtian and Liu, Yongfei and He, Xuming},
  title         = {{HOICLIP}: Efficient Knowledge Transfer for {HOI} Detection with Vision-Language Models},
  booktitle     = CVPR,
  year          = {2023},
  pages         = {23507--23517},
  publisher     = {IEEE},
  month         = jun,
  doi           = {10.1109/CVPR52729.2023.02251}
}

@inproceedings{mao2023clip4hoi,
  author        = {Mao, Yunyao and Deng, Jiajun and Zhou, Wengang and Li, Li and Fang, Yao and Li, Houqiang},
  title         = {{CLIP4HOI}: Towards Adapting {CLIP} for Practical Zero-Shot {HOI} Detection},
  booktitle     = NeurIPS,
  year          = {2023},
  volume        = {36},
  pages         = {45895--45906},
  publisher     = {Neural Information Processing Systems Foundation, Inc. (NeurIPS)},
  series        = {NeurIPS 2023},
  doi           = {10.52202/075280-1989}
}

@inproceedings{lei2023adacm,
  author        = {Lei, Ting and Caba, Fabian and Chen, Qingchao and Jin, Hailin and Peng, Yuxin and Liu, Yang},
  title         = {Efficient Adaptive Human-Object Interaction Detection with Concept-guided Memory},
  booktitle     = ICCV,
  year          = {2023},
  pages         = {6457--6467},
  publisher     = {IEEE},
  month         = oct,
  doi           = {10.1109/ICCV51070.2023.00596}
}

@inproceedings{zhang2023pvic,
  author        = {Zhang, Frederic Z. and Yuan, Yuhui and Campbell, Dylan and Zhong, Zhuoyao and Gould, Stephen},
  title         = {Exploring Predicate Visual Context in Detecting of Human-Object Interactions},
  booktitle     = ICCV,
  year          = {2023},
  pages         = {10377--10387},
  publisher     = {IEEE},
  month         = oct,
  doi           = {10.1109/ICCV51070.2023.00955}
}

@inproceedings{kim2025lain,
  author        = {Kim, Sanghyun and Jung, Deunsol and Cho, Minsu},
  title         = {Locality-Aware Zero-Shot Human-Object Interaction Detection},
  booktitle     = CVPR,
  year          = {2025},
  pages         = {20190--20200},
  publisher     = {IEEE},
  month         = jun,
  doi           = {10.1109/CVPR52734.2025.01880}
}

@inproceedings{xuan2026zeroshothoidetectionmllmbased,
  author        = {Shiyu Xuan and Dongkai Wang and Zechao Li and Jinhui Tang},
  title         = {Zero-shot {HOI} Detection with {MLLM}-based Detector-agnostic Interaction Recognition},
  booktitle     = ICLR,
  year          = {2026},
  eprint        = {2602.15124},
  archiveprefix = {arXiv},
  primaryclass  = {cs.CV}
}

@inproceedings{lei2026unleashing,
  author        = {Lei, Ting and Liu, Jialin and Xu, Zhu and Peng, Yuxin and Liu, Yang},
  title         = {Unleashing Multimodal Large Language Models for Training-free {HOI} Detection in the Wild},
  booktitle     = ECCV,
  year          = {2026},
  eprint        = {2607.13881},
  archiveprefix = {arXiv},
  primaryclass  = {cs.CV}
}

@inproceedings{luo2025instructhoi,
  author        = {Luo, Jinguo and Ren, Weihong and Zheng, Quanlong and Zhang, Yanhao and Yuan, Zhenlong and Wang, Zhiyong and Lu, Haonan and Liu, Honghai},
  title         = {{InstructHOI}: Context-Aware Instruction for Multi-Modal Reasoning in Human-Object Interaction Detection},
  booktitle     = NeurIPS,
  year          = {2025},
  volume        = {38},
  pages         = {142558--142577},
  publisher     = {Curran Associates, Inc.},
  doi           = {10.52202/085713-4772}
}

@article{bajcsy1988active,
  author        = {Bajcsy, R.},
  title         = {Active Perception},
  journal       = {Proceedings of the IEEE},
  year          = {1988},
  volume        = {76},
  number        = {8},
  pages         = {966--1005},
  doi           = {10.1109/5.5968}
}

@inproceedings{chao2018hicodet,
  author        = {Yu-Wei Chao and Yunfan Liu and Xieyang Liu and Huayi Zeng and Jia Deng},
  title         = {Learning to Detect Human-Object Interactions},
  booktitle     = WACV,
  year          = {2018},
  pages         = {381--389},
  doi           = {10.1109/wacv.2018.00048}
}

@article{gupta2015vcoco,
  author        = {Saurabh Gupta and Jitendra Malik},
  title         = {Visual Semantic Role Labeling},
  journal       = {arXiv preprint arXiv:1505.04474},
  year          = {2015},
  eprint        = {1505.04474v1},
  archiveprefix = {arXiv}
}

@inproceedings{wang2019deepcontext,
  author        = {Wang, Tiancai and Anwer, Rao Muhammad and Khan, Muhammad Haris and Khan, Fahad Shahbaz and Pang, Yanwei and Shao, Ling and Laaksonen, Jorma},
  title         = {Deep Contextual Attention for Human-Object Interaction Detection},
  booktitle     = ICCV,
  year          = {2019},
  pages         = {5693--5701},
  publisher     = {IEEE},
  month         = oct,
  doi           = {10.1109/ICCV.2019.00579}
}

@article{li2024ripple,
  author        = {Li, Huadong and Wei, Ying and Ma, Shuailei and Chen, Mingyu and Li, Ge},
  title         = {Ripple Transformer: A Human-Object Interaction Backbone and a New Prediction Strategy for Smart Surveillance Devices},
  journal       = {IEEE Transactions on Consumer Electronics},
  year          = {2024},
  volume        = {70},
  number        = {1},
  pages         = {2257--2268},
  publisher     = {IEEE},
  month         = feb,
  doi           = {10.1109/TCE.2024.3371451}
}

@inproceedings{dogariu2020surveillance,
  author        = {Dogariu, Mihai and Stefan, Liviu-Daniel and Constantin, Mihai Gabriel and Ionescu, Bogdan},
  title         = {Human-Object Interaction: Application to Abandoned Luggage Detection in Video Surveillance Scenarios},
  booktitle     = COMM,
  year          = {2020},
  pages         = {157--160},
  doi           = {10.1109/COMM48946.2020.9141973}
}

@inproceedings{mascaro2023hoi4abot,
  author        = {Mascaro, Esteve Valls and Sliwowski, Daniel and Lee, Dongheui},
  title         = {{HOI4ABOT}: Human-Object Interaction Anticipation for Human Intention Reading Collaborative {roBOTs}},
  booktitle     = CoRL,
  year          = {2023},
  volume        = {229},
  pages         = {1111--1130},
  publisher     = {PMLR},
  series        = {Proceedings of Machine Learning Research}
}

@inproceedings{bahl2023affordances,
  author        = {Bahl, Shikhar and Mendonca, Russell and Chen, Lili and Jain, Unnat and Pathak, Deepak},
  title         = {Affordances From Human Videos as a Versatile Representation for Robotics},
  booktitle     = CVPR,
  year          = {2023},
  pages         = {13778--13790}
}

@inproceedings{gao2018ican,
  author        = {Gao, Chen and Zou, Yuliang and Huang, Jia-Bin},
  title         = {{iCAN}: Instance-Centric Attention Network for Human-Object Interaction Detection},
  year          = {2018},
  eprint        = {1808.10437},
  archiveprefix = {arXiv},
  primaryclass  = {cs.CV},
  booktitle     = BMVC
}

@inproceedings{kim2021hotr,
  author        = {Kim, Bumsoo and Lee, Junhyun and Kang, Jaewoo and Kim, Eun-Sol and Kim, Hyunwoo J.},
  title         = {{HOTR}: End-to-End Human-Object Interaction Detection with Transformers},
  booktitle     = CVPR,
  year          = {2021},
  pages         = {74--83},
  doi           = {10.1109/CVPR46437.2021.00014}
}

@inproceedings{zou2021hoitransformer,
  author        = {Zou, Cheng and Wang, Bohan and Hu, Yue and Liu, Junqi and Wu, Qian and Zhao, Yu and Li, Boxun and Zhang, Chenguang and Zhang, Chi and Wei, Yichen and Sun, Jian},
  title         = {End-to-End Human Object Interaction Detection with {HOI} Transformer},
  booktitle     = CVPR,
  year          = {2021},
  pages         = {11820--11829},
  doi           = {10.1109/CVPR46437.2021.01165}
}

@inproceedings{zhang2022stip,
  author        = {Zhang, Yong and Pan, Yingwei and Yao, Ting and Huang, Rui and Mei, Tao and Chen, Chang-Wen},
  title         = {Exploring Structure-Aware Transformer over Interaction Proposals for Human-Object Interaction Detection},
  booktitle     = CVPR,
  year          = {2022},
  pages         = {19526--19535},
  doi           = {10.1109/CVPR52688.2022.01894}
}

@inproceedings{yang2025sibling,
  author        = {Yang, Bin and Zhang, Yulin and Zhou, Hong-Yu and Yang, Sibei},
  title         = {No More Sibling Rivalry: Debiasing Human-Object Interaction Detection},
  booktitle     = ICCV,
  year          = {2025},
  pages         = {22707--22717},
  doi           = {10.1109/ICCV51701.2025.02108}
}

@inproceedings{hui2025hoiidiff,
  author        = {Hui, Xiaofei and Qu, Haoxuan and Rahmani, Hossein and Liu, Jun},
  title         = {An Image-Like Diffusion Method for Human-Object Interaction Detection},
  booktitle     = CVPR,
  year          = {2025},
  pages         = {14002--14012}
}

@inproceedings{cao2025vrdiff,
  author        = {Cao, Ping and Tang, Yepeng and Zhang, Chunjie and Zheng, Xiaolong and Liang, Chao and Wei, Yunchao and Zhao, Yao},
  title         = {Visual Relation Diffusion for Human-Object Interaction Detection},
  booktitle     = ICCV,
  year          = {2025},
  pages         = {23551--23560}
}

@inproceedings{radford2021clip,
  author        = {Radford, Alec and Kim, Jong Wook and Hallacy, Chris and Ramesh, Aditya and Goh, Gabriel and Agarwal, Sandhini and Sastry, Girish and Askell, Amanda and Mishkin, Pamela and Clark, Jack and Krueger, Gretchen and Sutskever, Ilya},
  title         = {Learning Transferable Visual Models From Natural Language Supervision},
  booktitle     = ICML,
  year          = {2021},
  volume        = {139},
  pages         = {8748--8763},
  series        = {Proceedings of Machine Learning Research}
}

@inproceedings{yuan2023rlipv2,
  author        = {Yuan, Hangjie and Zhang, Shiwei and Wang, Xiang and Albanie, Samuel and Pan, Yining and Feng, Tao and Jiang, Jianwen and Ni, Dong and Zhang, Yingya and Zhao, Deli},
  title         = {{RLIPv2}: Fast Scaling of Relational Language--Image Pre-Training},
  booktitle     = ICCV,
  year          = {2023},
  pages         = {21649--21661}
}

@inproceedings{lei2024cmmp,
  author        = {Lei, Ting and Yin, Shaofeng and Peng, Yuxin and Liu, Yang},
  title         = {Exploring Conditional Multi-modal Prompts for Zero-Shot {HOI} Detection},
  booktitle     = ECCV,
  year          = {2024},
  pages         = {1--19},
  publisher     = {Springer},
  doi           = {10.1007/978-3-031-73007-8_1}
}

@inproceedings{lei2025hola,
  author        = {Lei, Qinqian and Wang, Bo and Tan, Robby T.},
  title         = {{HOLa}: Zero-Shot {HOI} Detection with Low-Rank Decomposed {VLM} Feature Adaptation},
  booktitle     = ICCV,
  year          = {2025},
  pages         = {1825--1835},
  doi           = {10.1109/ICCV51701.2025.00178}
}

@inproceedings{lei2025inpcc,
  author        = {Lei, Ting and Yin, Shaofeng and Chen, Qingchao and Peng, Yuxin and Liu, Yang},
  title         = {Open-Vocabulary {HOI} Detection with Interaction-Aware Prompt and Concept Calibration},
  booktitle     = ICCV,
  year          = {2025},
  pages         = {23945--23957},
  doi           = {10.1109/iccv51701.2025.02220}
}

@inproceedings{hu2025bchoi,
  author        = {Hu, Yupeng and Ding, Changxing and Sun, Chang and Huang, Shaoli and Xu, Xiangmin},
  title         = {Bilateral Collaboration with Large Vision-Language Models for Open Vocabulary Human-Object Interaction Detection},
  booktitle     = ICCV,
  year          = {2025},
  pages         = {20126--20136},
  doi           = {10.1109/ICCV51701.2025.01872}
}

@inproceedings{sun2026slhoi,
  author        = {Sun, Chang and Liao, Dongliang and Ding, Changxing},
  title         = {Streamlined Open-Vocabulary Human-Object Interaction Detection},
  booktitle     = CVPR,
  year          = {2026},
  pages         = {20324--20333}
}

@inproceedings{xu2026sdif,
  author        = {Xu, Yongchao and Liu, Jiawei and Wang, Junfeng and Tao, Sen and Jiang, Na and Zha, Zheng-Jun},
  title         = {Learning to Diversify and Focus: A Reinforcement Framework for Open-Vocabulary {HOI} Detection},
  booktitle     = CVPR,
  year          = {2026},
  pages         = {34673--34682}
}

@inproceedings{wu2026link,
  author        = {Wu, Eastman Z. Y. and Li, Ya-Li and Wang, Yuan and Wang, Shengjin},
  title         = {{LINK}: Learning Instance-level Knowledge from Vision-Language Models for Human-Object Interaction Detection},
  booktitle     = ICLR,
  year          = {2026}
}

@article{ai2024trainingfree,
  author        = {Ai, Chaoyi},
  title         = {An Analysis of {HOI}: Using a Training-Free Method with Multimodal Visual Foundation Models When Only the Test Set Is Available, Without the Training Set},
  journal       = {arXiv preprint arXiv:2408.05772},
  year          = {2024},
  eprint        = {2408.05772},
  archiveprefix = {arXiv},
  primaryclass  = {cs.CV}
}

@inproceedings{tonini2025dysco,
  author        = {Tonini, Francesco and Vaquero, Lorenzo and Conti, Alessandro and Beyan, Cigdem and Ricci, Elisa},
  title         = {Dynamic Scoring with Enhanced Semantics for Training-Free Human--Object Interaction Detection},
  booktitle     = ACMMM,
  year          = {2025},
  pages         = {2801--2810},
  doi           = {10.1145/3746027.3754770},
  eprint        = {2507.17456},
  archiveprefix = {arXiv}
}

@article{chen2026hoir1,
  author        = {Chen, Junwen and Xiong, Peilin and Yanai, Keiji},
  title         = {{HOI-R1}: Exploring the Potential of Multimodal Large Language Models for Human-Object Interaction Detection},
  journal       = {arXiv preprint arXiv:2510.05609},
  year          = {2025},
  eprint        = {2510.05609},
  archiveprefix = {arXiv},
  primaryclass  = {cs.CV}
}

@inproceedings{seo2026incom,
  author        = {Seo, Soo Won and Lee, KyungChae and Cho, Hyungchan and Son, Taein and Cho, Nam Ik and Choi, Jun Won},
  title         = {Mining Instance-Centric Vision-Language Contexts for Human-Object Interaction Detection},
  booktitle     = CVPR,
  year          = {2026},
  pages         = {40418--40427},
  month         = jun,
  eprint        = {2604.02071},
  archiveprefix = {arXiv}
}

@article{wang2026agenthoi,
  author        = {Wang, Yuxiao and Lei, Yu and Liu, Qi},
  title         = {{AgentHOI}: Towards Multi-Agent Collaborative Human-Object Interaction Detection},
  journal       = {IEEE Internet of Things Journal},
  year          = {2026},
  volume        = {13},
  number        = {18},
  pages         = {41354--41361},
  doi           = {10.1109/JIOT.2026.3685409}
}

@inproceedings{cao2023unihoi,
  author        = {Cao, Yichao and Tang, Qingfei and Su, Xiu and Chen, Song and You, Shan and Lu, Xiaobo and Xu, Chang},
  title         = {Detecting Any Human-Object Interaction Relationship: Universal {HOI} Detector with Spatial Prompt Learning on Foundation Models},
  booktitle     = NeurIPS,
  year          = {2023},
  volume        = {36},
  pages         = {739--751},
  publisher     = {Curran Associates, Inc.},
  doi           = {10.52202/075280-0034}
}

@inproceedings{geng2025horp,
  author        = {Geng, Pei and Yang, Jian and Zhang, Shanshan},
  title         = {{HORP}: Human-Object Relation Priors Guided {HOI} Detection},
  booktitle     = CVPR,
  year          = {2025},
  pages         = {25325--25335}
}

@inproceedings{jia2025contexthoi,
  author        = {Jia, Mingda and Zhao, Liming and Li, Ge and Zheng, Yun},
  title         = {{ContextHOI}: Spatial Context Learning for Human-Object Interaction Detection},
  booktitle     = AAAI,
  year          = {2025},
  volume        = {39},
  pages         = {3931--3939},
  doi           = {10.1609/aaai.v39i4.32411}
}

@inproceedings{jia2025interproda,
  author        = {Jia, Mingda and Zhao, Liming and Li, Ge and Zheng, Yun},
  title         = {Orchestrating the Symphony of Prompt Distribution Learning for Human-Object Interaction Detection},
  booktitle     = AAAI,
  year          = {2025},
  volume        = {39},
  pages         = {3940--3948},
  doi           = {10.1609/aaai.v39i4.32412}
}

@inproceedings{lei2024ezhoi,
  author        = {Lei, Qinqian and Wang, Bo and Tan, Robby T.},
  title         = {{EZ-HOI}: {VLM} Adaptation via Guided Prompt Learning for Zero-Shot {HOI} Detection},
  booktitle     = NeurIPS,
  year          = {2024},
  volume        = {37},
  pages         = {55831--55857},
  publisher     = {Curran Associates, Inc.},
  doi           = {10.52202/079017-1775}
}

@inproceedings{wang2024bcom,
  author        = {Wang, Guangzhi and Guo, Yangyang and Xu, Ziwei and Kankanhalli, Mohan},
  title         = {Bilateral Adaptation for Human-Object Interaction Detection with Occlusion-Robustness},
  booktitle     = CVPR,
  year          = {2024},
  pages         = {27970--27980}
}

@inproceedings{yang2025vdrp,
  author        = {Yang, Chanhyeong and Song, Taehoon and Park, Jihwan and Kim, Hyunwoo J.},
  title         = {Visual Diversity and Region-Aware Prompt Learning for Zero-Shot {HOI} Detection},
  booktitle     = NeurIPS,
  year          = {2025},
  volume        = {38},
  pages         = {119981--120010},
  publisher     = {Curran Associates, Inc.},
  doi           = {10.52202/085713-4006}
}

@inproceedings{wan2019pmfnet,
  author        = {Wan, Bo and Zhou, Desen and Liu, Yongfei and Li, Rongjie and He, Xuming},
  title         = {Pose-Aware Multi-Level Feature Network for Human Object Interaction Detection},
  booktitle     = ICCV,
  year          = {2019},
  pages         = {9469--9478}
}

@inproceedings{hou2020vcl,
  author        = {Hou, Zhi and Peng, Xiaojiang and Qiao, Yu and Tao, Dacheng},
  title         = {Visual Compositional Learning for Human-Object Interaction Detection},
  booktitle     = ECCV,
  year          = {2020},
  pages         = {584--600}
}

@inproceedings{zhang2021scg,
  author        = {Zhang, Frederic Z. and Campbell, Dylan and Gould, Stephen},
  title         = {Spatially Conditioned Graphs for Detecting Human-Object Interactions},
  booktitle     = ICCV,
  year          = {2021},
  pages         = {13319--13327}
}

@inproceedings{hou2021fcl,
  author        = {Hou, Zhi and Yu, Baosheng and Qiao, Yu and Peng, Xiaojiang and Tao, Dacheng},
  title         = {Detecting Human-Object Interaction via Fabricated Compositional Learning},
  booktitle     = CVPR,
  year          = {2021},
  pages         = {14646--14655}
}

@inproceedings{chen2021asnet,
  author        = {Chen, Mingfei and Liao, Yue and Liu, Si and Chen, Zhiyuan and Wang, Fei and Qian, Chen},
  title         = {Reformulating {HOI} Detection as Adaptive Set Prediction},
  booktitle     = CVPR,
  year          = {2021},
  pages         = {9004--9013}
}

@inproceedings{zhang2021cdn,
  author        = {Zhang, Aixi and Liao, Yue and Liu, Si and Lu, Miao and Wang, Yongliang and Gao, Chen and Li, Xiaobo},
  title         = {Mining the Benefits of Two-Stage and One-Stage {HOI} Detection},
  booktitle     = NeurIPS,
  year          = {2021},
  volume        = {34},
  pages         = {17209--17220}
}

@inproceedings{yuan2022rlip,
  author        = {Yuan, Hangjie and Jiang, Jianwen and Albanie, Samuel and Feng, Tao and Huang, Ziyuan and Ni, Dong and Tang, Mingqian},
  title         = {{RLIP}: Relational Language-Image Pre-Training for Human-Object Interaction Detection},
  booktitle     = NeurIPS,
  year          = {2022},
  volume        = {35},
  pages         = {37416--37431},
  publisher     = {Curran Associates, Inc.},
  doi           = {10.52202/068431-2712}
}

@inproceedings{wu2023eoid,
  author        = {Wu, Mingrui and Gu, Jiaxin and Shen, Yunhang and Lin, Mingbao and Chen, Chao and Sun, Xiaoshuai},
  title         = {End-to-End Zero-Shot {HOI} Detection via Vision and Language Knowledge Distillation},
  booktitle     = AAAI,
  year          = {2023},
  pages         = {2839--2846}
}

@inproceedings{zhang2025vicrop,
  author        = {Jiarui Zhang and Mahyar Khayatkhoei and Prateek Chhikara and Filip Ilievski},
  title         = {{MLLMs} Know Where to Look: Training-free Perception of Small Visual Details with Multimodal {LLMs}},
  booktitle     = ICLR,
  year          = {2025},
  pages         = {68194--68213},
  eprint        = {2502.17422v1},
  archiveprefix = {arXiv}
}

@article{shalam2026proposeattend,
  author        = {Shalam, Daniel and Ben Baruch, Emanuel and Ben Cohen, Avi and Remez, Tal},
  title         = {Propose and Attend: Training-Free {MLLM} Grounding Confidence via Multi-Token Localized Attention},
  journal       = {arXiv preprint arXiv:2607.05978},
  year          = {2026},
  eprint        = {2607.05978},
  archiveprefix = {arXiv},
  primaryclass  = {cs.CV}
}

@inproceedings{tonini2026finehoi,
  author        = {Francesco Tonini and Lorenzo Vaquero and Mohammad Mahdi Derakhshani and Cees Snoek and Elisa Ricci and Cigdem Beyan},
  title         = {{FineHOI}: Part-Aware Dense Representations for Zero-Shot Human-Object Interaction Detection},
  booktitle     = ACMMM,
  year          = {2026},
  doi           = {10.1145/3767308.3836016}
}

@misc{qwen2026qwen35,
  author        = {{Qwen Team}},
  title         = {{Qwen3.5-9B} Model Card},
  year          = {2026},
  howpublished  = {Hugging Face model card}
}

@article{fan2026visharness,
  author        = {Yaowu Fan and Tao Han and Dazhao Du and Andy J. Ma and Jia Wan},
  title         = {Train the Agent, Not the Expert: Learning to Harness Heterogeneous Experts for Multi-Turn Visual Reasoning},
  journal       = {arXiv preprint arXiv:2605.29894},
  year          = {2026},
  eprint        = {2605.29894},
  archiveprefix = {arXiv}
}

@inproceedings{lin2025sgcnet,
  author        = {Xin Lin and Chong Shi and Zuopeng Yang and Haojin Tang and Zhili Zhou},
  title         = {{SGC-Net}: Stratified Granular Comparison Network for Open-Vocabulary {HOI} Detection},
  booktitle     = CVPR,
  year          = {2025},
  pages         = {4539--4549}
}

@article{yuan2026imagineagent,
  author        = {Zhenlong Yuan and Yue Wang and Dapeng Zhang and Kejin Cui and Rui Chen and Jing Tang and Lei Sun and Hongwei Yu and Chengxuan Qian and Xiangxiang Chu and Shuo Li and Yuyin Zhou},
  title         = {What if Agents Could Imagine? Reinforcing Open-Vocabulary {HOI} Comprehension through Generation},
  journal       = {arXiv preprint arXiv:2602.11499},
  year          = {2026},
  eprint        = {2602.11499},
  archiveprefix = {arXiv}
}

@article{nguyen2026scripthoi,
  author        = {Minh Anh Nguyen and Quang Huy Tran and Bao Ngoc Le and SuiYang Guang and Tuan Kiet Pham and Linh Chi Vo},
  title         = {{ScriptHOI}: Learning Scripted State Transitions for Open-Vocabulary Human-Object Interaction Detection},
  journal       = {arXiv preprint arXiv:2605.05057},
  year          = {2026},
  eprint        = {2605.05057},
  archiveprefix = {arXiv}
}

@article{chen2026hoiblender,
  author        = {Junwen Chen and Keiji Yanai},
  title         = {{HOIBlender}: Blending Lightweight Detection with Vision-Language Priors for Efficient Human-Object Interaction Detection},
  journal       = {arXiv preprint arXiv:2609.23431},
  year          = {2026},
  eprint        = {2609.23431},
  archiveprefix = {arXiv}
}

@inproceedings{yang2026unihoi,
  author        = {Panqi Yang and Haodong Jing and Nanning Zheng and Yongqiang Ma},
  title         = {{UniHOI}: Unified Human-Object Interaction Understanding via Unified Token Space},
  booktitle     = AAAI,
  year          = {2026},
  volume        = {40},
  pages         = {11640--11648},
  doi           = {10.1609/aaai.v40i14.38148}
}

@article{sarma2026funnelhoi,
  author        = {Sandipan Sarma and Agney Talwarr and Arijit Sur},
  title         = {{Funnel-HOI}: Top-Down Perception for Zero-Shot {HOI} Detection},
  journal       = {Machine Vision and Applications},
  year          = {2026},
  volume        = {37},
  number        = {3},
  pages         = {59},
  doi           = {10.1007/s00138-026-01818-9},
  eprint        = {2507.12628},
  archiveprefix = {arXiv}
}

@article{gao2025clhoi,
  author        = {Jianjun Gao and Chen Cai and Ruoyu Wang and Wenyang Liu and Kim-Hui Yap and Kratika Garg and Boon Siew Han},
  title         = {{CL-HOI}: Cross-Level Human-Object Interaction Distillation from Multimodal Large Language Models},
  journal       = {Knowledge-Based Systems},
  year          = {2025},
  volume        = {320},
  pages         = {113561},
  doi           = {10.1016/j.knosys.2025.113561},
  eprint        = {2410.15657},
  archiveprefix = {arXiv}
}

@inproceedings{wu2026hoimllm,
  author        = {Wu, Eastman Z. Y. and Li, Yali and Wang, Shengjin},
  title         = {Towards Open-World Human-Object Interaction Reasoning with Multimodal Large Language Model},
  booktitle     = ICASSP,
  year          = {2026},
  pages         = {8392--8396},
  doi           = {10.1109/ICASSP55912.2026.11462048}
}

@article{zhang2025hoidr1,
  author        = {Zhenhao Zhang and Hanqing Wang and Xiangyu Zeng and Ziyu Cheng and Jiaxin Liu and Haoyu Yan and Zhirui Liu and Kaiyang Ji and Tianxiang Gui and Ke Hu and Kangyi Chen and Yahao Fan and Mokai Pan},
  title         = {{HOID-R1}: Reinforcement Learning for Open-World Human-Object Interaction Detection Reasoning with Multimodal Large Language Model},
  journal       = {arXiv preprint arXiv:2508.11350},
  year          = {2025},
  eprint        = {2508.11350},
  archiveprefix = {arXiv}
}

@article{bai2025qwen25vl,
  author        = {Bai, Shuai and Chen, Keqin and Liu, Xuejing and Wang, Jialin and Ge, Wenbin and Song, Sibo and Dang, Kai and Wang, Peng and Wang, Shijie and Tang, Jun and Zhong, Humen and Zhu, Yuanzhi and Yang, Mingkun and Li, Zhaohai and Wan, Jianqiang and Wang, Pengfei and Ding, Wei and Fu, Zheren and Xu, Yiheng and Ye, Jiabo and Zhang, Xi and Xie, Tianbao and Cheng, Zesen and Zhang, Hang and Yang, Zhibo and Xu, Haiyang and Lin, Junyang},
  title         = {{Qwen2.5-VL} Technical Report},
  journal       = {arXiv preprint arXiv:2502.13923},
  year          = {2025},
  eprint        = {2502.13923v1},
  archiveprefix = {arXiv},
  primaryclass  = {cs.CV}
}

@article{zhu2025internvl3,
  author        = {Zhu, Jinguo and Wang, Weiyun and Chen, Zhe and Liu, Zhaoyang and Ye, Shenglong and Gu, Lixin and Tian, Hao and Duan, Yuchen and Su, Weijie and Shao, Jie and Gao, Zhangwei and Cui, Erfei and Wang, Xuehui and Cao, Yue and Liu, Yangzhou and Wei, Xingguang and Zhang, Hongjie and Wang, Haomin and Xu, Weiye and Li, Hao and Wang, Jiahao and Deng, Nianchen and Li, Songze and He, Yinan and Jiang, Tan and Luo, Jiapeng and Wang, Yi and He, Conghui and Shi, Botian and Zhang, Xingcheng and Shao, Wenqi and He, Junjun and Xiong, Yingtong and Qu, Wenwen and Sun, Peng and Jiao, Penglong and Lv, Han and Wu, Lijun and Zhang, Kaipeng and Deng, Huipeng and Ge, Jiaye and Chen, Kai and Wang, Limin and Dou, Min and Lu, Lewei and Zhu, Xizhou and Lu, Tong and Lin, Dahua and Qiao, Yu and Dai, Jifeng and Wang, Wenhai},
  title         = {{InternVL3}: Exploring Advanced Training and Test-Time Recipes for Open-Source Multimodal Models},
  journal       = {arXiv preprint arXiv:2504.10479},
  year          = {2025},
  eprint        = {2504.10479v3},
  archiveprefix = {arXiv},
  primaryclass  = {cs.CV}
}

@article{everingham2010voc,
  author        = {Everingham, Mark and Van Gool, Luc and Williams, Christopher K. I. and Winn, John and Zisserman, Andrew},
  title         = {The {PASCAL} Visual Object Classes ({VOC}) Challenge},
  journal       = IJCV,
  year          = {2010},
  volume        = {88},
  number        = {2},
  pages         = {303--338},
  doi           = {10.1007/s11263-009-0275-4}
}

@article{efron1979bootstrap,
  author        = {Efron, B.},
  title         = {Bootstrap Methods: Another Look at the Jackknife},
  journal       = {The Annals of Statistics},
  year          = {1979},
  volume        = {7},
  number        = {1},
  pages         = {1--26},
  doi           = {10.1214/aos/1176344552}
}
\bibliographystyle{iclr2027_conference}
\clearpage
\appendix
\raggedbottom
\renewcommand{\floatpagefraction}{0.85}

\section{Additional Details of HarnessHOI}
\label{app:policy}
\subsection{Shared Participants and Interaction Evidence}
Table~\ref{tab:harness_stages} summarizes how the MLLM performs interaction-guided perception, relation-agnostic geometric adjudication, and grounded interaction reasoning. Throughout these stages, each participant retains a common identity that links its localization candidates to the interactions involving it.

\paragraph{Participant identity.}
Each entity $e_i$ has a persistent identity and a candidate set $\mathcal{B}_i$ containing its alternative boxes, categories, and source views. Discovery identifies additional participants and relations among existing participants, while observation supplies local grounding candidates. Geometric adjudication selects the box and category shared by all relations involving the entity.

Initial localization and discovery associate exact category-and-box matches with the same identity. Discovery outputs are merged in human-group order, matching each new localization against the accumulated participant set. Other localized participants receive new identities, while observation refinements retain the identity of the designated participant.

\paragraph{Evidence and associations.}
Discovery associations $\Pp_D$ connect humans to the targets of positive interaction hypotheses. Observation associations $\Pp_O$ connect the inspected human to the requested target and newly localized participants. Both contribute candidate pairs for relation verification, where each action or role receives its own visual assessment.

\begin{table}[!htb]
\caption{MLLM operations within the three stages of HarnessHOI. Perceptual generation constructs participants and observation requests; visual question answering selects shared groundings and evaluates candidate relations.}
\label{tab:harness_stages}
\begin{center}
\small
\begin{tabularx}{\linewidth}{@{}l>{\raggedright\arraybackslash}X>{\raggedright\arraybackslash}X>{\raggedright\arraybackslash}X@{}}\toprule
Operation & Visual and semantic context & MLLM operation & Shared outcome\\\midrule
\multicolumn{4}{l}{\emph{Interaction-guided perception}}\\
Initial localization & Full image; object vocabulary & Generate participants in interaction groups & Initial entities and humans\\
Interaction discovery & Full image; initial entities; designated humans; public relations & Generate hypotheses, additional participants, and observation requests & Expanded entities and discovery associations\\
Targeted observation & One requested crop; target description; local instance context & Generate instance updates and newly visible participants & Candidate geometry and observation associations\\
\midrule
\multicolumn{4}{l}{\emph{Relation-agnostic geometric adjudication}}\\
Grounding selection & Full image and acquired crops; candidate boxes for one identity & Answer which candidate matches the visible instance & One grounding per identity\\
\midrule
\multicolumn{4}{l}{\emph{Grounded interaction reasoning}}\\
Relation assessment & Full image and a relevant crop; selected participants; public relation definition & Answer pair, action, agent, visibility, or binding questions & Relation supports and final detections\\\bottomrule
\end{tabularx}
\end{center}
\end{table}

\subsection{Interaction-Guided Perception}
\paragraph{Initial localization and grouped discovery.}
Initial localization groups each person with the identifiable targets involved in their interactions. We then divide the initially localized humans into groups of at most four for discovery. Conditioned on the full image and initial entity set, each group is examined for additional interactions and participants. Newly discovered humans participate in subsequent geometric and relational assessment, while discovery groups are defined by the initial human set.

A hypothesis specifies the human, action, and target identities, together with semantic roles for V-COCO. Each person may participate in multiple actions with multiple targets. Existing targets are referenced by identity, and new participants are localized by category and complete visible extent. A hypothesis with unresolved target localization can request a local observation, specifying the human, target description, action and role, and region to inspect.

\paragraph{Observation selection.}
The harness executes at most two observation requests per image. Requests concerning an unlocalized target take precedence over requests to complete an existing participant's boundary. Within each priority, different humans receive their first observation before a second request for the same human is considered. Requests for the same known target are deduplicated by identity; unresolved targets are distinguished by their associated human and description. Each discovery group may propose two requests, while the image-level budget governs execution after merging.

Given the target description and instances within the requested region, the MLLM refines eligible groundings and localizes additional visible participants. A single observation can therefore provide localization candidates for several participants in the same crop.

\paragraph{Local-to-global geometry.}
For a crop with image-pixel boundaries $[L,T,R,B]$, let $(x,y)$ be a point normalized to $[0,1000]$ within the crop. Its image coordinate is
\begin{equation}
 \begin{bmatrix}x'\\y'\end{bmatrix}
 =\begin{bmatrix}L\\T\end{bmatrix}
 +\frac{1}{1000}
  \begin{bmatrix}R-L&0\\0&B-T\end{bmatrix}
  \begin{bmatrix}x\\y\end{bmatrix}.
 \label{eq:app_crop_mapping}
\end{equation}
Here $(x',y')$ is a full-image pixel position. For an image of size $W\times H$, $(1000x'/W,1000y'/H)$ gives its normalized coordinates. Both box corners are transformed to this full-image representation before updating the candidate set. An existing identity is eligible for refinement when all its candidate boxes lie within the crop; boundary-crossing instances provide contextual references. Localized boxes cover the visible extent of each instance.

\paragraph{Observation associations.}
Let $h_\rho$ denote the human associated with observation $\rho$, $t_\rho$ its known target when available, and $\E_\rho^{\mathrm{new}}$ the newly localized identities. For the acquired observations $\mathcal{R}_{\mathrm{obs}}$, the corresponding relation associations are
\begin{equation}
 \Pp_O=\bigcup_{\rho\in\mathcal{R}_{\mathrm{obs}}}
 \bigl\{(h_\rho,o):o\in\E_\rho^{\mathrm{new}}\cup\{t_\rho\},\ o\neq h_\rho\bigr\},
 \label{eq:app_observation_pairs}
\end{equation}
where $\{t_\rho\}$ is omitted for an unlocalized target. These associations connect the requested target and newly localized participants to the human whose interaction prompted the observation, forming candidate pairs for grounded relation verification.

\subsection{Relation-Agnostic Geometric Adjudication}
\paragraph{One grounding per identity.}
For an identity with multiple grounding candidates, the MLLM compares their category and box descriptions against the full image and acquired crops. Each candidate is assigned an answer letter, and the candidate with the highest answer support is selected. The judgment considers the participant's identity, category, and complete visible extent to establish a shared grounding across actions and roles. Identities with one candidate retain their original grounding.

The selected geometry is shared by every relation involving the participant: for relations $d_1$ and $d_2$ containing identity $i$, $b_i(d_1)=b_i(d_2)=b_i^*$, where $b_i^*=b_{i,m_i^*}$. With at most two observations, an identity has at most three candidates: its initial or discovery grounding and up to two local revisions.

\subsection{Grounded Interaction Reasoning}
\paragraph{Question construction.}
The question context comprises the selected participants' identities, categories, and image-coordinate boxes, the public action or role definition, and the corresponding image views. Pair questions assess the presence of an interaction, and action questions verify a specific predicate. Agent, role-state, and binding questions respectively assess action occurrence, target visibility, and role identity. The prompt formulations appear in Appendix~\ref{app:prompts}.

\paragraph{Visual context.}
Localization and discovery use the full image, local observation uses the requested crop, and geometric adjudication receives all acquired views. Relation questions use the full image and, when available, one local view associated with the queried participants. The source view $v_{i,m_i^*}$ of a selected grounding takes priority, with the most recent observation chosen when both participants have local source views. If neither selected grounding has a local source view, the crop is taken from the most recent compatible observation request.

A compatible request concerns the same human and agrees with the question's specified action and target, or leaves those attributes unspecified. For role questions, it concerns the queried role or general participant localization. Agent questions admit requests for any role, pair questions use general participant requests, and no-interaction questions impose no action or role restriction. The full image provides the visual context when no compatible local view is available.

\paragraph{Answer support.}
Answer supports are obtained by normalizing the raw first-token logits according to Equation~\ref{eq:support}. Pair, action, agent, and binding questions use \{\texttt{false}, \texttt{true}, \texttt{unknown}\}; role-state questions use \{\texttt{missing}, \texttt{visible}, \texttt{unknown}\}; geometry questions use the candidate letters. Each label corresponds to one token. These normalized supports determine the selected geometry and final detection scores, with the generation temperature applied to perceptual generation alone.

\paragraph{Candidate retention and multi-label prediction.}
Discovery and observation associations complement visual pair assessment in Equation~\ref{eq:route}, with $\Pp_D\cup\Pp_O\subseteq\K$. Each retained HICO-DET pair is evaluated over all positive actions compatible with its target category, using a separate answer distribution for each action. The \texttt{no\_interaction} label is evaluated independently over all directed pairs.

\paragraph{Physical and missing role targets.}
V-COCO uses the agent support $q_A$, role-state supports $q_V$ and $q_M$, and binding support $q_B$ defined in the main text. Physical candidates share the action and visibility factors and receive individual binding assessments, giving $s_{\mathrm{physical}}=q_Aq_Vq_B$. The NULL alternative combines action and missing-target support as $s_{\mathrm{NULL}}=q_Aq_M$. Since visible, missing, and unknown states share one answer distribution, $q_V+q_M\leq1$; the combined score of any one physical candidate and its NULL alternative is bounded by $q_A$.

\subsection{Complete Inference Procedure}
Table~\ref{tab:inference_procedure} summarizes the complete inference procedure, from participant localization and observation to grounded interaction prediction.
\begin{table}[!htb]
\caption{Inference procedure for one image. All generation and question-answering operations use the same frozen MLLM.}
\label{tab:inference_procedure}
\begin{center}
\small
\begin{tabularx}{\linewidth}{@{}r>{\raggedright\arraybackslash}X@{}}\toprule
Step & Operation\\\midrule
1 & Localize initial entities $\E_0$ and identify the initial humans $\Hh_0$. Initialize each entity's geometry candidates with its returned box.\\
2 & Partition $\Hh_0$ into groups of at most four. For each group, discover interactions and participants using the full image and $\E_0$. Associate the resulting hypotheses and observation requests with the expanded participant set.\\
3 & Select at most two image-level observation requests. Localize participants in each requested crop, map their boxes to image coordinates, and update geometry candidates and observation associations.\\
4 & For each multi-candidate identity, ask the geometry question over the full image and acquired crops. Select one grounding and construct the shared entity state $\E^*$.\\
5 & Assess all directed human-entity pairs. Form $\K$ from pair support, discovery associations, and observation associations. For HICO-DET, independently assess no-interaction for all pairs.\\
6 & For HICO-DET, evaluate every compatible positive action on each retained pair through a visual question. For V-COCO, assess each human's actions and role states, followed by role bindings for retained pairs.\\
7 & Compute the HICO-DET and V-COCO scores using Equations~\ref{eq:hico_score} and~\ref{eq:vcoco}. Form the final detections from the selected boxes and continuous scores, retaining maximum support for exact duplicates.\\\bottomrule
\end{tabularx}
\end{center}
\end{table}

\FloatBarrier
\section{Experimental Setup and Evaluation Protocols}
\label{app:protocol}
\subsection{Detection and Semantic-Role Evaluation}
\paragraph{HICO-DET.}
We evaluate all 9,658 official test images over 600 human-object interaction classes, comprising 520 positive-interaction classes and 80 \texttt{no\_interaction} classes. Each prediction consists of a human box, an object box, an interaction class, and a confidence score. A detection is correct when its class matches the annotation and both boxes reach IoU 0.5. Predictions are matched in descending score order within each class, with at most one true positive per annotation. Rare contains 138 classes with fewer than ten training instances, and Non-Rare contains the remaining 462 classes.

Default evaluation includes all test images. Known Object evaluates each interaction class on images containing an annotated instance of its object category. Both settings use the same predictions, with Known Object restricting the images considered when assessing false positives.

\paragraph{Average precision.}
HICO-DET uses the VOC 2007 eleven-point definition~\citep{everingham2010voc}. For class $c$, let $p_c(r)$ denote precision at an attained recall value $r$, and let $\mathcal{R}_{\mathrm{AP}}=\{0,0.1,\ldots,1\}$. Then
\begin{equation}
 \mathrm{AP}_c=\frac{1}{11}\sum_{t\in\mathcal{R}_{\mathrm{AP}}}
       \max_{r\geq t}p_c(r),
 \qquad
 \mathrm{mAP}(\mathcal{C})=\frac{1}{|\mathcal{C}|}
       \sum_{c\in\mathcal{C}}\mathrm{AP}_c,
 \label{eq:app_ap}
\end{equation}
where a recall threshold with no qualifying prediction contributes zero. Full mAP averages all 600 classes. We report AP as percentages and compute differences before rounding.

\paragraph{V-COCO.}
We evaluate all 4,946 official test images. Agent evaluation measures whether the localized person performs one of 26 actions. Role evaluation additionally identifies the target fulfilling the corresponding semantic role and averages over 25 action-role categories, including \texttt{point}. Actions such as \texttt{hit}, \texttt{eat}, and \texttt{cut} distinguish an object from an instrument. Conversely, \texttt{stand}, \texttt{walk}, \texttt{run}, and \texttt{smile} have agent annotations without a target role.

For a role with an annotated physical target, both the person and role target must match at IoU 0.5. When the role target is annotated as missing, Scenario 1 (S1) requires a matching person and a NULL target prediction. Scenario 2 (S2) relaxes target localization for that missing-role annotation. Both scenarios follow the benchmark's treatment of incompletely annotated people and integrate the precision-recall envelope.

\paragraph{V-COCO reporting conventions.}
\label{app:role_conventions}
The primary S1/S2 metric for \method{}, direct-generation baselines, and ablations is the 25-role mean, including \texttt{point}. Appendix~\ref{app:roles} additionally gives the 24-role mean excluding \texttt{point} and the 26-action agent mean. Published comparison results retain their sources' role definitions and missing-target conventions; the S1/S2 columns follow the scenario designations in those sources.

\paragraph{Coordinate conventions.}
Predicted boxes use coordinates normalized within the input view and are mapped to image pixels for evaluation. HICO-DET then shifts all four zero-based coordinates by one and computes IoU with inclusive pixel extents, following its annotation convention. Crop coordinates are mapped to the full image as specified in Appendix~\ref{app:policy}. All official test images are evaluated, including empty outputs and valid detections from partially completed inference.

\subsection{Class Partitions and Development Settings}
\paragraph{Open-vocabulary partitions.}
We report rare-first unseen combinations (RF-UC), non-rare-first unseen combinations (NF-UC), unseen objects (UO), and unseen verbs (UV). The Unseen subsets contain 120, 120, 100, and 84 classes, respectively; their Seen complements contain 480, 480, 500, and 516 classes. We use the class masks released with GEN-VLKT~\citep{liao2022genvlkt}. Each partition aggregates the same class AP values from the same predictions. In particular,
\begin{equation}
 \mathrm{mAP}_{\mathrm{Full}}
 =\frac{|\mathcal{C}_{U}|\mathrm{mAP}_{U}
       +|\mathcal{C}_{S}|\mathrm{mAP}_{S}}{600},
 \label{eq:app_partition_mean}
\end{equation}
where $\mathcal{C}_{U}$ and $\mathcal{C}_{S}$ are the benchmark's Unseen and Seen class sets. \method{} uses the full public interaction vocabulary and the same frozen model and inference policy across these partitions. The class masks are applied to the resulting AP values, so the Full result is shared by all four settings.

\paragraph{Development protocol.}
Prompts and inference policy are developed using evaluation feedback from fixed subsets of the official train set: 128 images for optimization and 256 for confirmation, with four additional images for interface checks. Development feedback also includes train-image inspection and evaluation on the complete train set. The final configuration is evaluated on every image in the official test sets.

The optimization and confirmation sets are sampled without replacement from disjoint source subsets, excluding the interface images. Sampling uses fixed seeds, and each image retains its index in the official train list for generation across both development-subset and complete-train-set runs. The image lists and seed assignments will accompany the released evaluation configurations.

\paragraph{Model inputs.}
The model remains frozen throughout development and evaluation. Its inference inputs comprise the image, generated participants, and public action and role definitions. Target-benchmark annotations are used for development feedback, evaluation, and the ground-truth-based analyses described below; they are not supplied to the model at inference.

\subsection{Model and Inference Configuration}
\label{app:implementation}
We use frozen Qwen3.5-9B for participant localization, interaction discovery, local observation, and grounded question answering. Its native visual encoder processes both the full image and local crops. Table~\ref{tab:inference_configuration} summarizes the inference settings.

\begin{table}[!htb]
\caption{Inference configuration. Generation settings apply to localization, discovery, and observation; visual question answering reads raw first-token logits over its answer set.}
\label{tab:inference_configuration}
\begin{center}
\small
\begin{tabularx}{\linewidth}{@{}l>{\raggedright\arraybackslash}X@{}}\toprule
Setting & Value\\\midrule
\multicolumn{2}{l}{\emph{Model and visual inputs}}\\
Backbone & Frozen Qwen3.5-9B; shared weights across stages\\
Precision & BF16 weights and KV cache\\
Maximum context & 16,384 tokens\\
Full-image processor & Roughly 3 MP minimum and 17 MP maximum, depending on aspect ratio\\
Crop processor & Roughly 0.8 MP\\
\midrule
\multicolumn{2}{l}{\emph{Generation and visual question answering}}\\
Initial localization output & At most 4,096 tokens\\
Discovery / observation output & At most 2,048 tokens per response\\
Generation sampling & Temperature 0.7; top-$p$ 0.8; top-$k$ 20; min-$p$ 0\\
Generation penalties & Presence 1.5; frequency 0; repetition 1\\
Question answering & One output token; raw logits for the designated labels\\
Generation seeds & Base seed 66; seeds 67 and 68 for independent runs; fixed image and stage offsets\\
\midrule
\multicolumn{2}{l}{\emph{Harness settings}}\\
Discovery group & At most four initially localized humans\\
Observation budget & At most two requests per image\\
Visual context & At most the full image and two distinct crops\\
\midrule
Software & Torch 2.10.0; vLLM 0.19.1; Transformers 5.15.1; xgrammar 0.2.3\\\bottomrule
\end{tabularx}
\end{center}
\end{table}

\paragraph{Image views.}
The image processor preserves aspect ratio and aligns the input dimensions to the model's visual grid. Local crops are extracted from the original image and encoded by the same backbone.

\paragraph{Generation and scoring.}
Perceptual generation produces participants, interaction hypotheses, and localization updates. Visual question answering uses one output token with decoding temperature zero; answer support is computed from the raw logits before output-grammar modification. Independent generation runs preserve the image and stage seed offsets and vary the base seed as specified in Appendix~\ref{app:stability}.

\paragraph{Detection output.}
We retain continuous detection scores and merge exact duplicate predictions by maximum support. No score threshold or non-maximum suppression is applied. V-COCO predictions include agent actions, physical-role targets, and NULL alternatives, scored according to Equation~\ref{eq:vcoco}.

\subsection{Baseline Inference}
\label{app:baselines}
We compare three inference strategies using frozen Qwen3.5-9B on the complete test sets. Direct generation predicts participants and relations in one response, with a shared person box across actions. The initial-instance baseline applies visual pair assessment and relation verification to the initially localized participants. \method{} additionally performs interaction-guided discovery, observation, and geometric adjudication. Direct generation uses the same full-image processing, sampling settings, and initial output budget.

\paragraph{Direct-generation confidence.}
For each generated interaction record, let $\mathcal{J}$ contain the token positions overlapping its emitted interaction label. The direct-generation score is the geometric mean of the corresponding token probabilities:
\begin{equation}
 s_{\mathrm{gen}}
 =\exp\!\left(
   \frac{1}{|\mathcal{J}|}
   \sum_{j\in\mathcal{J}}\log p_\theta(y_j\mid I,y_{<j})
 \right).
 \label{eq:app_direct_score}
\end{equation}
We use token probabilities before output-grammar modification. HICO-DET scores the emitted verb-object label; V-COCO scores the action label and assigns this score to its agent prediction and explicitly generated physical or NULL roles.

\subsection{Evaluation Coverage}
Evaluation includes all 9,658 HICO-DET and 4,946 V-COCO test images. Valid predictions from partially completed inference are evaluated normally. Images with no predictions, numbering 98 and 12 on the two benchmarks, respectively, remain in the evaluation population.

\subsection{Evaluation Implementations}
We use a common evaluator for \method{}, its ablations, and the released predictions re-evaluated in Tables~\ref{tab:common} and~\ref{tab:common-open}. Tables~\ref{tab:evaluator_comparison} and~\ref{tab:evaluator_vcoco} compare this implementation with the original HICO-DET code under Octave and the original V-COCO evaluator on the same predictions, including direct generation and the principal perceptual and relational ablations.

On HICO-DET, the original code under Octave gives 32.97\% Default and 36.61\% Known Object Full mAP, compared with 33.20\% and 36.85\% from the common evaluator. The difference follows from the recall-threshold sequences constructed by the two implementations, which sample the precision envelope at slightly different boundaries in eleven-point AP. V-COCO role scores agree at the reported precision.
\begin{table}[!htb]
\caption{Evaluation implementations on identical HICO-DET predictions (Full mAP, \%). Common denotes the evaluator used throughout this paper; Original denotes the benchmark code under Octave.}
\label{tab:evaluator_comparison}
\begin{center}
\small
\begin{tabular}{lrrrr}\toprule
& \multicolumn{2}{c}{Default} & \multicolumn{2}{c}{Known Object}\\
\cmidrule(lr){2-3}\cmidrule(l){4-5}Method & Common & Original & Common & Original\\
\midrule
Direct generation & 15.62 & 15.50 & 17.27 & 17.11\\
Initial-instance baseline & 31.45 & 31.23 & 35.05 & 34.82\\
w/o targeted observation & 32.80 & 32.56 & 36.39 & 36.15\\
First-candidate geometry & 32.95 & 32.71 & 36.55 & 36.30\\
\method{} & 33.20 & 32.97 & 36.85 & 36.61\\
\bottomrule
\end{tabular}
\end{center}
\end{table}
\begin{table}[!htb]
\caption{Evaluation implementations on identical V-COCO predictions (role AP, \%). Common denotes the evaluator used in this paper; Original denotes the benchmark implementation. Both average the same 25 roles.}
\label{tab:evaluator_vcoco}
\begin{center}
\small
\begin{tabular}{lrrrr}\toprule
& \multicolumn{2}{c}{S1} & \multicolumn{2}{c}{S2}\\
\cmidrule(lr){2-3}\cmidrule(l){4-5}Method & Common & Original & Common & Original\\
\midrule
Direct generation & 16.41 & 16.41 & 21.20 & 21.20\\
Joint role question & 41.10 & 41.10 & 46.30 & 46.30\\
\method{} & 44.65 & 44.65 & 49.70 & 49.70\\
\bottomrule
\end{tabular}
\end{center}
\end{table}

\section{Extended Benchmark Results}
\label{app:comparisons}

\subsection{Regular and Open-Vocabulary Comparisons}
Tables~\ref{tab:extended} and~\ref{tab:extended-open} provide the complete regular and open-vocabulary comparisons corresponding to Tables~\ref{tab:main} and~\ref{tab:open}. They include additional backbone configurations, MLLM-based methods, retrieval baselines, and the training-free variant of DA-HOI.
\begin{table}[ht]
\caption{Complete regular-setting comparison corresponding to Table~\ref{tab:main}, including additional model configurations and training-free baselines. HICO-DET reports Default and Known Object mAP (\%); V-COCO reports S1/S2 role AP (\%). HarnessHOI uses the 25-role definition. Within each method group, the best results are bold; the strongest previous training-free results are underlined. Dashes denote unavailable entries.}
\label{tab:extended}
\begin{center}
\small
\setlength{\tabcolsep}{2pt}
\renewcommand{\arraystretch}{1.08}
\resizebox{\linewidth}{!}{%
\begin{tabular}{llcccccccc}
\toprule
& & \multicolumn{3}{c}{HICO-DET Default} & \multicolumn{3}{c}{HICO-DET Known Object} & \multicolumn{2}{c}{V-COCO}\\
\cmidrule(lr){3-5}\cmidrule(lr){6-8}\cmidrule(l){9-10}
Method & Model & Full & Rare & Non-Rare & Full & Rare & Non-Rare & S1 & S2\\
\midrule
\multicolumn{10}{l}{\emph{Task-optimized methods}}\\
QPIC~\citep{tamura2021qpic} & R101 & 29.90 & 23.92 & 31.69 & 32.38 & 26.06 & 34.27 & 58.30 & 60.70\\
UPT~\citep{zhang2022upt} & R101-DC5 & 32.62 & 28.62 & 33.81 & 36.08 & 31.41 & 37.47 & 61.30 & 67.10\\
GEN-VLKT~\citep{liao2022genvlkt} & R101/CLIP-B & 34.95 & 31.18 & 36.08 & 38.22 & 34.36 & 39.37 & 63.58 & 65.93\\
HOICLIP~\citep{ning2023hoiclip} & R50/CLIP-B & 34.69 & 31.12 & 35.74 & 37.61 & 34.47 & 38.54 & 63.50 & 64.80\\
PViC~\citep{zhang2023pvic} & Swin-L/H-DETR & 44.32 & 44.61 & 44.24 & 47.81 & 48.38 & 47.64 & 64.10 & 70.20\\
CLIP4HOI~\citep{mao2023clip4hoi} & R50/CLIP-B & 35.33 & 33.95 & 35.74 & 37.19 & 35.27 & 37.77 & -- & 66.30\\
UniHOI-s~\citep{cao2023unihoi} & R50/BLIP-2 & 40.06 & 39.91 & 40.11 & 42.20 & 42.60 & 42.08 & 65.58 & 68.27\\
CMMP~\citep{lei2024cmmp} & R50/CLIP-L & 38.14 & 37.75 & 38.25 & -- & -- & -- & -- & 64.00\\
BCOM~\citep{wang2024bcom} & R50/CLIP-L & 39.34 & 39.90 & 39.17 & 42.24 & 42.86 & 42.05 & 65.80 & 69.90\\
LAIN~\citep{kim2025lain} & R50/CLIP-B & 36.02 & 35.70 & 36.11 & -- & -- & -- & -- & 65.10\\
HOLa~\citep{lei2025hola} & R50/CLIP-L & 39.05 & 38.66 & 39.17 & -- & -- & -- & 60.30 & 66.00\\
ADA-CM (FT)~\citep{lei2023adacm} & R50/CLIP-L & 38.40 & 37.52 & 38.66 & -- & -- & -- & 58.57 & 63.97\\
VRDiff~\citep{cao2025vrdiff} & R50/CLIP-L & 41.40 & 41.69 & 41.31 & 44.25 & 44.65 & 44.13 & -- & --\\
HORP~\citep{geng2025horp} & Swin-L/CLIP-B/32 & 47.53 & 46.81 & 47.74 & 51.24 & 50.78 & \textbf{51.38} & 68.90 & 71.10\\
VDRP~\citep{yang2025vdrp} & R50/CLIP-B & 39.07 & 39.08 & 39.06 & -- & -- & -- & 60.60 & 66.20\\
ContextHOI~\citep{jia2025contexthoi} & R101/CLIP-L & 42.09 & 42.41 & 41.99 & -- & -- & -- & 67.30 & --\\
InterProDa~\citep{jia2025interproda} & R50/CLIP-L & 42.67 & 45.21 & 41.92 & -- & -- & -- & 67.60 & --\\
BC-HOI~\citep{hu2025bchoi} & R50/BLIP-2 & 43.01 & 45.76 & 42.18 & -- & -- & -- & 68.20 & 70.61\\
NMSR~\citep{yang2025sibling} & R50/BLIP-2 & 43.98 & 43.27 & 44.00 & 45.90 & 45.29 & 46.32 & 71.10 & 73.30\\
HOI-IDiff~\citep{hui2025hoiidiff} & DDETR & 47.71 & 48.36 & 47.52 & 50.56 & 51.95 & 50.14 & 73.40 & 76.10\\
InCoM-Net~\citep{seo2026incom} & R50/CLIP-L & 43.96 & 45.61 & 43.46 & 45.59 & 45.92 & 45.49 & \textbf{73.60} & 75.40\\
SL-HOI~\citep{sun2026slhoi} & DINOv3-L & 45.05 & 47.71 & 44.25 & -- & -- & -- & -- & --\\
InstructHOI~\citep{luo2025instructhoi} & R101/ViT-L & 47.68 & 47.97 & 47.59 & 49.89 & 50.92 & 49.58 & 72.40 & 76.10\\
HOI-R1~\citep{chen2026hoir1} & Qwen2.5-VL-3B & 18.33 & 16.03 & 19.02 & 19.83 & 16.25 & 20.90 & -- & --\\
HOI-R1~\citep{chen2026hoir1} & Qwen3-VL-4B & 23.35 & 23.58 & 23.28 & 25.12 & 24.11 & 25.42 & -- & --\\
HOI-R1~\citep{chen2026hoir1} & Rex-Omni-3B & 25.24 & 22.57 & 26.04 & 27.00 & 23.21 & 28.14 & -- & --\\
DA-HOI (FT, DETR)~\citep{xuan2026zeroshothoidetectionmllmbased} & R50/Qwen2.5-VL-3B & 44.58 & 46.17 & 44.08 & -- & -- & -- & 69.40 & 71.70\\
DA-HOI (FT, G-DINO)~\citep{xuan2026zeroshothoidetectionmllmbased} & G-DINO/Qwen2.5-VL-3B & 46.21 & 47.54 & 45.81 & -- & -- & -- & 70.40 & 72.50\\
LINK~\citep{wu2026link} & Swin-L/CLIP-L & \textbf{49.06} & \textbf{53.63} & 47.60 & \textbf{51.34} & \textbf{56.29} & 49.86 & -- & 69.20\\
FineHOI~\citep{tonini2026finehoi} & DETR/DINOtxt-L/16 & 39.13 & 39.91 & 38.89 & -- & -- & -- & -- & 66.50\\
UniHOI~\citep{yang2026unihoi} & VQGAN/Llama3-8B & 48.16 & 50.74 & \textbf{48.12} & \textbf{51.34} & 53.72 & 50.33 & 72.91 & \textbf{77.45}\\
Funnel-HOI~\citep{sarma2026funnelhoi} & R50/CLIP-B/32 & 34.36 & 31.07 & 35.33 & 37.24 & 33.76 & 38.27 & 61.60 & 64.60\\
ScriptHOI~\citep{nguyen2026scripthoi} & -- & 37.10 & 37.80 & 36.90 & 40.00 & 41.40 & 39.60 & 67.80 & 69.40\\
HOIBlender-2XL~\citep{chen2026hoiblender} & DINOv2-B/BLIP-2 & 44.49 & 43.21 & 44.86 & 47.45 & 47.76 & 47.36 & -- & --\\
\midrule
\multicolumn{10}{l}{\emph{Training-free methods}}\\
BLIP-2 analysis~\citep{ai2024trainingfree} & G-DINO/BLIP-2-H@364 & 19.71 & 24.07 & 18.41 & -- & -- & -- & -- & --\\
Qwen2.5-VL direct~\citep{chen2026hoir1} & 72B-AWQ & 20.71 & 29.62 & 18.05 & \underline{22.93} & \underline{31.87} & \underline{20.26} & -- & --\\
Qwen3-VL direct~\citep{chen2026hoir1} & 8B & 17.41 & 21.96 & 16.05 & 19.17 & 23.39 & 17.91 & -- & --\\
LongCLIP retrieval~\citep{tonini2025dysco} & R50/LongCLIP-L & 23.13 & 31.32 & 20.68 & -- & -- & -- & 40.13 & 45.11\\
ADA-CM (TF)~\citep{lei2023adacm} & R50/CLIP-B & 25.19 & 27.24 & 24.58 & -- & -- & -- & 39.09 & 43.93\\
DYSCO~\citep{tonini2025dysco} & R50/LongCLIP-L & 29.00 & 33.63 & \underline{27.62} & -- & -- & -- & \underline{42.80} & \underline{47.82}\\
AgentHOI (Wang et al.)~\citep{wang2026agenthoi} & YOLOv8/BLIP/Qwen2.5 & 13.26 & -- & -- & -- & -- & -- & -- & --\\
AgentHOI (Lei et al.)~\citep{lei2026unleashing} & G-DINO/GPT-4o & 29.61 & \underline{40.33} & 26.41 & -- & -- & -- & -- & --\\
DA-HOI (TF)~\citep{xuan2026zeroshothoidetectionmllmbased} & R50/Qwen2.5-VL-3B & \underline{31.50} & -- & -- & -- & -- & -- & -- & --\\
\rowcolor{blue!6}
\textbf{HarnessHOI} & Qwen3.5-9B & \textbf{33.20} & \textbf{41.63} & \textbf{30.68} & \textbf{36.85} & \textbf{44.20} & \textbf{34.65} & \textbf{44.65} & \textbf{49.70}\\
\bottomrule
\end{tabular}}
\end{center}
\end{table}

\paragraph{DA-HOI and vision-language baselines.}
The fine-tuned DA-HOI~\citep{xuan2026zeroshothoidetectionmllmbased} comparisons include both DETR and GroundingDINO configurations under regular and open-vocabulary evaluation. Its training-free variant combines R50-DETR proposals with deterministic MLLM generation. DYSCO~\citep{tonini2025dysco} uses LongCLIP-L for regular evaluation and CLIP ViT-B/16 for open-vocabulary evaluation.

\begin{table}[ht]
\caption{Complete open-vocabulary comparison corresponding to Table~\ref{tab:open}, including additional methods and training-free baselines. Each partition reports Unseen, Seen, and Full mAP (\%) using predicted boxes. HarnessHOI uses the same predictions across all four partitions. Within each method group, the best results are bold; the strongest previous training-free results are underlined. Dashes denote unavailable entries. CLIP baseline results are reported by LAIN.}
\label{tab:extended-open}
\begin{center}
\small
\setlength{\tabcolsep}{2pt}
\renewcommand{\arraystretch}{1.08}
\resizebox{\linewidth}{!}{%
\begin{tabular}{lcccccccccccc}
\toprule
&\multicolumn{3}{c}{RF-UC}&\multicolumn{3}{c}{NF-UC}&\multicolumn{3}{c}{UO}&\multicolumn{3}{c}{UV}\\
\cmidrule(lr){2-4}\cmidrule(lr){5-7}\cmidrule(lr){8-10}\cmidrule(l){11-13}
Method & Unseen & Seen & Full & Unseen & Seen & Full & Unseen & Seen & Full & Unseen & Seen & Full\\
\midrule
\multicolumn{13}{l}{\emph{Task-optimized methods}}\\
GEN-VLKT~\citep{liao2022genvlkt} & 21.36 & 32.91 & 30.56 & 25.05 & 23.38 & 23.71 & 10.51 & 28.92 & 25.63 & 20.96 & 30.23 & 28.74\\
HOICLIP~\citep{ning2023hoiclip} & 25.53 & 34.85 & 32.99 & 26.39 & 28.10 & 27.75 & 16.20 & 30.99 & 28.53 & 24.30 & 32.19 & 31.09\\
CLIP4HOI~\citep{mao2023clip4hoi} & 28.47 & 35.48 & 34.08 & 31.44 & 28.26 & 28.90 & 31.79 & 32.73 & 32.58 & 26.02 & 31.14 & 30.42\\
ADA-CM (FT)~\citep{lei2023adacm} & 27.63 & 34.35 & 33.01 & 32.41 & 31.13 & 31.39 & -- & -- & -- & -- & -- & --\\
BCOM~\citep{wang2024bcom} & 28.52 & 35.04 & 33.74 & 33.12 & 31.76 & 32.03 & -- & -- & -- & -- & -- & --\\
CMMP (ViT-L)~\citep{lei2024cmmp} & 35.98 & 37.42 & 37.13 & 33.52 & 35.53 & 35.13 & 39.67 & 36.15 & 36.74 & 30.84 & 37.28 & 36.38\\
InterProDa~\citep{jia2025interproda} & 36.38 & 40.88 & 39.58 & 33.64 & 36.47 & 35.50 & -- & -- & -- & -- & -- & --\\
INP-CC~\citep{lei2025inpcc} & 17.38 & 24.74 & 23.12 & -- & -- & -- & -- & -- & -- & -- & -- & --\\
SGC-Net~\citep{lin2025sgcnet} & 23.27 & 28.34 & 27.22 & -- & -- & -- & -- & -- & -- & -- & -- & --\\
VDRP (ViT-L)~\citep{yang2025vdrp} & 36.72 & 38.48 & 38.13 & 37.48 & 36.21 & 36.46 & 39.36 & 37.50 & 37.81 & 31.16 & 38.16 & 37.18\\
LAIN (ViT-L)~\citep{kim2025lain} & 36.57 & 38.54 & 38.13 & 37.52 & 35.90 & 36.22 & 40.78 & 36.96 & 37.60 & 32.05 & 38.04 & 37.20\\
EZ-HOI (ViT-L)~\citep{lei2024ezhoi} & 34.24 & 37.35 & 36.73 & 36.33 & 34.47 & 34.84 & 38.17 & 36.02 & 36.38 & 28.82 & 38.15 & 36.84\\
UniHOI-s (BLIP-2)~\citep{cao2023unihoi} & 28.68 & 33.16 & 32.27 & 28.45 & 32.63 & 31.79 & 19.72 & 34.76 & 31.56 & 26.05 & 36.78 & 34.68\\
HOLa (ViT-B)~\citep{lei2025hola} & 30.61 & 35.08 & 34.19 & 35.25 & 31.64 & 32.36 & 36.45 & 33.02 & 33.59 & 27.91 & 35.09 & 34.09\\
NMSR (CLIP)~\citep{yang2025sibling} & 35.48 & -- & -- & 36.12 & -- & -- & 31.01 & -- & -- & 30.47 & -- & --\\
VRDiff (ViT-L)~\citep{cao2025vrdiff} & 38.92 & 40.83 & 40.45 & 39.33 & 39.03 & 39.09 & 41.52 & 39.66 & 39.97 & 32.14 & 40.61 & 39.43\\
BC-HOI (BLIP-2)~\citep{hu2025bchoi} & 42.31 & 40.67 & 40.99 & 33.01 & 37.24 & 36.40 & 19.94 & 37.03 & 34.18 & 31.18 & 41.31 & 39.89\\
SD-IF~\citep{xu2026sdif} & 28.01 & 30.09 & 29.63 & -- & -- & -- & -- & -- & -- & -- & -- & --\\
InCoM-Net~\citep{seo2026incom} & 37.69 & 41.46 & 40.71 & 39.45 & 38.42 & 38.62 & -- & -- & -- & -- & -- & --\\
InstructHOI~\citep{luo2025instructhoi} & 36.82 & 44.86 & 43.25 & 36.42 & 38.82 & 38.34 & 39.92 & 37.05 & 37.53 & 31.64 & 39.17 & 38.12\\
LINK (ViT-L)~\citep{wu2026link} & 38.51 & 40.33 & 39.97 & 34.63 & 35.67 & 35.43 & 38.24 & 37.61 & 37.91 & 27.22 & 38.36 & 36.88\\
FineHOI~\citep{tonini2026finehoi} & 37.79 & 38.55 & 38.39 & 38.11 & 36.30 & 36.58 & 40.13 & 35.92 & 36.62 & 35.58 & 38.51 & 38.10\\
Funnel-HOI~\citep{sarma2026funnelhoi} & 25.90 & 34.43 & 32.70 & 26.63 & 28.32 & 28.00 & 15.28 & 31.50 & 28.66 & 26.10 & 32.56 & 31.66\\
ImagineAgent~\citep{yuan2026imagineagent} & 37.85 & 36.49 & 36.76 & -- & -- & -- & -- & -- & -- & -- & -- & --\\
SL-HOI~\citep{sun2026slhoi} & 40.53 & 42.99 & 42.49 & -- & -- & -- & -- & -- & -- & -- & -- & --\\
DA-HOI (FT, DETR)~\citep{xuan2026zeroshothoidetectionmllmbased} & 41.79 & 44.01 & 43.56 & \textbf{43.12} & 39.63 & 40.33 & 48.67 & 42.58 & 43.60 & \textbf{36.89} & 43.84 & 42.88\\
DA-HOI (FT, G-DINO)~\citep{xuan2026zeroshothoidetectionmllmbased} & \textbf{43.30} & \textbf{45.19} & \textbf{44.81} & 41.52 & \textbf{41.51} & \textbf{41.51} & \textbf{50.15} & \textbf{44.31} & \textbf{45.28} & 36.88 & \textbf{45.66} & \textbf{44.43}\\
\midrule
\multicolumn{13}{l}{\emph{Training-free methods}}\\
CLIP baseline~\citep{kim2025lain} & 28.79 & 22.00 & 23.36 & 28.52 & 22.06 & 23.36 & 28.66 & \underline{22.29} & 23.36 & 26.16 & \underline{22.90} & 23.36\\
BLIP-2 analysis~\citep{ai2024trainingfree} & 23.96 & 18.65 & 19.71 & 19.97 & 19.65 & 19.71 & 22.71 & 19.12 & 19.71 & 21.96 & 19.35 & 19.71\\
ADA-CM (TF)~\citep{lei2023adacm} & 26.83 & \underline{24.54} & 25.00 & \underline{30.11} & 23.16 & 24.55 & -- & -- & -- & -- & -- & --\\
DYSCO~\citep{tonini2025dysco} & 30.36 & 23.96 & 25.24 & 27.56 & \underline{24.58} & 25.18 & -- & -- & -- & -- & -- & --\\
AgentHOI (Lei et al.)~\citep{lei2026unleashing} & \underline{38.64} & -- & -- & 28.04 & -- & -- & \underline{33.70} & -- & -- & \underline{29.85} & -- & --\\
DA-HOI (TF)~\citep{xuan2026zeroshothoidetectionmllmbased} & -- & -- & \underline{31.50} & -- & -- & \underline{31.50} & -- & -- & \underline{31.50} & -- & -- & \underline{31.50}\\
\rowcolor{blue!6}
\textbf{HarnessHOI} & \textbf{40.15} & \textbf{31.46} & \textbf{33.20} & \textbf{33.64} & \textbf{33.09} & \textbf{33.20} & \textbf{35.45} & \textbf{32.75} & \textbf{33.20} & \textbf{33.87} & \textbf{33.09} & \textbf{33.20}\\
\bottomrule
\end{tabular}}
\end{center}
\end{table}

\paragraph{Other comparison configurations.}
GEN-VLKT uses its large and small configurations for regular and open-vocabulary evaluation, respectively. VDRP follows its default ViT-B configuration for regular evaluation and its ViT-L configuration for open-vocabulary evaluation. The UO result of EZ-HOI uses its HICO-DET-pretrained detector. The CLIP baseline follows LAIN~\citep{kim2025lain}; direct Qwen inference and LongCLIP retrieval follow HOI-R1 and DYSCO. LINK's V-COCO S2 result follows its ICLR version. V-COCO reporting conventions are specified in Appendix~\ref{app:role_conventions}.

\FloatBarrier
\subsection{Models and Inference Resources}
Table~\ref{tab:resources} compares the models, inference strategies, and additional resources used by related methods. \method{} uses one frozen MLLM for participant acquisition, geometric adjudication, and relation verification. The public vocabulary and development protocol are specified in Appendix~\ref{app:protocol}.
\begin{table}[!htb]
\caption{Resource and organization comparison. VisHarness provides a related visual-reasoning formulation; the remaining systems address HOI detection.}
\label{tab:resources}
\begin{center}
\footnotesize
\begin{tabularx}{\linewidth}{@{}l>{\raggedright\arraybackslash}X>{\raggedright\arraybackslash}X>{\raggedright\arraybackslash}X@{}}\toprule
System & Capability source & Task organization & Additional resources\\\midrule
HOI-R1 & SFT+GRPO-tuned MLLM & Joint interaction reasoning and localization & HOI instruction and reward supervision\\
DA-HOI (FT) & Adapted Qwen2.5-VL-3B & Spatial-aware pooling and deterministic matching & DETR or GroundingDINO; HOI supervision\\
DA-HOI (TF) & Frozen Qwen2.5-VL-3B & Candidate-conditioned deterministic generation & R50 DETR proposals\\
AgentHOI (Wang et al.) & BLIP and Qwen2.5 & Hypothesis proposal, multi-agent debate, and consensus & YOLOv8\\
AgentHOI (Lei et al.) & GPT-4o or Qwen2.5-VL-72B & Recognition, re-mining, reassignment, grounding & GroundingDINO\\
DYSCO & CLIP/LongCLIP & Visual-semantic registry and scoring & R50 detector; target-domain visual support\\
VisHarness & Trained visual agent & Learned multi-turn expert selection & Visual experts; policy training\\
HarnessHOI & One frozen Qwen3.5-9B & Interaction-guided perception, geometric adjudication, and grounded reasoning & No auxiliary models or retrieved visual support sets\\\bottomrule
\end{tabularx}
\end{center}
\end{table}

\paragraph{Agent-based HOI inference.}
AgentHOI by Wang et al.~\citep{wang2026agenthoi} combines hypothesis proposal with multi-agent debate, whereas AgentHOI by Lei et al.~\citep{lei2026unleashing} combines multi-round interaction reasoning with instance-specific visual grounding. Table~\ref{tab:extended} includes both methods under regular evaluation, while Table~\ref{tab:extended-open} includes the open-vocabulary results reported by Lei et al. Wang et al. also report 14.58 mAP on V-COCO without an S1/S2 designation.

\paragraph{Task-optimized configurations.}
The extended tables include the learned interaction representations of FineHOI~\citep{tonini2026finehoi}, SGC-Net~\citep{lin2025sgcnet}, Funnel-HOI~\citep{sarma2026funnelhoi}, ScriptHOI~\citep{nguyen2026scripthoi}, and HOIBlender~\citep{chen2026hoiblender}, with their reported backbone configurations. UniHOI by Yang et al.~\citep{yang2026unihoi} jointly learns detection and generation in a unified token space; UniHOI-s~\citep{cao2023unihoi} denotes the earlier BLIP-2-based detector. Table~\ref{tab:extended} also includes the three SFT+GRPO backbones reported by HOI-R1~\citep{chen2026hoir1}. MLLM distillation in CL-HOI~\citep{gao2025clhoi} and generative evaluation in HOI-MLLM~\citep{wu2026hoimllm} are presented with their respective metrics in Appendix~\ref{app:additional_settings}.

\paragraph{Tool-augmented interaction reasoning.}
ImagineAgent~\citep{yuan2026imagineagent} learns interaction reasoning through supervised fine-tuning and tool-augmented reinforcement learning, combining image cropping, retrieval, and generative modeling to augment its observations. Related visual-agent work learns to select heterogeneous visual experts across reasoning rounds~\citep{fan2026visharness}. \method{} uses the same frozen MLLM for visual exploration and relation verification, with geometric adjudication establishing the participant groundings shared across actions and roles.

\subsection{Common-Evaluator Comparison}
Tables~\ref{tab:common} and~\ref{tab:common-open} evaluate the released predictions of AgentHOI by Lei et al. and \method{} with identical matching and AP computation, complementing the source-reported comparisons above.
\begin{table}[!htb]
\caption{Common-evaluator comparison (AP, \%; differences in points, computed before rounding). AgentHOI (Lei et al.) uses GPT-4o and GroundingDINO; HarnessHOI uses Qwen3.5-9B.}
\label{tab:common}
\begin{center}
\small
\begin{tabular}{lrrr}\toprule
HICO-DET metric & AgentHOI (Lei et al.) & HarnessHOI & Difference\\\midrule
Default Full &29.68&33.20&3.52\\
Default Rare &40.49&41.63&1.14\\
Default Non-Rare &26.45&30.68&4.23\\
Known Object Full &33.71&36.85&3.14\\
Known Object Rare &43.39&44.20&0.81\\
Known Object Non-Rare &30.82&34.65&3.83\\\bottomrule
\end{tabular}
\end{center}
\end{table}
We evaluate the released AgentHOI~\citep{lei2026unleashing} predictions with the authors' default score completion, sigmoid, box exponent 0.3, and threshold 0.1.\footnote{AgentHOI release identifier: \texttt{5599af3794f6b6006be156da353e54f9ca64f486}.} The release covers 9,546 test images, including 23 without boxes. The remaining 112 official test images receive empty predictions. Both methods are evaluated over the complete test set using identical inclusive-IoU, matching, and sorting conventions.

\begin{table}[!htb]
\caption{HICO-DET class-partition analysis (Default AP, \%). The same HarnessHOI predictions are evaluated under four standard class partitions. Published and common-evaluator results of AgentHOI (Lei et al.) are shown separately. Bold identifies HarnessHOI's Unseen results.}
\label{tab:common-open}
\begin{center}
\small
\begin{tabular}{lrrrr}\toprule
& RF-UC & NF-UC & UO & UV\\\midrule
AgentHOI (Lei et al.), reported Unseen &38.64&28.04&33.70&29.85\\
AgentHOI (Lei et al.), common evaluator Unseen &39.11&28.56&33.97&30.13\\
\rowcolor{blue!6} HarnessHOI, Unseen &\textbf{40.15}&\textbf{33.64}&\textbf{35.45}&\textbf{33.87}\\
HarnessHOI, Seen &31.46&33.09&32.75&33.09\\
HarnessHOI, Full &33.20&33.20&33.20&33.20\\\bottomrule
\end{tabular}
\end{center}
\end{table}
\method{} consistently outperforms AgentHOI under the same evaluator, with improvements on both Rare and Non-Rare classes under Default and Known Object evaluation. The gains also extend to Unseen classes in all four open-vocabulary settings (Table~\ref{tab:common-open}).

\subsection{HICO-DET Interaction Classes}
\paragraph{Positive interactions and no interaction.}
Table~\ref{tab:hico_groups} reports results for the 520 positive-interaction and 80 no-interaction classes. Default mAP is 36.84\% and 9.55\%, respectively, giving a class-weighted Full mAP of 33.20\%. The corresponding Known Object results are 39.79\% and 17.73\%. Positive actions and no-interaction labels are assessed through separate visual questions.
\begin{table}[!htb]
\caption{HICO-DET interaction-type and frequency breakdowns. The upper and lower blocks each partition all 600 classes. GT and final TP counts are shared by Default and Known Object evaluation; AP is reported in percent.}
\label{tab:hico_groups}
\begin{center}
\small
\begin{tabular}{lrrrrr}\toprule
Class group & Classes & GT & TP & Default & Known Object\\
\midrule
Positive interactions & 520 & 29,110 & 19,965 & 36.84 & 39.79\\
No interaction & 80 & 4,295 & 2,196 & 9.55 & 17.73\\
\midrule
Rare & 138 & 981 & 598 & 41.63 & 44.20\\
Non-Rare & 462 & 32,424 & 21,563 & 30.68 & 34.65\\
\bottomrule
\end{tabular}
\end{center}
\end{table}

\paragraph{Frequency and evaluation domain.}
Rare and Non-Rare mAP average the classes within each frequency group, whereas coverage and true-positive counts aggregate individual annotations. Default and Known Object evaluation yield the same 22,161 true positives. Their AP differs because Known Object restricts the images considered for each class and therefore changes the false-positive distribution.

\subsection{V-COCO Actions and Semantic Roles}
\label{app:roles}
Table~\ref{tab:role_detail} reports S1/S2 AP and matching statistics for all 25 roles. Of the 12,423 physical-role and 3,969 missing-role annotations, the predictions match 9,728 and 3,854, respectively. S1 and S2 produce the same final true-positive counts in these results, while their missing-role criteria change the matching order and precision-recall curves. S2 exceeds S1 for every role, with small gaps for roles with few missing annotations such as surf, skateboard, and drink, and large gaps for roles with many missing annotations such as lay and eat/instr. The corresponding 24-role means, excluding \texttt{point}, are 46.09\% and 51.07\%; the 26-action agent mean is 75.58\%.
\begin{table}[!htb]
\caption{Complete V-COCO role results. GT and TP are annotation and final true-positive counts; physical and missing targets are separated. TP counts coincide under S1 and S2, while their score-ordered precision--recall curves differ. The last row sums counts and averages AP over 25 roles.}
\label{tab:role_detail}
\begin{center}
\small
\begin{tabular}{lrrrrrr}\toprule
& \multicolumn{2}{c}{GT} & \multicolumn{2}{c}{TP} & \multicolumn{2}{c}{AP (\%)}\\
\cmidrule(lr){2-3}\cmidrule(lr){4-5}\cmidrule(l){6-7}Action / role & Physical & Missing & Physical & Missing & S1 & S2\\
\midrule
hold / obj & 2,733 & 875 & 2,234 & 854 & 42.95 & 52.39\\
sit / instr & 1,089 & 827 & 747 & 786 & 38.93 & 47.88\\
ride / instr & 505 & 51 & 428 & 47 & 63.13 & 63.40\\
look / obj & 2,752 & 595 & 1,944 & 583 & 26.27 & 29.18\\
hit / instr & 315 & 34 & 286 & 34 & 43.63 & 46.52\\
hit / obj & 226 & 123 & 187 & 121 & 57.19 & 57.68\\
eat / obj & 334 & 187 & 271 & 181 & 46.80 & 54.71\\
eat / instr & 100 & 421 & 64 & 411 & 37.20 & 48.48\\
jump / instr & 402 & 233 & 371 & 228 & 62.40 & 66.18\\
lay / instr & 230 & 157 & 136 & 152 & 18.74 & 43.50\\
talk\_on\_phone / instr & 246 & 39 & 176 & 39 & 45.71 & 49.00\\
carry / obj & 433 & 39 & 347 & 36 & 26.17 & 27.29\\
throw / obj & 214 & 30 & 194 & 30 & 50.01 & 52.25\\
catch / obj & 201 & 45 & 187 & 44 & 32.42 & 32.70\\
cut / instr & 230 & 39 & 139 & 39 & 18.86 & 21.49\\
cut / obj & 203 & 66 & 159 & 64 & 49.20 & 58.92\\
work\_on\_computer / instr & 375 & 35 & 304 & 33 & 50.37 & 53.68\\
ski / instr & 360 & 64 & 249 & 64 & 39.58 & 46.82\\
surf / instr & 464 & 22 & 408 & 22 & 73.10 & 73.60\\
skateboard / instr & 402 & 15 & 378 & 15 & 80.28 & 80.78\\
drink / instr & 78 & 4 & 71 & 4 & 40.99 & 41.49\\
kick / obj & 163 & 17 & 131 & 17 & 46.78 & 47.28\\
point / instr & 21 & 10 & 13 & 10 & 10.05 & 16.97\\
read / obj & 86 & 25 & 71 & 24 & 43.83 & 58.55\\
snowboard / instr & 261 & 16 & 233 & 16 & 71.64 & 71.84\\
\midrule
Total / mean & 12,423 & 3,969 & 9,728 & 3,854 & 44.65 & 49.70\\
\bottomrule
\end{tabular}
\end{center}
\end{table}

\paragraph{Agent-level action recognition.}
Table~\ref{tab:agent_detail} reports all 26 agent actions, including the four without target roles. \method{} obtains 75.58\% mean agent AP by assessing each action on the localized person. For role-bearing actions, the same support is combined with target visibility and binding to produce role predictions.
\begin{table}[!htb]
\caption{V-COCO agent results for all 26 actions (AP, \%). Agent detection assesses action occurrence and person localization. The mean AP is 75.58\%.}
\label{tab:agent_detail}
\begin{center}
\small
\begin{tabular}{lrrrlrrr}\toprule
Action & GT & TP & AP & Action & GT & TP & AP\\
\midrule
hold & 3,608 & 3,533 & 85.42 & catch & 246 & 243 & 59.84\\
stand & 4,118 & 4,013 & 78.45 & cut & 269 & 260 & 83.41\\
sit & 1,916 & 1,838 & 84.87 & run & 687 & 681 & 80.02\\
ride & 556 & 523 & 80.74 & work\_on\_computer & 410 & 393 & 80.02\\
walk & 597 & 566 & 76.98 & ski & 424 & 420 & 88.57\\
look & 3,347 & 3,276 & 43.27 & surf & 486 & 484 & 92.99\\
hit & 349 & 344 & 81.31 & skateboard & 417 & 417 & 94.04\\
eat & 521 & 508 & 82.90 & smile & 1,415 & 1,392 & 84.45\\
jump & 635 & 626 & 84.40 & drink & 82 & 82 & 58.05\\
lay & 387 & 366 & 78.22 & kick & 180 & 180 & 65.29\\
talk\_on\_phone & 285 & 285 & 90.84 & point & 31 & 29 & 28.09\\
carry & 472 & 456 & 50.55 & read & 111 & 108 & 75.03\\
throw & 244 & 242 & 68.87 & snowboard & 277 & 274 & 88.36\\
\bottomrule
\end{tabular}
\end{center}
\end{table}

\subsection{Results under Related Evaluation Protocols}
\label{app:additional_settings}
We summarize related results on MLLM distillation, generative HOI prediction, and instruction-conditioned detection under their respective evaluation protocols.

\paragraph{Cross-level distillation results.}
Table~\ref{tab:clhoi-reported} presents CL-HOI under full supervision and MLLM distillation, including its ResNet-50 and Swin-L configurations. The reported metrics are HICO-DET Full mAP and the authors' V-COCO Role AP. The distillation setting transfers image-level interaction knowledge from the MLLM to an instance-level detector.

\begin{table}[!htb]
\caption{CL-HOI~\citep{gao2025clhoi} results under full supervision and MLLM distillation. Values follow the authors' HICO-DET Full mAP and V-COCO Role AP reporting (\%).}
\label{tab:clhoi-reported}
\begin{center}
\small
\renewcommand{\arraystretch}{1.08}
\begin{tabular}{llcc}
\toprule
Supervision & Backbone & HICO-DET Full mAP & V-COCO Role AP\\
\midrule
Full & R50+Transformer & 34.10 & 65.44\\
MLLM & R50+Transformer & 15.62 & 26.74\\
MLLM & Swin-L+Transformer & 17.55 & 36.63\\
\bottomrule
\end{tabular}
\end{center}
\end{table}

\paragraph{Generative interaction results.}
Table~\ref{tab:hoimllm-reported} reports HOI-MLLM with Qwen2.5-VL-3B using mean F1, precision, and recall. Its end-to-end configuration jointly predicts participants and interactions, the R50-DETR configuration supplies predicted object boxes, and Oracle supplies ground-truth boxes. These settings examine interaction reasoning with different sources of participant localization.

\begin{table}[!htb]
\caption{HOI-MLLM~\citep{wu2026hoimllm} results with Qwen2.5-VL-3B under its mean F1, precision, and recall evaluation (\%). End-to-end jointly predicts objects and interactions; R50-DETR supplies predicted object boxes; Oracle supplies ground-truth boxes.}
\label{tab:hoimllm-reported}
\begin{center}
\small
\renewcommand{\arraystretch}{1.08}
\begin{tabular}{lccccc}
\toprule
& \multicolumn{3}{c}{mF1} & mPrec & mRec\\
\cmidrule(lr){2-4}
Setting & Full & Rare & Non-Rare & Full & Full\\
\midrule
\multicolumn{6}{l}{\emph{HICO-DET}}\\
End-to-end & 24.28 & 25.75 & 23.27 & 26.72 & 26.26\\
R50-DETR & 30.32 & 27.12 & 31.68 & 35.71 & 36.21\\
Oracle & 50.61 & 51.16 & 49.90 & 51.37 & 56.47\\
\midrule
\multicolumn{6}{l}{\emph{V-COCO}}\\
End-to-end & 51.69 & -- & -- & 61.57 & 45.81\\
R50-DETR & 69.28 & -- & -- & 74.44 & 65.56\\
Oracle & 81.94 & -- & -- & 85.32 & 79.35\\
\bottomrule
\end{tabular}
\end{center}
\end{table}

\paragraph{Instruction-conditioned detection.}
HOID-R1 by Zhang et al.~\citep{zhang2025hoidr1} evaluates fine-grained, precise, and open-vocabulary instructions on a reannotated HICO-DET benchmark (Table~\ref{tab:hoidr1-reported}). Its Seen and Unseen groups refer to image subsets. Human and object mIoU, action accuracy, and mAP measure localization and interaction recognition under these instruction settings.

\begin{table}[!htb]
\caption{HOID-R1~\citep{zhang2025hoidr1} results on its reannotated HICO-DET benchmark with fine-grained, precise, and open-vocabulary instructions. Seen and Unseen refer to image subsets. H-mIoU, O-mIoU, and A-Acc measure human localization, object localization, and action accuracy; mAP is reported in percent.}
\label{tab:hoidr1-reported}
\begin{center}
\small
\renewcommand{\arraystretch}{1.08}
\begin{tabular}{llcccc}
\toprule
Annotation & Images & H-mIoU & O-mIoU & A-Acc & mAP\\
\midrule
Fine-grained & Seen & $0.72\pm0.026$ & $0.73\pm0.028$ & $0.79\pm0.031$ & $55.07\pm2.11$\\
& Unseen & $0.71\pm0.025$ & $0.70\pm0.027$ & $0.75\pm0.030$ & $52.98\pm2.18$\\
Precise & Seen & $0.70\pm0.025$ & $0.72\pm0.027$ & $0.76\pm0.030$ & $53.81\pm2.13$\\
& Unseen & $0.71\pm0.026$ & $0.69\pm0.025$ & $0.72\pm0.029$ & $51.19\pm2.21$\\
Open-vocabulary & Seen & $0.69\pm0.024$ & $0.71\pm0.027$ & $0.75\pm0.032$ & $51.68\pm2.36$\\
& Unseen & $0.68\pm0.023$ & $0.68\pm0.022$ & $0.72\pm0.030$ & $50.03\pm2.14$\\
\bottomrule
\end{tabular}
\end{center}
\end{table}

\FloatBarrier
\section{Harness Design Analysis}
\label{app:ablations}
\label{app:guidance}
We examine how interaction-guided perception, geometric adjudication, and grounded interaction reasoning contribute to HOI detection. Table~\ref{tab:comparison_design} summarizes the experimental conditions used to evaluate each component.
\begin{table}[!htb]
\caption{Design of the controlled comparisons on the complete official test sets. Each comparison varies the indicated aspect of inference while retaining the common conditions listed below.}
\label{tab:comparison_design}
\begin{center}
\small
\begin{tabularx}{\linewidth}{@{}l>{\raggedright\arraybackslash}X>{\raggedright\arraybackslash}X@{}}\toprule
Comparison & Aspect examined & Common conditions\\\midrule
Overall inference & Direct generation, initial-instance baseline, and the complete harness & Backbone, full-image processing, public vocabulary, and evaluation\\
Perceptual components & Discovery participants, local observation, and geometric selection & Common initial localization; relation questions evaluated on each resulting participant set\\
Interaction guidance & Discovery cues, observation location, geometry criterion, or proposal context & Inputs and budgets specified for each replacement in Appendix~\ref{app:guidance}\\
Relation candidates & Pair sources and labels on the proposed pairs & Selected participants, image views, and relation supports\\
Role verification & Joint questions and compositional confidence & Physical and NULL candidates and participant geometry; joint questions use the corresponding binding or role-state views\\
Repeated generation & Response count and duplicate fusion & Ordered response prefixes and original generation scores\\
Generation stability & Generation seed & Backbone, inference policy, and evaluation population\\
Backbone generality & Different frozen MLLM backbones & Harness configuration, public vocabulary, seed 66, and evaluation\\\bottomrule
\end{tabularx}
\end{center}
\end{table}

\subsection{Effectiveness of the Complete Harness}
Table~\ref{tab:harness_analysis} compares direct generation, the initial-instance baseline, and the complete harness under the scoring rules in Appendix~\ref{app:baselines}. Grounded verification over the initial participants substantially improves detection over direct generation. Interaction-guided perception and geometric adjudication further improve results on both benchmarks.

\subsection{Complete Ablation Results}
\label{app:complete_ablations}
Tables~\ref{tab:full_ablation} and~\ref{tab:full_ablation_vcoco} report the complete HICO-DET and V-COCO metrics for the ablations examined in this section.
\begin{table}[!htb]
\caption{Complete HICO-DET harness comparisons under Default and Known Object evaluation (mAP, \%). All variants use the same backbone and evaluator.}
\label{tab:full_ablation}
\begin{center}
\small
\begin{tabular}{lrrrrrr}
\toprule
& \multicolumn{3}{c}{Default} & \multicolumn{3}{c}{Known Object}\\
\cmidrule(lr){2-4}\cmidrule(l){5-7}
Variant & Full & Rare & Non-Rare & Full & Rare & Non-Rare\\
\midrule
\method{} & 33.20 & 41.63 & 30.68 & 36.85 & 44.20 & 34.65\\
Direct HOI generation & 15.62 & 22.13 & 13.68 & 17.27 & 23.25 & 15.49\\
Initial-instance baseline & 31.45 & 39.79 & 28.96 & 35.05 & 41.30 & 33.18\\
w/o discovered participants & 32.55 & 40.91 & 30.05 & 36.15 & 42.42 & 34.28\\
w/o targeted observation & 32.80 & 41.24 & 30.28 & 36.39 & 42.62 & 34.53\\
First-candidate geometry & 32.95 & 41.30 & 30.46 & 36.55 & 42.89 & 34.65\\
Proposed pairs, proposed labels & 18.35 & 21.08 & 17.53 & 20.39 & 22.13 & 19.87\\
Proposed pairs, all compatible labels & 29.85 & 34.19 & 28.55 & 33.09 & 36.03 & 32.21\\
Visual pair assessment only & 32.30 & 40.58 & 29.83 & 35.85 & 41.90 & 34.04\\
Discovery and observation only & 28.09 & 32.41 & 26.80 & 31.20 & 34.26 & 30.28\\
Binary relation readout & 33.05 & 41.37 & 30.56 & 36.70 & 42.89 & 34.85\\
Repeated participant localization & 32.45 & 40.84 & 29.95 & 36.05 & 42.36 & 34.16\\
Non-directed observation & 32.70 & 41.17 & 30.17 & 36.30 & 42.49 & 34.45\\
Relation-conditioned adjudication & 32.79 & 41.11 & 30.30 & 36.38 & 42.62 & 34.51\\
Proposal-conditioned verification & 32.50 & 40.84 & 30.01 & 36.09 & 42.36 & 34.22\\
\bottomrule
\end{tabular}
\end{center}
\end{table}

\begin{table}[!htb]
\caption{Complete V-COCO harness comparisons corresponding to Table~\ref{tab:full_ablation}, reporting role AP over 25 roles under S1 and S2 (\%).}
\label{tab:full_ablation_vcoco}
\begin{center}
\small
\begin{tabular}{lrr}
\toprule
Variant & S1 & S2\\
\midrule
\method{} & 44.65 & 49.70\\
Direct HOI generation & 16.41 & 21.20\\
Initial-instance baseline & 42.90 & 47.20\\
w/o discovered participants & 44.02 & 48.93\\
w/o targeted observation & 44.33 & 49.21\\
First-candidate geometry & 44.38 & 49.48\\
Proposed pairs, proposed labels & 36.43 & 39.81\\
Proposed pairs, all compatible labels & 42.27 & 47.14\\
Visual pair assessment only & 43.23 & 47.77\\
Discovery and observation only & 42.48 & 47.19\\
Binary relation readout & 44.52 & 49.57\\
w/o action factor & 5.83 & 6.57\\
w/o visibility factor & 44.13 & 49.55\\
w/o NULL predictions & 39.20 & 49.43\\
Joint action-and-role question & 41.10 & 46.30\\
Repeated participant localization & 43.87 & 48.88\\
Non-directed observation & 44.07 & 49.08\\
Relation-conditioned adjudication & 44.18 & 49.16\\
Proposal-conditioned verification & 43.92 & 48.84\\
\bottomrule
\end{tabular}
\end{center}
\end{table}

\begin{table}[!htb]
\caption{Complete HICO-DET class-partition ablations (Default mAP, \%). All partitions use the same predictions for each variant; Full values appear in Table~\ref{tab:full_ablation}.}
\label{tab:ablation_open}
\begin{center}
\small
\begin{tabular}{lrrrr}\toprule
& \multicolumn{2}{c}{RF-UC} & \multicolumn{2}{c}{NF-UC}\\
\cmidrule(lr){2-3}\cmidrule(l){4-5}Variant & Unseen & Seen & Unseen & Seen\\
\midrule
\method{} & 40.15 & 31.46 & 33.64 & 33.09\\
Direct generation & 21.82 & 14.07 & 12.80 & 16.33\\
Initial-instance baseline & 38.33 & 29.73 & 31.59 & 31.42\\
w/o discovered participants & 39.17 & 30.89 & 33.26 & 32.37\\
w/o targeted observation & 40.00 & 31.00 & 32.88 & 32.78\\
First-candidate geometry & 39.85 & 31.23 & 33.48 & 32.82\\
Proposed pairs, proposed labels & 23.33 & 17.10 & 18.41 & 18.33\\
Proposed pairs, all compatible labels & 36.52 & 28.18 & 30.61 & 29.66\\
Visual pair assessment only & 39.39 & 30.53 & 32.50 & 32.25\\
Discovery and observation only & 35.08 & 26.34 & 28.71 & 27.94\\
Binary relation readout & 40.00 & 31.31 & 33.48 & 32.94\\
\bottomrule
\end{tabular}

\par\medskip
\begin{tabular}{lrrrr}\toprule
& \multicolumn{2}{c}{UO} & \multicolumn{2}{c}{UV}\\
\cmidrule(lr){2-3}\cmidrule(l){4-5}Variant & Unseen & Seen & Unseen & Seen\\
\midrule
\method{} & 35.45 & 32.75 & 33.87 & 33.09\\
Direct generation & 17.09 & 15.33 & 19.26 & 15.03\\
Initial-instance baseline & 33.91 & 30.96 & 32.03 & 31.36\\
w/o discovered participants & 34.55 & 32.15 & 33.33 & 32.42\\
w/o targeted observation & 35.27 & 32.31 & 33.12 & 32.75\\
First-candidate geometry & 34.91 & 32.56 & 33.33 & 32.89\\
Proposed pairs, proposed labels & 20.73 & 17.87 & 18.61 & 18.31\\
Proposed pairs, all compatible labels & 31.91 & 29.44 & 30.63 & 29.72\\
Visual pair assessment only & 34.82 & 31.80 & 32.79 & 32.22\\
Discovery and observation only & 30.27 & 27.65 & 29.11 & 27.92\\
Binary relation readout & 35.18 & 32.62 & 33.66 & 32.95\\
\bottomrule
\end{tabular}
\end{center}
\end{table}

Table~\ref{tab:ablation_open} reports the complete ablations under RF-UC, NF-UC, UO, and UV. Each variant is evaluated across the four class partitions using its predictions from regular evaluation. \method{} obtains Unseen mAP of 40.15\%, 33.64\%, 35.45\%, and 33.87\%, respectively. Its advantage over direct generation extends across combinations, object categories, and actions. The pair and label ablations examine the contribution of candidate selection and compatible-action verification within these class groups.

\subsection{Interaction-Guided Perception}
\label{app:coverage}
Interaction-guided perception uses action and role hypotheses to discover relevant participants and select regions for closer observation. We first evaluate participant discovery and local observation, then compare them with repeated localization and non-directed observation to examine the contribution of interaction guidance.

\paragraph{Participant discovery and targeted observation.}
The discovered-participant ablation excludes identities introduced by discovery and performs observation and relation assessment over the remaining participants. The observation ablation retains global discovery and its single localization per instance, omitting local observation and the corresponding geometric adjudication. These component configurations are reported in Table~\ref{tab:harness_analysis}, with full metrics in Tables~\ref{tab:full_ablation} and~\ref{tab:full_ablation_vcoco}.

\paragraph{Guidance for participant discovery.}
The repeated-localization control receives the same full image, initial participant identities, and designated human groups as interaction-guided discovery. Its question directly requests additional visible participants associated with those humans. Both conditions use the same generation budget and the observation requests produced by interaction-guided discovery, followed by the same observation, geometric adjudication, and relation verification. Interaction-guided discovery improves detection on both benchmarks (Table~\ref{tab:harness_controls}), supporting the use of interaction semantics to direct participant search.

\paragraph{Selection of local observations.}
To evaluate region selection, the non-directed control retains the discovery output, target descriptions, observation count, and crop dimensions. Each crop center is sampled uniformly from valid image positions using a fixed seed. Both conditions use the same image processing and localization instructions, followed by geometric and relational assessment on their acquired observations. The gains from targeted observation in Table~\ref{tab:harness_controls} demonstrate the value of selecting regions according to the interaction hypothesis.

\begin{figure}[!htb]
\centering
\includegraphics[width=\linewidth]{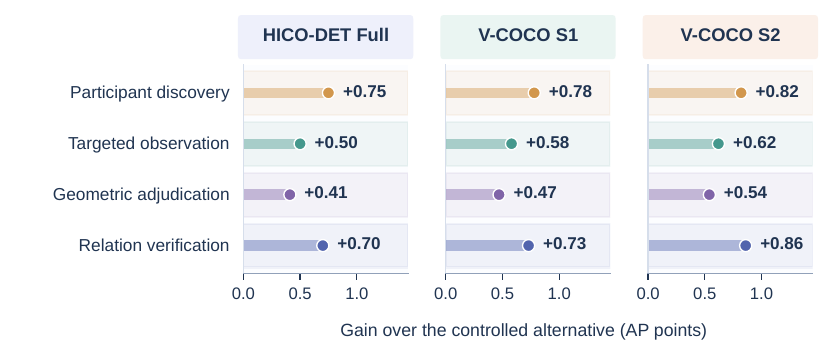}
\caption{\textbf{Interaction guidance and visual judgment.} Each point reports the AP gain of the complete harness over the corresponding controlled alternative in Table~\ref{tab:harness_controls}: repeated localization, non-directed observation, relation-conditioned adjudication, and proposal-conditioned verification. The three panels use the same scale.}
\label{fig:guidance_controls}
\end{figure}

\paragraph{Stage-wise participant coverage.}
We measure grounding coverage and relation-candidate coverage to evaluate participant acquisition at successive stages. Grounding coverage requires participant boxes that match an annotation at IoU 0.5, while candidate coverage additionally requires the corresponding action or role label. Each annotation is counted once.

Initial localization and discovery are evaluated on their respective participant sets. The all-localization condition uses the accumulated grounding candidates, and the selected-grounding condition uses one adjudicated candidate per identity. Physical interactions require matching human and target instances with distinct identities; HICO-DET also requires the target category to match. Missing roles require a matching person, together with the appropriate NULL prediction for candidate coverage.

\begin{table}[!htb]
\caption{Coverage of annotated interactions through participant construction and relation assessment. Grounding coverage requires matching participant boxes at IoU 0.5; the final candidate domain additionally requires the action or role label. V-COCO includes 3,969 missing-role annotations, for which participant grounding requires the person alone.}
\label{tab:coverage_stages}
\begin{center}
\small
\begin{tabular}{lrr}\toprule
Participant and relation domain & HICO-DET & V-COCO\\\midrule
Initial participant groundings & 22,314 & 14,248\\
After interaction discovery & 22,885 & 14,465\\
All localization candidates & 22,897 & 14,488\\
Selected shared groundings & 22,895 & 14,473\\
Final action or role candidates & 22,232 & 13,628\\\midrule
Annotated interactions or roles & 33,405 & 16,392\\\bottomrule
\end{tabular}
\end{center}
\end{table}

As shown in Table~\ref{tab:coverage_stages}, discovery adds matching groundings for 571 HICO-DET interactions and 217 V-COCO roles. Local observation extends coverage by a further 12 and 23 annotations, respectively. The final selected groundings cover 22,895 HICO-DET interactions and 14,473 V-COCO roles, providing the participant pairs for subsequent verification.

\paragraph{Interaction configurations.}
We group the interaction annotations by participant and action structure. The multiple-human-box group contains images with at least two distinct annotated human boxes. The HICO-DET multiple-target group contains annotations whose action has at least two distinct physical targets within the image. For V-COCO, the corresponding multiple-target-state group treats each distinct physical target and NULL as target states, while reporting physical-role annotations. Small targets occupy at most 1\% of the image area. The multiple-action group contains human-target pairs annotated with at least two actions, including NULL targets in V-COCO. Groups may overlap, and their sizes count annotations.

Missing-role annotations form a separate V-COCO group and contribute to the full totals. Their grounding and candidate coverage follow the person and NULL criteria above. The small-target group includes physical targets alone.

\begin{table}[!htb]
\caption{Interaction coverage in the final action or role candidate domain. A ground-truth annotation is covered when at least one candidate has the required participant localization and interaction label. The initial-instance baseline verifies relations over the initially localized participants. Small targets occupy at most 1\% of the image area. Scene groups can overlap; missing-role coverage requires the person and the corresponding NULL role candidate.}
\label{tab:coverage}
\begin{center}
\small
\begin{tabular}{lrrr}\toprule
Interaction group & Annotations & \shortstack{Initial-instance\\baseline} & \method{}\\\midrule
\multicolumn{4}{l}{\emph{HICO-DET}}\\
All interactions & 33,405 & 21,354 & 22,232\\
Multiple annotated human boxes & 30,441 & 19,498 & 20,288\\
Multiple targets for the same action & 11,549 & 6,172 & 6,437\\
Small targets & 5,960 & 2,184 & 2,356\\
\midrule
\multicolumn{4}{l}{\emph{V-COCO}}\\
All roles & 16,392 & 12,703 & 13,628\\
Multiple annotated human boxes & 8,183 & 5,909 & 6,553\\
Multiple target states per action & 4,468 & 3,026 & 3,375\\
Small targets & 4,343 & 2,761 & 3,208\\
Multiple actions per human--target state & 10,072 & 8,375 & 8,909\\
Missing role targets & 3,969 & 3,863 & 3,863\\
\bottomrule
\end{tabular}
\end{center}
\end{table}

\begin{table}[!htb]
\caption{Stage-wise coverage within complex-scene groups. Initial, Discovery, and Observed count annotations with matching participant geometry in the cumulative candidate sets; Selected uses adjudicated boxes; Relations additionally requires the correct action or role candidate. Counts use IoU 0.5 and overlapping annotation-defined groups.}
\label{tab:group_stages}
\begin{center}
\small
\begin{tabular}{lrrrrrr}\toprule
Group & GT & Initial & Discovery & Observed & Selected & Relations\\
\midrule
\multicolumn{7}{l}{\emph{HICO-DET}}\\
All interactions & 33,405 & 22,314 & 22,885 & 22,897 & 22,895 & 22,232\\
\shortstack[l]{Multiple annotated\\human boxes} & 30,441 & 20,323 & 20,848 & 20,858 & 20,856 & 20,288\\
\shortstack[l]{Multiple targets\\per action} & 11,549 & 6,578 & 6,728 & 6,733 & 6,731 & 6,437\\
Small targets & 5,960 & 2,387 & 2,489 & 2,493 & 2,492 & 2,356\\
\midrule
\multicolumn{7}{l}{\emph{V-COCO}}\\
All roles & 16,392 & 14,248 & 14,465 & 14,488 & 14,473 & 13,628\\
Missing roles & 3,969 & 3,863 & 3,863 & 3,863 & 3,863 & 3,863\\
\shortstack[l]{Multiple actions per\\human--target state} & 10,072 & 8,939 & 9,086 & 9,102 & 9,088 & 8,909\\
\shortstack[l]{Multiple annotated\\human boxes} & 8,183 & 7,100 & 7,213 & 7,219 & 7,214 & 6,553\\
\shortstack[l]{Multiple target states\\per action} & 4,468 & 3,663 & 3,756 & 3,761 & 3,759 & 3,375\\
Small targets & 4,343 & 3,407 & 3,518 & 3,529 & 3,522 & 3,208\\
\bottomrule
\end{tabular}
\end{center}
\end{table}

\begin{figure}[!htb]
\centering
\includegraphics[width=\linewidth]{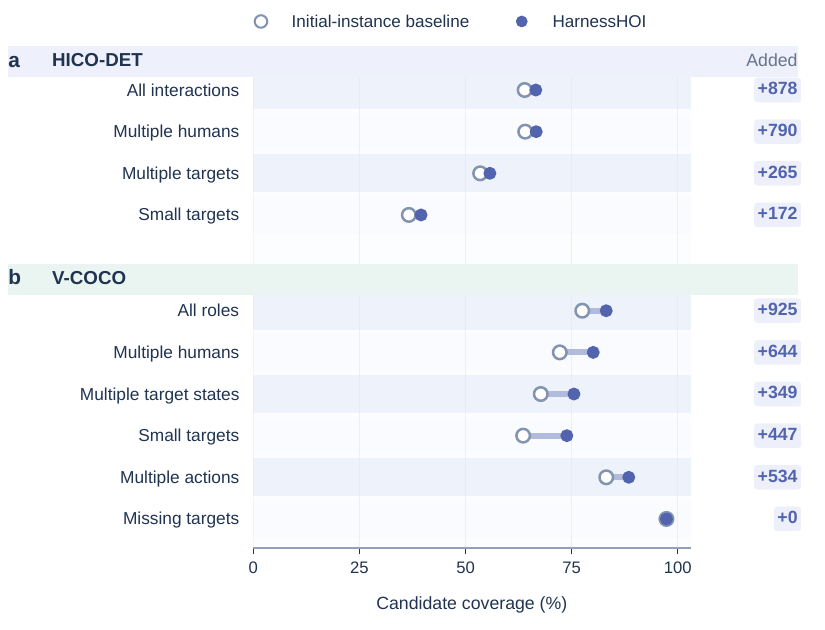}
\caption{\textbf{Candidate coverage across interaction configurations.} Open and filled markers compare the initial-instance baseline and the complete harness. Coverage is normalized by the number of annotations in each group; the right column reports additional covered annotations. Groups can overlap. The underlying counts and group definitions appear in Table~\ref{tab:coverage}.}
\label{fig:coverage_structure}
\end{figure}

\paragraph{Coverage in complex scenes.}
The full harness covers 878 additional HICO-DET interactions and 925 additional V-COCO roles over the initial-instance baseline (Table~\ref{tab:coverage} and Figure~\ref{fig:coverage_structure}). These gains include 172 and 447 annotations with small targets, respectively, and 534 V-COCO roles involving multiple actions. Table~\ref{tab:group_stages} reports the corresponding stage-wise coverage. For V-COCO small targets, coverage increases from 3,407 after initial localization to 3,518 after discovery and 3,529 after observation, with 3,208 annotations represented in the final relation candidates.

The gains on small targets demonstrate the effectiveness of interaction-guided perception in recovering visually inconspicuous participants. Improvements on multiple-action pairs further show that the acquired groundings support several interactions involving the same entities.

\subsection{Relation-Agnostic Geometric Adjudication}
We evaluate geometric adjudication by retaining discovery and observation and replacing visual selection with the first box associated with each identity (Table~\ref{tab:harness_analysis}). This comparison examines the contribution of selecting a shared participant grounding from the acquired localization candidates.

On HICO-DET and V-COCO, 479 and 424 identities adopt an observation-derived grounding, respectively. The selected grounding is shared across all relations involving the participant. The improvement over first-candidate geometry in Table~\ref{tab:harness_analysis} confirms the effectiveness of visual adjudication for interaction grounding.

\paragraph{Criterion for shared grounding.}

Relation-conditioned and relation-agnostic adjudication receive identical candidate boxes and image views for each instance. Relation-conditioned adjudication adds the earliest discovery or observation association involving that instance and asks which candidate best localizes the participant in that interaction. Relation-agnostic adjudication uses the identity, category, and visible extent criterion in Section~\ref{sec_geometric_adjudication}. Instances without a recorded association use the common identity-based criterion. Both conditions select one shared grounding per instance and apply the same subsequent relation verification.

With candidate boxes and image views fixed, relation-agnostic adjudication improves detection on both benchmarks (Table~\ref{tab:harness_controls}). The results support selecting shared groundings according to participant identity, category, and visible extent, independently of the proposed relation.

\subsection{Grounded Interaction Reasoning}
\label{app:reasoning}
We examine grounded interaction reasoning through complementary pair selection, multi-label verification, and compositional semantic-role assessment. These comparisons evaluate which participant pairs and relations are considered and how their visual evidence determines the final predictions.
\paragraph{Context for relation verification.}
The proposal-conditioned control appends the original discovery proposals associated with the queried participants to the final action, agent, role-state, and binding questions. The harness condition uses the selected participants, public relation definitions, and image evidence. Pair retention, candidate relations, boxes, views, answer labels, and score composition are identical. Both conditions assess the full candidate domain, including relations absent from the discovery proposals. Grounded verification improves detection over proposal-conditioned questions on these identical candidates (Table~\ref{tab:harness_controls}), demonstrating the benefit of assessing relations directly from the acquired visual evidence.

\begin{table}[!htb]
\caption{Analysis of grounded interaction reasoning. HICO-DET reports Default Full mAP; V-COCO reports S1/S2 role AP (\%). Panel (a) varies pair retention with fixed visual assessments; panel (b) varies the retained labels on identical proposed pairs with fixed scores; panel (c) compares joint and compositional verification on the same role candidates, with each joint question receiving the views of its corresponding binding or role-state question. Best results within each panel are bold.}
\label{tab:harness_reasoning}
\begin{center}
\small
\begin{tabular}{lrrr}\toprule
& HICO-DET & \multicolumn{2}{c}{V-COCO}\\
\cmidrule(lr){2-2}\cmidrule(l){3-4}
Variant & Full & S1 & S2\\\midrule
\multicolumn{4}{l}{\emph{(a) Sources of relation candidates}}\\
Discovery and observation & 28.09 & 42.48 & 47.19\\
Visual pair assessment & 32.30 & 43.23 & 47.77\\
Combined & \textbf{33.20} & \textbf{44.65} & \textbf{49.70}\\
\midrule
\multicolumn{4}{l}{\emph{(b) Labels retained on identical proposed pairs}}\\
Proposed labels & 18.35 & 36.43 & 39.81\\
All compatible labels & \textbf{29.85} & \textbf{42.27} & \textbf{47.14}\\
\midrule
\multicolumn{4}{l}{\emph{(c) Semantic-role verification}}\\
Joint action-and-role question & -- & 41.10 & 46.30\\
Compositional verification & -- & \textbf{44.65} & \textbf{49.70}\\
\bottomrule
\end{tabular}
\end{center}
\end{table}

\paragraph{Complementary relation candidates.}
Table~\ref{tab:harness_reasoning}(a) compares visual pair assessment, discovery and observation associations $\mathcal{P}_D\cup\mathcal{P}_O$, and their union $\mathcal{K}$. Each configuration retains the original relation scores on its selected pairs, together with HICO-DET no-interaction and V-COCO agent and NULL predictions. Combining the two sources achieves the best results on both benchmarks, improving HICO-DET mAP by 5.11 points over discovery and observation associations and V-COCO AP by 1.42/1.93 points over visual pair assessment. The gains demonstrate that associations established during perception complement the relations identified by visual pair assessment.

\paragraph{Coverage and selective verification.}
The final candidates cover 22,232 HICO-DET interactions and 13,628 V-COCO roles, retaining 97.1\% and 94.2\% of the annotations covered by the selected groundings. Positive-action and role-binding verification uses 39,791 of 416,280 human-entity pairs on HICO-DET and 17,804 of 251,376 on V-COCO. Thus, pair selection reduces the respective domains by 90.4\% and 92.9\% while retaining high coverage of the grounded interactions.

Coverage measures the existence of matching candidates. Detection evaluation additionally assigns each prediction to its best-matching annotation and removes duplicates in score order. A prediction overlapping several annotations contributes at most one true positive, so final matching counts can differ from candidate coverage. The ranking bounds in Appendix~\ref{app:ranking} follow the evaluator's assignment rule.

\paragraph{Multi-label interaction verification.}
We evaluate label expansion on the pairs represented by explicitly proposed and scored relations, denoted by $\mathcal{P}_{\mathrm{prop}}$. The proposed-label condition retains these relations; the compatible-label condition includes all assessed category-compatible positive actions in HICO-DET or all 25 action-role categories in V-COCO on the same pairs. An observation association enters $\mathcal{P}_O$, whereas membership in $\mathcal{P}_{\mathrm{prop}}$ requires a scored relation proposal. Geometry, image evidence, scores, and all no-interaction, agent, and NULL predictions remain fixed.

As shown in Table~\ref{tab:harness_reasoning}(b), label expansion increases HICO-DET mAP from 18.35\% to 29.85\% and V-COCO AP from 36.43\%/39.81\% to 42.27\%/47.14\%. The gains demonstrate the importance of verifying multiple compatible relations over shared participants, extending prediction beyond the labels proposed during discovery.

\paragraph{Answer support and role factors.}
Role visibility and binding are assessed conditionally on the action, while $q_A$ supplies the evidence that the person performs it. For a given person, action, and role, physical candidates share $q_Aq_V$ and are ordered by their binding support $q_B$. Across people and images, action and visibility also contribute to the confidence ranking used for role AP. Table~\ref{tab:full_ablation_vcoco} examines these factors and the treatment of unresolved and missing targets.

The binary-readout ablation removes unknown support and renormalizes false/true or missing/visible within each final question, preserving pair retention and score composition. The action-factor ablation uses $q_Vq_B$ for physical targets and $q_M$ for NULL targets. The visibility-factor ablation uses $q_Aq_B$ for physical targets and preserves NULL scores. Both retain the original agent predictions. The NULL ablation removes missing-target predictions while retaining physical-role and agent outputs. All four conditions are derived from the full method's answer supports.

\paragraph{Compositional role verification.}
We compare compositional verification with a joint action-and-role question on identical participants, geometry, and role candidates. Each physical-target joint question uses the same full-image and crop inputs as its corresponding binding question; each NULL joint question uses those of its corresponding role-state question. The joint question evaluates action occurrence together with the target's role; its NULL formulation evaluates action occurrence together with the absence of a visible role target. True-label support directly scores each candidate, with agent predictions unchanged.

Compositional verification improves V-COCO S1/S2 AP from 41.10\%/46.30\% to 44.65\%/49.70\% (Table~\ref{tab:harness_reasoning}(c)), with gains on 15 of 25 roles in S1 and 20 of 25 in S2. The final true-positive counts remain unchanged, showing that the improvement comes from more accurate ranking of role predictions. Image-paired bootstrap with 1,000 resamples gives 95\% percentile intervals of [3.02, 4.10] and [2.85, 3.96] points for the S1 and S2 gains.

\begin{table}[!htb]
\caption{Joint and compositional role verification on identical candidates (AP, \%). $\Delta$ denotes compositional minus joint AP in points. Both formulations have the same final true-positive count for every role.}
\label{tab:joint_role_detail}
\begin{center}
\small
\begin{tabular}{lrrrrrr}\toprule
& \multicolumn{3}{c}{S1} & \multicolumn{3}{c}{S2}\\
\cmidrule(lr){2-4}\cmidrule(l){5-7}Action / role & Joint & Compositional & $\Delta$ & Joint & Compositional & $\Delta$\\
\midrule
hold / obj & 46.08 & 42.95 & -3.13 & 52.08 & 52.39 & +0.31\\
sit / instr & 38.97 & 38.93 & -0.04 & 45.02 & 47.88 & +2.86\\
ride / instr & 56.50 & 63.13 & +6.63 & 57.05 & 63.40 & +6.35\\
look / obj & 31.49 & 26.27 & -5.22 & 35.52 & 29.18 & -6.34\\
hit / instr & 27.68 & 43.63 & +15.95 & 30.64 & 46.52 & +15.88\\
hit / obj & 30.73 & 57.19 & +26.46 & 31.48 & 57.68 & +26.20\\
eat / obj & 45.98 & 46.80 & +0.82 & 54.14 & 54.71 & +0.57\\
eat / instr & 10.91 & 37.20 & +26.29 & 53.88 & 48.48 & -5.40\\
jump / instr & 57.87 & 62.40 & +4.53 & 64.18 & 66.18 & +2.00\\
lay / instr & 22.76 & 18.74 & -4.02 & 40.67 & 43.50 & +2.83\\
talk\_on\_phone / instr & 46.79 & 45.71 & -1.08 & 49.43 & 49.00 & -0.43\\
carry / obj & 25.85 & 26.17 & +0.32 & 26.31 & 27.29 & +0.98\\
throw / obj & 46.92 & 50.01 & +3.09 & 48.35 & 52.25 & +3.90\\
catch / obj & 31.30 & 32.42 & +1.12 & 32.69 & 32.70 & +0.01\\
cut / instr & 7.89 & 18.86 & +10.97 & 9.43 & 21.49 & +12.06\\
cut / obj & 49.12 & 49.20 & +0.08 & 56.87 & 58.92 & +2.05\\
work\_on\_computer / instr & 50.41 & 50.37 & -0.04 & 52.07 & 53.68 & +1.61\\
ski / instr & 39.84 & 39.58 & -0.26 & 43.22 & 46.82 & +3.60\\
surf / instr & 67.33 & 73.10 & +5.77 & 67.90 & 73.60 & +5.70\\
skateboard / instr & 81.07 & 80.28 & -0.79 & 81.51 & 80.78 & -0.73\\
drink / instr & 41.58 & 40.99 & -0.59 & 42.14 & 41.49 & -0.65\\
kick / obj & 45.16 & 46.78 & +1.62 & 45.61 & 47.28 & +1.67\\
point / instr & 7.13 & 10.05 & +2.92 & 11.40 & 16.97 & +5.57\\
read / obj & 47.62 & 43.83 & -3.79 & 54.86 & 58.55 & +3.69\\
snowboard / instr & 70.49 & 71.64 & +1.15 & 71.05 & 71.84 & +0.79\\
\midrule
Mean & 41.10 & 44.65 & +3.55 & 46.30 & 49.70 & +3.40\\
\bottomrule
\end{tabular}
\end{center}
\end{table}

\begin{figure}[!htb]
\centering
\includegraphics[width=\linewidth]{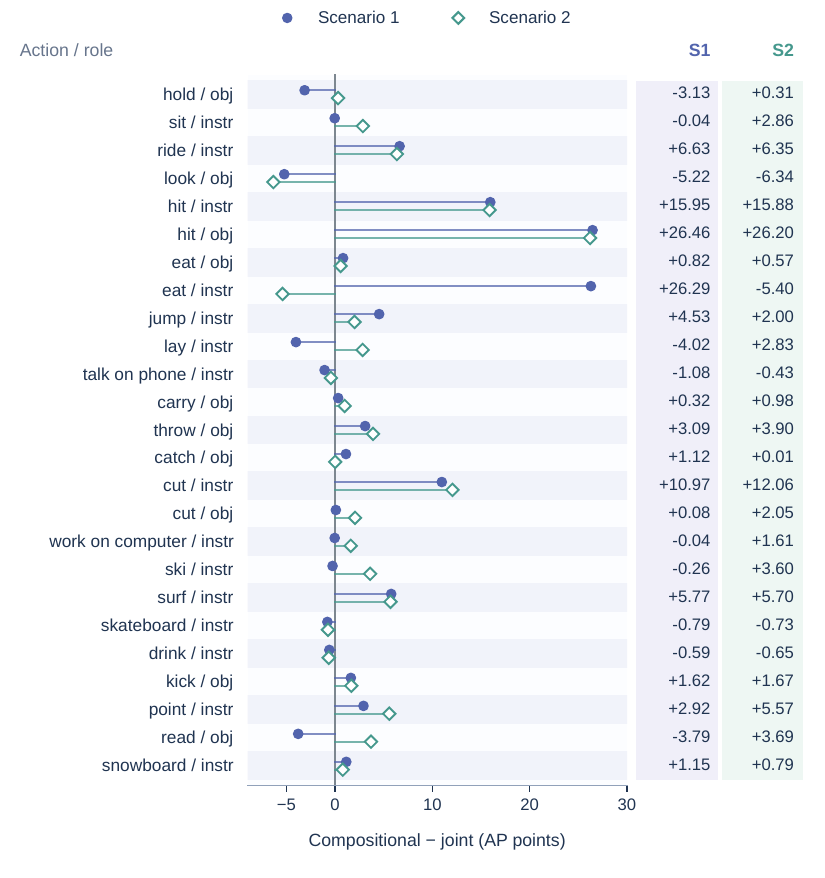}
\caption{\textbf{Compositional verification across semantic roles.} AP differences between compositional and joint verification for all 25 V-COCO roles on the same candidates. Each joint question receives the views of its corresponding binding or role-state question. Filled circles and open diamonds denote S1 and S2 on a common scale. Values in the right columns correspond to Table~\ref{tab:joint_role_detail}.}
\label{fig:role_composition}
\end{figure}

\paragraph{Role-specific confidence composition.}
Table~\ref{tab:joint_role_detail} and Figure~\ref{fig:role_composition} report the comparison for all 25 roles. Compositional verification improves both roles of \texttt{hit} and the instrument role of \texttt{cut} under S1 and S2, with a 26.29-point gain for \texttt{eat/instr} under S1. The gains demonstrate the effectiveness of combining shared action evidence with the visibility and binding evidence of each role target.

\paragraph{Image-paired resampling.}
For the joint and compositional role comparison, we draw 1,000 paired bootstrap samples of the test images with replacement~\citep{efron1979bootstrap}, using seed 20260920. Each sample contains the original number of images and assigns identical image multiplicities to both methods' predictions and annotations. We recompute AP under the original matching rules and class denominator, assigning zero to classes absent from a sample. The 2.5th and 97.5th percentiles of the paired AP differences form the confidence interval for each scenario.

\paragraph{Missing role targets.}
Explicit missing-role assessment represents actions whose targets are not visible in the image. Removing NULL predictions reduces S1 AP from 44.65\% to 39.20\%, while the complete method obtains 49.70\% under S2 (Table~\ref{tab:full_ablation_vcoco}). This comparison demonstrates the importance of distinguishing a missing role target from an absent action.

\subsection{Backbone Generality}
\label{app:transfer}
We apply the identical harness configuration, public vocabulary, and evaluator to Qwen2.5-VL-7B~\citep{bai2025qwen25vl} and InternVL3-8B~\citep{zhu2025internvl3} on the complete test sets under generation seed 66. Table~\ref{tab:backbone_transfer} reports the complete-harness results for each backbone.
\begin{table}[!htb]
\caption{Backbone generality on the complete test sets under generation seed 66. Each backbone uses the identical harness configuration, public vocabulary, and evaluator. HICO-DET reports Default Full mAP; V-COCO reports S1/S2 role AP (\%).}
\label{tab:backbone_transfer}
\begin{center}
\small
\begin{tabular}{lrrr}\toprule
Backbone & HICO-DET Full & V-COCO S1 & V-COCO S2\\\midrule
Qwen3.5-9B & 33.20 & 44.65 & 49.70\\
InternVL3-8B & 32.11 & 43.60 & 48.40\\
Qwen2.5-VL-7B & 31.61 & 43.10 & 47.90\\\bottomrule
\end{tabular}
\end{center}
\end{table}

InternVL3-8B obtains 32.11\% HICO-DET Full mAP and 43.60\%/48.40\% V-COCO S1/S2 AP, while Qwen2.5-VL-7B obtains 31.61\% and 43.10\%/47.90\%, respectively. The same three-stage harness supports action-triplet and semantic-role prediction across these backbones, demonstrating its applicability to different MLLMs.

\FloatBarrier
\section{Inference Computation and Stability}
\label{app:inference}
We analyze the computational cost and generation stability of \method{}. Comparisons with repeated direct generation examine the benefit of coordinating perception and reasoning, while ground-truth-assisted ranking separates candidate quality from confidence estimation.

\subsection{Computational Structure}
Perceptual generation comprises participant localization, interaction discovery, and local observation. Discovery groups share the initially localized participants and can be processed concurrently. Let $H_0$ be the number of initially localized humans and $R$ the number of local observations. With four humans per discovery group, the number of generative responses is
\begin{equation}
 N_{\mathrm{gen}}=1+\left\lceil\frac{H_0}{4}\right\rceil+R,
 \qquad 0\leq R\leq2.
 \label{eq:app_generation_complexity}
\end{equation}
For $C$ distinct crop windows, the visual context comprises $1+C$ views, with $C\leq R$. Their encoded representations are shared across the associated generation and verification requests.

Candidate verification evaluates independent questions through first-token answer scores. Geometric adjudication selects the participant groundings used in pair assessment and action or role verification. Independent questions within each stage are processed with continuous batching. Let $H=|\Hh^*|$, $N=|\E^*|$, $P=H(N-1)$, $K=|\K|$, and let $G$ count identities with multiple geometry candidates. The number of scoring requests on HICO-DET is
\begin{equation}
 N_{\mathrm{score}}^{\mathrm{HICO}}
 =G+2P+\sum_{(h,o)\in\K}|\mathcal{A}^{+}(c_o^*)|.
 \label{eq:app_hico_complexity}
\end{equation}
The two pair terms correspond to interaction assessment and no-interaction assessment. Positive-action verification covers the category-compatible actions of each retained pair. On V-COCO, the scoring workload is
\begin{equation}
 N_{\mathrm{score}}^{\mathrm{VCOCO}}
 =G+51H+P+25K.
 \label{eq:app_vcoco_complexity}
\end{equation}
Agent-action and role-state assessment contribute $26H$ and $25H$ requests, respectively; each retained pair contributes 25 role-binding requests. In both benchmarks, the verification workload depends on the number of grounded participants and retained relation candidates.

\subsection{Measured Computation}
Table~\ref{tab:runtime_cost} reports the per-image computation and hardware cost of the harness. Perceptual generation requires an average of 2.51 and 2.78 responses per image on HICO-DET and V-COCO, respectively, with a median of two on both benchmarks. Candidate verification processes independent scoring questions in batches. Table~\ref{tab:query_breakdown} reports their mean, median, and 95th percentile for each type of question, reflecting the variation in candidate-set size across images.
\begin{table}[!htb]
\caption{Per-image inference workload and hardware cost. Generative responses and independent scoring requests are reported separately. Output tokens refer to perceptual generation; P95 denotes the 95th percentile over images.}
\label{tab:runtime_cost}
\begin{center}
\small
\begin{tabular}{lrr}\toprule
Quantity & HICO-DET & V-COCO\\
\midrule
\multicolumn{3}{l}{\emph{Perceptual generation}}\\
Generative responses (mean) & 2.51 & 2.78\\
Generative responses (median) & 2 & 2\\
Generative responses (P95) & 5 & 6\\
Output tokens (mean) & 563.29 & 615.26\\
Local observations (mean) & 0.19 & 0.40\\
Distinct image views (mean) & 1.18 & 1.38\\
\midrule
\multicolumn{3}{l}{\emph{Candidate verification}}\\
Scoring requests (mean) & 119.81 & 311.27\\
Scoring requests (median) & 28 & 158\\
Scoring requests (P95) & 470 & 1,011\\
\midrule
\multicolumn{3}{l}{\emph{Hardware measurements}}\\
GPU-worker seconds / image & 13.26 & 22.07\\
Scoring input cache hits (\%) & 88.43 & 89.93\\
\bottomrule
\end{tabular}
\end{center}
\end{table}
\begin{table}[!htb]
\caption{Mean independent scoring requests per image. Each request evaluates a candidate question through first-token answer scores, with independent requests processed in continuous batches.}
\label{tab:query_breakdown}
\begin{center}
\small
\begin{tabular}{lrr}\toprule
Scoring task & HICO-DET & V-COCO\\
\midrule
Geometry selection & 0.17 & 0.33\\
Pair assessment & 43.10 & 50.82\\
No-interaction assessment & 43.10 & --\\
Positive-action verification & 33.44 & --\\
Agent assessment & -- & 86.73\\
Role visibility & -- & 83.40\\
Role binding & -- & 89.99\\
\bottomrule
\end{tabular}
\end{center}
\end{table}

\paragraph{Hardware cost.}
Hardware cost is measured independently as amortized GPU-worker time, computed by summing the recorded processing time across GPU workers and dividing by the number of images. The measurement includes input preparation, model inference, and result collection, with all recorded inference work counted once; model loading and evaluation are excluded. On HICO-DET and V-COCO, the measured costs are 13.26 and 22.07 GPU-worker seconds per image, respectively.

\paragraph{Execution settings.}
Hardware measurements use frozen Qwen3.5-9B in BF16 on NVIDIA A6000 GPUs, with one model replica per GPU, four active images per replica, and up to 32 concurrent requests per GPU. The harness measurement uses four GPUs. Continuous batching schedules generation and scoring requests, while shared visual representations and prefix caching reuse the common image and prompt context. The cache hit rates in Table~\ref{tab:runtime_cost} measure the fraction of scoring input tokens served from the prefix cache.

\paragraph{Ablation computation.}
Pair, label, and factor ablations use the same relation supports to evaluate alternative candidate sets and score compositions. The V-COCO joint action-and-role control evaluates new questions with the view assignment specified in Appendix~\ref{app:reasoning}. Perceptual ablations use the participants and views produced by each configuration. Shared computations are counted once when measuring a complete inference configuration.

\subsection{Allocating Inference Computation}
We compare the harness with repeated direct generation to examine how coordinated perception and reasoning affect detection performance and computational cost. Both methods use the same frozen backbone, full-image processing, sampling settings, and initial output budget. Repeated generation independently predicts participants and relations in each response, whereas the harness uses interaction hypotheses to acquire visual evidence for grounded verification.

\paragraph{Repeated generation and fusion.}
We combine the first $K\in\{1,2,4,8\}$ responses in a predetermined order using two fusion rules. Exact-match fusion retains the maximum score among predictions with identical categories, roles, target states, and boxes. Pair-IoU suppression removes lower-scored duplicates at threshold 0.5 within the same category, role, and target state. The hardware cost at each $K$ sums the recorded GPU-worker processing time of the constituent generation runs per image. Tables~\ref{tab:inference_analysis} and~\ref{tab:generation_fusion} report both rules on the same response sets.
\begin{table}[!htb]
\caption{Detection performance and measured hardware cost. HICO-DET reports Default Full mAP and V-COCO reports S1/S2 role AP (\%). Cost is amortized GPU-worker seconds per image. Repeated generation uses pair-IoU suppression; Table~\ref{tab:generation_fusion} compares both fusion rules.}
\label{tab:inference_analysis}
\begin{center}
\small
\begin{tabular}{lrrrrr}\toprule
& \multicolumn{2}{c}{HICO-DET} & \multicolumn{3}{c}{V-COCO}\\
\cmidrule(lr){2-3}\cmidrule(l){4-6}Method & Full & Cost & S1 & S2 & Cost\\
\midrule
Direct generation, $K=1$ & 15.61 & 2.50 & 16.38 & 21.18 & 2.60\\
Direct generation, $K=2$ & 17.62 & 4.98 & 19.39 & 24.19 & 5.22\\
Direct generation, $K=4$ & 19.04 & 9.97 & 20.86 & 25.54 & 10.50\\
Direct generation, $K=8$ & 19.84 & 19.95 & 22.21 & 26.50 & 20.94\\
\method{} & 33.20 & 13.26 & 44.65 & 49.70 & 22.07\\
\bottomrule
\end{tabular}
\end{center}
\end{table}

\begin{table}[!htb]
\caption{Prediction fusion for repeated direct generation (AP, \%). Exact-match fusion retains the maximum score among identical detections. Pair-IoU suppression uses threshold 0.5 within matching categories, roles, and target states. Both rules use the same fixed response prefixes.}
\label{tab:generation_fusion}
\begin{center}
\small
\begin{tabular}{lrrrr}\toprule
Fusion & Responses & HICO-DET Full & V-COCO S1 & V-COCO S2\\
\midrule
Exact-match fusion & 1 & 15.62 & 16.41 & 21.20\\
 & 2 & 14.53 & 13.31 & 16.55\\
 & 4 & 12.78 & 9.43 & 11.61\\
 & 8 & 10.87 & 6.51 & 7.84\\
\midrule
Pair-IoU suppression & 1 & 15.61 & 16.38 & 21.18\\
 & 2 & 17.62 & 19.39 & 24.19\\
 & 4 & 19.04 & 20.86 & 25.54\\
 & 8 & 19.84 & 22.21 & 26.50\\
\bottomrule
\end{tabular}
\end{center}
\end{table}

\begin{figure}[!htb]
\centering
\includegraphics[width=\linewidth]{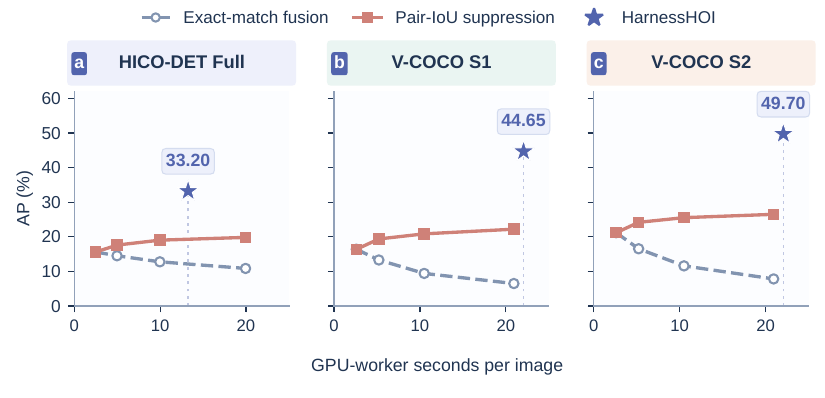}
\caption{\textbf{Detection performance and hardware cost.} Direct-generation curves use $K=1,2,4,8$ responses in increasing computation order, with exact-match fusion or pair-IoU suppression. Stars indicate the complete harness. Cost is amortized GPU-worker time per image; values are reported in Tables~\ref{tab:inference_analysis} and~\ref{tab:generation_fusion}.}
\label{fig:computation_frontier}
\end{figure}

\paragraph{Performance and computation.}
Figure~\ref{fig:computation_frontier} presents detection performance alongside hardware cost. With pair-IoU suppression, increasing the number of responses from one to eight raises HICO-DET mAP from 15.61\% to 19.84\% and V-COCO AP from 16.38\%/21.18\% to 22.21\%/26.50\%. The measured eight-response costs are 19.95 and 20.94 GPU-worker seconds per image, respectively. \method{} achieves 33.20\% and 44.65\%/49.70\%, with measured costs of 13.26 and 22.07 seconds per image. The gains over repeated generation demonstrate the effectiveness of coordinating visual exploration and relation verification within the harness.

\paragraph{Fusion and candidate construction.}
Repeated sampling produces additional interaction candidates and overlapping detections with slightly different boxes. Exact-match fusion retains these duplicates, and AP decreases as more responses are combined. Pair-IoU suppression improves AP by removing overlapping detections from the same responses. The following analysis evaluates candidate quality through the best AP attainable by reordering the predicted boxes and labels.

\subsection{Fixed-Candidate Ranking Analysis}
\label{app:ranking}
We evaluate the quality of the generated candidates by fixing their boxes and labels and using ground-truth annotations to determine the optimal ranking.

\paragraph{Single-response candidates.}
Table~\ref{tab:rank_upper} compares the harness with ground-truth-assisted ranking of a single direct-generation response. The resulting ranking upper bounds are 21.58\% HICO-DET mAP and 30.69\%/35.01\% V-COCO AP. \method{} exceeds these bounds by 11.62 and 13.96/14.69 points, respectively, demonstrating that its advantage extends beyond confidence estimation to the participants and relations available for prediction.

\begin{table}[!htb]
\caption{Ground-truth-assisted ranking of fixed direct-generation predictions. GT ranking retains the generated boxes and labels and optimally orders the candidates using annotations. HICO-DET reports Default Full mAP; V-COCO reports S1/S2 role AP (\%).}
\label{tab:rank_upper}
\begin{center}
\small
\begin{tabular}{lrrr}
\toprule
Method & HICO-DET Full & V-COCO S1 & V-COCO S2\\
\midrule
Direct HOI generation & 15.62 & 16.41 & 21.20\\
Direct HOI generation, GT ranking & 21.58 & 30.69 & 35.01\\
\method{} & 33.20 & 44.65 & 49.70\\
\bottomrule
\end{tabular}
\end{center}
\end{table}

\paragraph{Interaction candidates beyond repeated generation.}
With eight responses, exact-match fusion reaches ranking upper bounds of 30.50\% HICO-DET mAP and 41.90\%/46.40\% V-COCO AP. \method{} exceeds these bounds with its predicted scores by 2.70 and 2.75/3.30 points, respectively, and also exceeds the pair-IoU bounds of 30.09\% and 41.40\%/45.90\%. The advantage persists even when the repeated-generation predictions are ranked optimally, demonstrating the importance of interaction-guided participant discovery and relation construction for HOI detection.

\begin{table}[!htb]
\caption{Ground-truth ranking upper bounds for fixed candidates from repeated direct generation (AP, \%). Each row preserves the corresponding fused boxes, labels, and candidate count, and uses ground truth to place correct detections before false positives. HICO-DET averages all 600 classes. The last row uses the harness's predicted scores.}
\label{tab:repeated_upper}
\begin{center}
\small
\begin{tabular}{lrrrr}\toprule
Candidate construction & Responses & HICO-DET Full & V-COCO S1 & V-COCO S2\\
\midrule
Exact-match fusion & 1 & 21.58 & 30.69 & 35.01\\
 & 2 & 24.30 & 34.80 & 38.90\\
 & 4 & 27.80 & 39.20 & 43.10\\
 & 8 & 30.50 & 41.90 & 46.40\\
\midrule
Pair-IoU suppression & 1 & 21.55 & 30.58 & 34.90\\
 & 2 & 24.09 & 34.50 & 38.60\\
 & 4 & 27.39 & 38.80 & 42.70\\
 & 8 & 30.09 & 41.40 & 45.90\\
\midrule
\method{} (predicted scores) & -- & 33.20 & 44.65 & 49.70\\
\bottomrule
\end{tabular}
\end{center}
\end{table}

\paragraph{Ground-truth ranking.}
For each response count and fusion rule, we compute the maximum AP obtainable by reordering the fixed predictions. Ground-truth matching identifies one correct prediction per covered annotation and ranks these predictions before false positives. Boxes, labels, and candidate counts remain unchanged, and HICO-DET averages over all 600 classes.

\paragraph{Ranking bound under fixed matching.}
The evaluators assign each prediction to its best-matching ground-truth instance before duplicate assignment: HICO-DET uses participant-pair overlap, while V-COCO first selects the best-matching person and then assesses its role target. A duplicate does not switch to a second-best annotation. For fixed predicted boxes and labels, the set of annotations with an eligible matching prediction is therefore fixed. Let $M_c$ be the number of such annotations for class $c$ and $G_c>0$ its ground-truth count. Placing one eligible prediction for each of these annotations before false positives attains precision one up to recall $M_c/G_c$. Under the respective AP definitions, the class-wise bounds are
\begin{equation}
 \begin{aligned}
 U_c^{\mathrm{HICO}}
 &=\frac{\ind[M_c>0]}{11}
   \sum_{t\in\mathcal{R}_{\mathrm{AP}}}\ind\!\left[t\leq\frac{M_c}{G_c}\right],\\[3pt]
 U_c^{\mathrm{VCOCO}}
 &=\frac{M_c}{G_c}.
 \end{aligned}
 \label{eq:app_ranking_bound}
\end{equation}
Here $\mathcal{R}_{\mathrm{AP}}$ is the HICO-DET evaluator's recall-threshold sequence. V-COCO uses the matching rule of each role scenario, with S1 and S2 ranked separately. Averaging the class-wise bounds gives the results in Tables~\ref{tab:rank_upper} and~\ref{tab:repeated_upper}.

\subsection{Generation Stability}
\label{app:stability}
We evaluate each method with generation seeds 66, 67, and 68, keeping the backbone, inference policy, and complete test set fixed. For the harness, the seed controls participant localization, discovery, and observation, while grounded verification uses the same first-token scoring rule. Table~\ref{tab:independent_runs} reports the individual runs, and Table~\ref{tab:seed_hico} gives the HICO-DET results under Default, Known Object, Rare, and Non-Rare evaluation.

\begin{table}[ht]
\caption{Run-to-run results under generation seeds 66, 67, and 68. The backbone, inference policy, and official test population are fixed for each method. HICO-DET reports Default Full mAP and V-COCO reports S1/S2 role AP (\%). SD is the sample standard deviation across three runs.}
\label{tab:independent_runs}
\begin{center}
\small
\begin{tabular}{llrrr}\toprule
Method & Seed & HICO-DET Full & V-COCO S1 & V-COCO S2\\
\midrule
Direct generation & 66 & 15.62 & 16.41 & 21.20\\
 & 67 & 15.81 & 16.89 & 21.93\\
 & 68 & 15.55 & 16.58 & 21.74\\
 & Mean $\pm$ SD & 15.66 $\pm$ 0.13 & 16.63 $\pm$ 0.24 & 21.62 $\pm$ 0.38\\
\midrule
\method{} & 66 & 33.20 & 44.65 & 49.70\\
 & 67 & 32.79 & 44.80 & 49.75\\
 & 68 & 32.86 & 44.50 & 49.55\\
 & Mean $\pm$ SD & 32.95 $\pm$ 0.22 & 44.65 $\pm$ 0.15 & 49.67 $\pm$ 0.10\\
\bottomrule
\end{tabular}
\end{center}
\end{table}

\begin{table}[ht]
\caption{HICO-DET run-to-run results by evaluation setting and class frequency (mAP, \%). All three seeds evaluate the complete official test set.}
\label{tab:seed_hico}
\begin{center}
\small
\begin{tabular}{lrrrrrr}\toprule
& \multicolumn{3}{c}{Default} & \multicolumn{3}{c}{Known Object}\\
\cmidrule(lr){2-4}\cmidrule(l){5-7}Seed & Full & Rare & Non-Rare & Full & Rare & Non-Rare\\
\midrule
66 & 33.20 & 41.63 & 30.68 & 36.85 & 44.20 & 34.65\\
67 & 32.79 & 41.04 & 30.32 & 36.39 & 43.74 & 34.20\\
68 & 32.86 & 41.17 & 30.38 & 36.50 & 43.87 & 34.30\\
\midrule
Mean & 32.95 & 41.28 & 30.46 & 36.58 & 43.94 & 34.38\\
SD & 0.22 & 0.31 & 0.19 & 0.24 & 0.24 & 0.24\\
\bottomrule
\end{tabular}
\end{center}
\end{table}

\begin{figure}[ht]
\centering
\includegraphics[width=\linewidth]{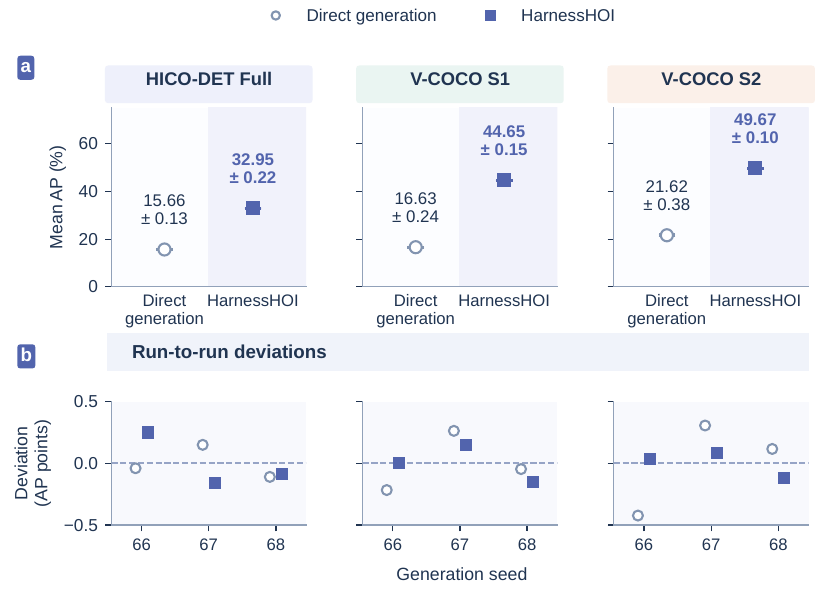}
\caption{\textbf{Consistency across generation runs.} (a) Mean AP with error bars indicating one sample standard deviation across the three generation seeds in Table~\ref{tab:independent_runs}. (b) Individual runs centered on each method's mean, with a common deviation scale across benchmarks. Each method retains the same backbone, inference policy, and test population across runs.}
\label{fig:generation_stability}
\end{figure}

\paragraph{Stability across benchmarks.}
Across three HICO-DET runs, \method{} obtains 32.95\% mean Default Full mAP with a sample standard deviation of 0.22 points, compared with 15.66\% $\pm$ 0.13 for direct generation. Its Default Rare and Non-Rare means are 41.28\% and 30.46\%, and its Known Object Full mean is 36.58\%. The gains remain consistent across seeds, evaluation settings, and frequency groups. V-COCO exhibits the same advantage in all three runs under both role-evaluation scenarios (Table~\ref{tab:independent_runs}).

\section{Qualitative Results}
\label{app:cases}
\begin{figure}[ht]
\centering
\includegraphics[width=\linewidth]{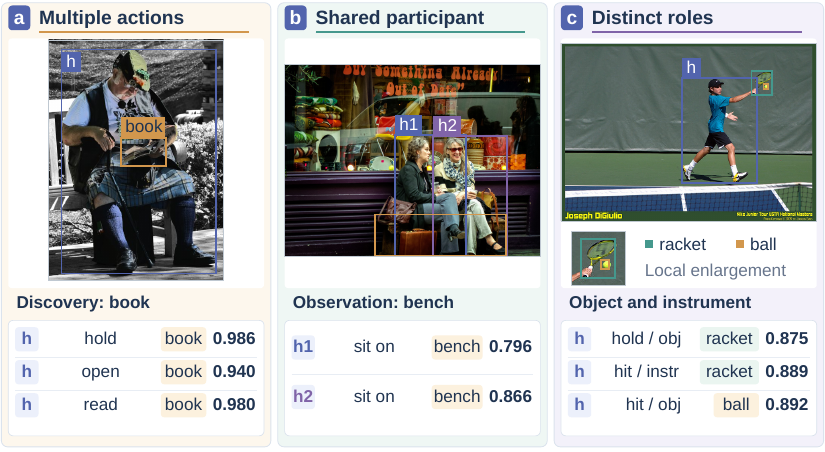}
\caption{\textbf{Interaction-guided perception and grounded reasoning.} (a) Participant discovery supports multiple actions on a shared person-book pair. (b) Targeted observation recovers a bench shared by two people. (c) Compositional verification assigns the object and instrument of a tennis action to the ball and racket. Boxes and labels indicate the predicted participants and relations.}
\label{fig:qualitative}
\end{figure}

Figure~\ref{fig:qualitative} illustrates how the harness coordinates participant discovery, targeted observation, and grounded reasoning. Interaction hypotheses guide the recovery of the book and shared bench. The recovered book supports both \textit{hold} and \textit{read}, and a common bench grounding supports the sitting relations of two people. In the tennis scene, the racket participates in both \textit{hold} and \textit{hit}, while role verification distinguishes the racket as the instrument and the ball as the object of hitting. These examples complement the coverage analysis in Appendix~\ref{app:coverage} by showing how the acquired participants support multiple actions and semantic roles.

\begin{figure}[!htb]
\centering
\includegraphics[width=\linewidth]{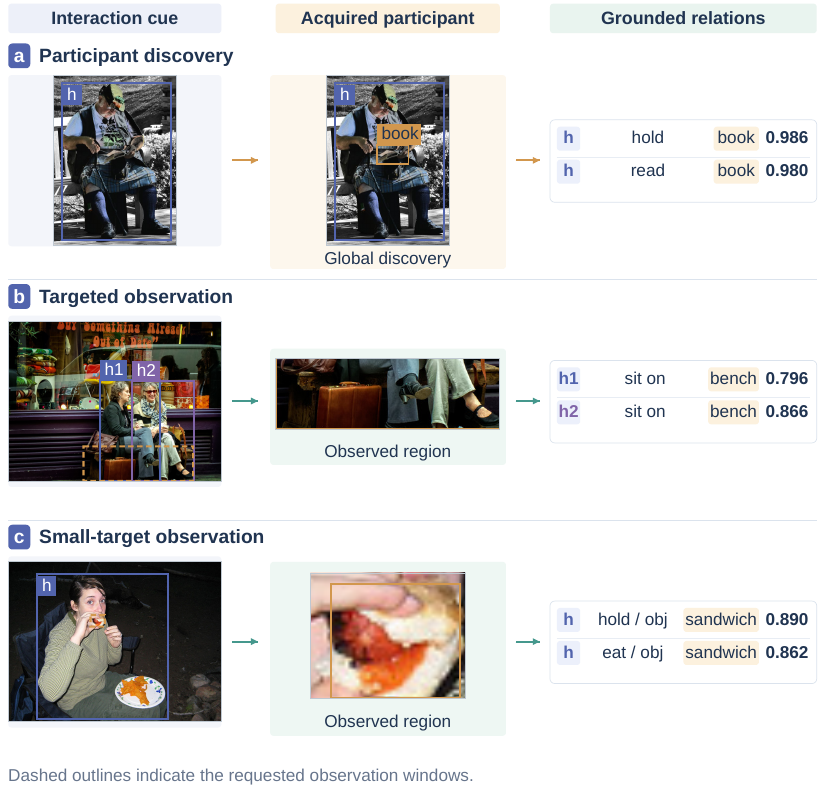}
\caption{\textbf{Interaction-guided participant acquisition.} Selected relations from the book, shared-bench, and sandwich examples. The book is introduced through global discovery. Two sitting hypotheses request the same local window, where observation introduces a bench used by both people. In the eating example, observation localizes the small food target, supporting both holding and eating. Each sequence connects the interaction cue, the acquired participant, and its final grounded relation questions.}
\label{fig:app_perception_cases}
\end{figure}

\begin{figure}[ht]
\centering
\includegraphics[width=\linewidth]{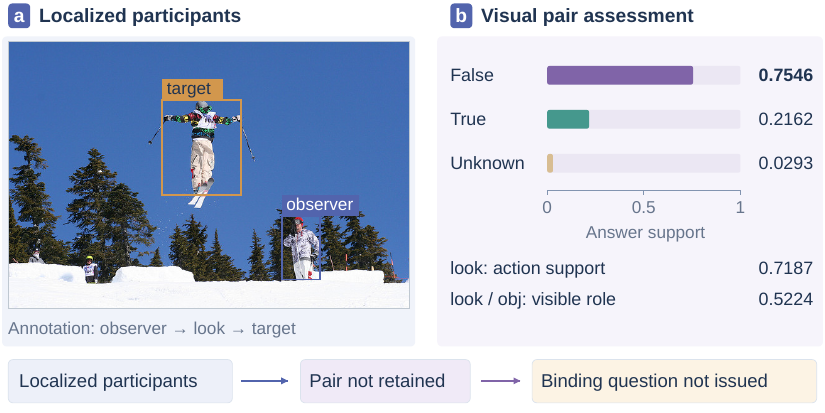}
\caption{\textbf{Pair retention and relation coverage.} Both participants match their annotated boxes, but visual pair assessment assigns its highest support to \textit{false}. Neither discovery nor observation associates them, so the pair is excluded from subsequent \textit{look}-role binding verification. Table~\ref{tab:gaze_questions} reports the corresponding question supports.}
\label{fig:failure}
\end{figure}

\begin{figure}[!htb]
\centering
\includegraphics[width=\linewidth]{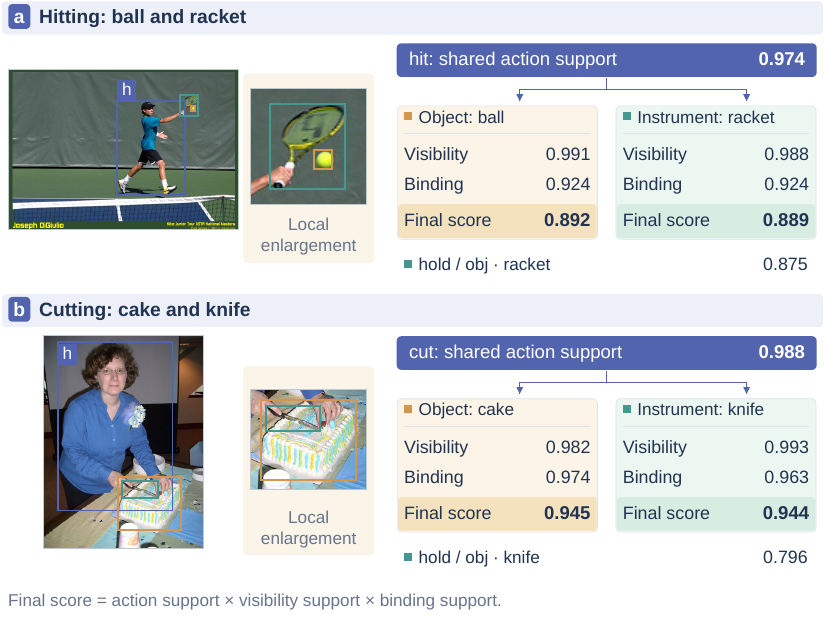}
\caption{\textbf{Compositional role verification.} Selected predictions from the tennis and cutting examples. The racket and knife fulfill instrument roles, whereas the ball and cake fulfill object roles. Each action contributes a shared agent support, combined with role-specific visibility and binding supports. Holding and the instrument role can share the same target grounding. }
\label{fig:app_role_cases}
\end{figure}

Figures~\ref{fig:app_perception_cases}, \ref{fig:failure}, and~\ref{fig:app_role_cases} provide detailed examples of participant recovery, pair selection, and semantic-role assignment, with the corresponding visual-question scores reported below.

\subsection{Participant Discovery and Targeted Observation}
\paragraph{Recovering a book.}
In HICO-DET image \texttt{00000116}, interaction discovery recovers a book omitted from initial localization and associates it with the person's holding and reading actions. The two actions are then verified independently using the same person-book groundings. Table~\ref{tab:case_grounding} reports the localization overlaps and relation scores.

\paragraph{A bench shared by two people.}
In HICO-DET image \texttt{00001195}, the sitting hypotheses of two people direct observation to a shared crop containing the supporting bench. Observation recovers the bench, and both sitting relations use its selected grounding. One person-bench pair is retained through the observation association and the other through visual pair assessment, illustrating their complementary contributions to relation coverage.

The same observation supplies an alternative suitcase box. Comparing both candidates against the acquired views, geometric adjudication selects the original box as the shared grounding for subsequent relation verification.

\paragraph{A small food target.}
In V-COCO image \texttt{18193}, the eating hypothesis directs observation to a small sandwich omitted from the initial participant set. The recovered sandwich is verified as the target of both holding and eating, showing how observation guided by one interaction can support another action involving the same participant. Table~\ref{tab:case_role_factors} gives the corresponding action, visibility, and binding factors.

\begin{table}[!htb]
\caption{Selected interaction groundings and scores in the qualitative cases. Each IoU is computed against the ground-truth boxes of the listed interaction annotation. Repeated targets share the same selected geometry across their relations.}
\label{tab:case_grounding}
\begin{center}
\small
\begin{tabular}{llrrr}\toprule
Case & Action and target role & Person IoU & Target IoU & Score\\\midrule
Book & hold book & 0.9419 & 0.6543 & 0.9859\\
 & read book & 0.9545 & 0.7631 & 0.9796\\
Shared bench & sit\_on bench & 0.9497 & 0.8209 & 0.7959\\
 & sit\_on bench & 0.8886 & 0.8209 & 0.8658\\
Sandwich & hold / obj & 0.9659 & 0.8844 & 0.8896\\
 & eat / obj & 0.9659 & 0.8844 & 0.8616\\
Tennis & hold / obj & 0.9615 & 0.9477 & 0.8748\\
 & hit / instr & 0.9615 & 0.9477 & 0.8889\\
 & hit / obj & 0.9615 & 0.8294 & 0.8924\\
Cutting & hold / obj & 0.9845 & 0.9382 & 0.7955\\
 & cut / instr & 0.9845 & 0.9382 & 0.9443\\
 & cut / obj & 0.9845 & 0.8939 & 0.9445\\
\bottomrule
\end{tabular}
\end{center}
\end{table}

\paragraph{Grounded multi-label verification.}
\label{app:qa_example}
For the person-book pair in HICO-DET image \texttt{00000116}, the selected boxes, full image, and public action definition specify the visual question: whether the person holds the book. The MLLM assesses this question over \{\texttt{false}, \texttt{true}, \texttt{unknown}\}. Normalizing the first-token logits by Equation~\ref{eq:support} gives a holding score of 0.9859. A separate reading question over the same participants gives 0.9796 (Table~\ref{tab:case_grounding}). Independent action assessments therefore retain both interactions on the shared groundings. Table~\ref{tab:case_role_factors} presents the corresponding factor composition for V-COCO roles.

\subsection{Pair Retention in a Distant Gaze Relation}
\paragraph{Distant gaze.}
V-COCO image \texttt{2473} contains an annotated gaze relation between two people. The observer and target boxes match their annotations at IoU 0.9560 and 0.9517. Although the agent and role-state questions favor looking and a visible target, the pair question assigns its highest support to \texttt{false} (Table~\ref{tab:gaze_questions}). Neither discovery nor observation associates this pair. Equation~\ref{eq:route} therefore excludes it from the binding domain, and no physical \texttt{look/obj} prediction is produced for that pair.

\begin{table}[!htb]
\caption{Visual-question supports for the gaze example. Agent and role-state questions assess the human's action and target visibility; the pair decision determines whether the specific target reaches binding verification.}
\label{tab:gaze_questions}
\begin{center}
\small
\begin{tabular}{lrrr}\toprule
Question & False / missing & True / visible & Unknown\\\midrule
Observer performs \texttt{look} & 0.2644 & 0.7187 & 0.0169\\
Visibility of \texttt{look/obj} & 0.4069 & 0.5224 & 0.0707\\
Observer-target pair & 0.7546 & 0.2162 & 0.0293\\\bottomrule
\end{tabular}
\end{center}
\end{table}

\par\noindent
In this example, pair selection excludes the gaze relation despite accurate participant localization. In the shared-bench example, observation associations and visual pair assessment retain different person-bench pairs, illustrating how their combination extends the relations available for verification.

\subsection{Shared Geometry and Distinct Semantic Roles}
\paragraph{Object and instrument in hitting.}
In V-COCO image \texttt{579060}, initial localization identifies the person, racket, and ball. The racket is both a held object and the instrument of hitting, while the ball fulfills the object role of hitting. The two hitting roles share action evidence, with separate visibility and binding questions identifying their respective targets (Table~\ref{tab:case_role_factors}).

\paragraph{Object and instrument in cutting.}
V-COCO image \texttt{468608} similarly distinguishes a knife as the cutting instrument and a cake as the object being cut. Both are available from initial localization, and their role predictions combine shared action evidence with target-specific assessments. The knife also participates in holding, illustrating the use of one grounding across different action-role assignments.

\begin{table}[!htb]
\caption{Compositional role scores in the qualitative examples. $q_A$, $q_V$, and $q_B$ denote action, visible-role, and binding support. The displayed score is computed from the unrounded factors.}
\label{tab:case_role_factors}
\begin{center}
\small
\begin{tabular}{llrrrr}\toprule
Case & Action / role & $q_A$ & $q_V$ & $q_B$ & $q_Aq_Vq_B$\\\midrule
Sandwich & hold / obj & 0.9770 & 0.9769 & 0.9320 & 0.8896\\
 & eat / obj & 0.9890 & 0.9525 & 0.9146 & 0.8616\\
Tennis & hold / obj & 0.9399 & 0.9669 & 0.9627 & 0.8748\\
 & hit / instr & 0.9740 & 0.9876 & 0.9241 & 0.8889\\
 & hit / obj & 0.9740 & 0.9914 & 0.9241 & 0.8924\\
Cutting & hold / obj & 0.9323 & 0.9234 & 0.9241 & 0.7955\\
 & cut / instr & 0.9875 & 0.9933 & 0.9626 & 0.9443\\
 & cut / obj & 0.9875 & 0.9820 & 0.9740 & 0.9445\\
\bottomrule
\end{tabular}
\end{center}
\end{table}

\FloatBarrier
\section{Prompts and Visual Questions}
\label{app:prompts}
We provide the prompt instructions used for perception and relation verification. Each prompt incorporates the relevant participants, image views, and public interaction definitions.

\subsection{Participant Localization}
Initial localization receives the full image and public object vocabulary and returns categories, boxes normalized to $[0,1000]$, and appearance references:
\begin{quote}
Localize the people and identifiable targets involved in the scene's interactions. Organize the output by interaction group: list one person and all associated targets immediately after that person, then proceed to the next person. Cover holding, using, riding, sitting or leaning, wearing, carrying, caring, and person-to-person interactions. Include identifiable nearby objects that may support no-interaction assessment, while prioritizing interaction participants over unrelated background objects.

Give one box for each complete visible instance and include each instance once. A person's box should cover the visible person; an object's box should cover the object itself. Describe each reference through visible appearance and position. Return the instance category, normalized box, and reference in the requested format.
\end{quote}

\subsection{Interaction Discovery and Observation Planning}
Discovery receives the full image, initial entities, designated human group, and public interaction definitions. The prompt requests hypotheses, additional participants, and observation regions:
\begin{quote}
Inspect every designated human. Give concise visible evidence and identify the interactions involving that person. A person can perform several actions and interact with several targets. Use existing instance identities when their targets are already localized. Add an omitted, visually identifiable participant with its category, full-image box, and appearance reference. For role-based actions, distinguish the target fulfilling each object or instrument role.

Examine held, worn, ridden, supporting, and distant visible targets. When the target cannot yet be localized reliably, retain its description and formulate a question for closer observation. Propose up to two regions, giving priority to unlocalized targets and then incomplete instance boundaries. Each request identifies the human, possible action and role, known target when available, target description, localization question, image region, and priority. The region should support localization of the complete visible instance. When the existing view resolves participant localization, return no observation request.
\end{quote}

\subsection{Targeted Observation}
Observation receives a crop, the target description, and the local instances eligible for refinement. Output coordinates are normalized within the crop:
\begin{quote}
Use this crop and target description to localize the relevant participants. Identify changes to the eligible instances and any additional visible participants. For an existing instance, retain its identity and report the changed category or box. Instances crossing the crop boundary provide context; update complete-instance boxes for eligible identities whose extents lie within the view.

Describe newly identified participants with complete visible-instance boxes and categories. Preserve the distinction between a person and the object they hold or use. Localize visible extent without extrapolating occluded boundaries. Return instance updates and new participants. If the observation provides no reliable localization change, return empty changes.
\end{quote}

Returned boxes are mapped to the full image by Equation~\ref{eq:app_crop_mapping}.

\subsection{Geometric Adjudication}
The geometry question receives the candidate category-and-box descriptions, full image, and acquired crops. Each candidate is assigned an answer letter:
\begin{quote}
Compare the candidate descriptions of this instance against the image evidence. Select the candidate that best matches its identity, category, and complete visible boundary. For a person, assess the full visible person rather than a hand or torso. For an object, assess the extent of that object rather than its contact region with another participant. Judge localization correspondence; the candidate order does not indicate quality. Answer with the letter of the selected candidate.
\end{quote}

Candidate supports are normalized by Equation~\ref{eq:support}, and the grounding is selected by Equation~\ref{eq:geometry_selection}.

\subsection{Grounded Relation Questions}
\paragraph{Shared instruction.}
Relation questions receive the selected participants, action or role definition, and corresponding image views. They share the following instruction:
\begin{quote}
Assess the specified instances from the image evidence. Use the given boxes to identify the person and target, and verify that those instances correspond to the visual question. Distinguish this person's action from another person's action and this target from another instance of its category. An interaction requires visual evidence beyond proximity or the object's typical use. Multiple actions can hold for the same participants. Apply the supplied action and role meanings and answer with one word from the designated answer set.
\end{quote}

Table~\ref{tab:question_semantics} lists the question types and answer sets.
\begin{table}[!htb]
\caption{Semantic decisions in grounded question answering. Answer supports are obtained from first-token logits normalized within the designated set.}
\label{tab:question_semantics}
\begin{center}
\small
\begin{tabularx}{\linewidth}{@{}l>{\raggedright\arraybackslash}Xl@{}}\toprule
Question & Visual judgment & Answer set\\\midrule
Pair & Whether any legal positive action or physical-role binding connects the specified human and target, including distant and non-contact relations & false / true / unknown\\
No interaction & Whether both instances and their boxes are valid and no legal positive interaction holds between them & false / true / unknown\\
Action & Whether the specified participant pair and action jointly have visual support & false / true / unknown\\
Agent & Whether the specified human performs the action, independently of target visibility & false / true / unknown\\
Role state & Given the action, whether its role has a localizable visible target, has a missing target, or remains unresolved & missing / visible / unknown\\
Binding & Given the action and visible role, whether this particular target fulfills that role & false / true / unknown\\\bottomrule
\end{tabularx}
\end{center}
\end{table}

\paragraph{Pair assessment.}
The pair question asks whether any legal positive interaction is visually supported. Its answers denote support for at least one relation (\texttt{true}), rejection of all such relations (\texttt{false}), or unresolved evidence (\texttt{unknown}). Pair retention follows Equation~\ref{eq:route}, combining this assessment with discovery and observation associations.

\paragraph{No-interaction and action assessment.}
No-interaction assessment verifies the participants' identity and geometry together with the absence of legal positive interactions. Each positive-action question evaluates a specific predicate under its supplied definition: for example, \texttt{hold} concerns bodily or manual support of the target, and \texttt{read} concerns engagement with its textual content. Each question produces a separate false/true/unknown distribution.

\paragraph{Role visibility and binding.}
Conditional on the action, \texttt{visible} denotes a localizable role target in the image, including one absent from the entity set; \texttt{missing} denotes the absence of such a target; and \texttt{unknown} denotes unresolved visibility. Binding assesses whether the specified instance fulfills the object or instrument role.

\subsection{Public Role Definitions}
\begin{table}[ht]
\caption{Complete public target-role meanings used by V-COCO questions. Role identifiers follow the released benchmark taxonomy.}
\label{tab:role_semantics}
\begin{center}
\small
\begin{tabular}{lll}\toprule
Action & Role & Target meaning\\\midrule
hold & obj & Object held by the person\\
sit & instr & Seat or surface supporting the sitting person\\
ride & instr & Vehicle or animal being ridden\\
look & obj & Target of the person's gaze\\
hit & instr & Implement used for hitting, such as a racket or bat\\
hit & obj & Target being hit, such as a ball\\
eat & obj & Food being eaten\\
eat & instr & Utensil used for eating\\
jump & instr & Target or supporting equipment involved in the jump\\
lay & instr & Object supporting the lying person\\
talk\_on\_phone & instr & Phone used for the conversation\\
carry & obj & Object being carried or transported\\
throw & obj & Object being thrown\\
catch & obj & Moving object being caught\\
cut & instr & Tool used for cutting\\
cut & obj & Material or food being cut\\
work\_on\_computer & instr & Computer being operated\\
ski & instr & Pair of skis used for skiing\\
surf & instr & Surfboard used on water\\
skateboard & instr & Skateboard used for the activity\\
drink & instr & Cup, bottle, or other drinking container\\
kick & obj & Target being kicked\\
point & instr & Target indicated by the gesture\\
read & obj & Book, newspaper, or other text being read\\
snowboard & instr & Snowboard used on snow\\
\bottomrule
\end{tabular}
\end{center}
\end{table}

Table 42 gives the complete target meanings used in V-COCO questions. The role keys follow the benchmark representation. In particular, \texttt{point/instr} denotes the target indicated by the pointing gesture. Action definitions complement these role meanings by specifying the person's activity: holding a phone and talking on a phone are assessed separately; standing on a board does not alone establish riding or jumping; and skiing, snowboarding, skateboarding, and surfing retain their distinct activity meanings. HICO-DET questions follow the public verb-object taxonomy, with definitions distinguishing predicates such as \texttt{cut} and \texttt{cut\_with}, \texttt{sit\_at} and \texttt{sit\_on}, and \texttt{hold} and \texttt{carry}.

\end{document}